%% file: main.tex
\documentclass{article}

\usepackage{microtype}
\usepackage{graphicx}
\usepackage{subfigure}
\usepackage{booktabs} % for professional tables

\usepackage{hyperref}
\usepackage{xr}
\usepackage{float}
\usepackage{comment}
\usepackage{siunitx}
\usepackage{relsize}
\usepackage{ifthen}
\usepackage[colorinlistoftodos]{todonotes}
\usepackage{graphics} % for pdf, bitmapped graphics files
\usepackage{rotating}
\usepackage{color}
\usepackage{enumerate}
\usepackage[T1]{fontenc}
\usepackage{psfrag}
\usepackage{epsfig} % for postscript graphics files
\usepackage{booktabs}
\usepackage{graphicx,url}
\usepackage{multirow}
\usepackage{array}
\usepackage{latexsym}
\usepackage{amsfonts}
\usepackage{amsmath}

\usepackage{amssymb}
\usepackage{xstring}
\usepackage{algorithmic}
\usepackage{multirow}
\usepackage{xcolor}
\usepackage{prettyref}
\usepackage{flexisym}
\usepackage{bigdelim}
\usepackage{breqn} % load this last
\usepackage{listings}

\usepackage{xspace}
\usepackage{bm}
\graphicspath{{./figures/}}
\usepackage{tikz}
\usetikzlibrary{matrix,calc}
\usepackage{lipsum}
\usepackage{mdwlist}

\makecompactlist{itemize}{stditemize}
\usepackage{enumitem}
\usepackage{caption}
\usepackage{epstopdf}
\usepackage{amsthm}
\usepackage{mathtools}

\usepackage{cleveref}
\hypersetup{%
	colorlinks=true,
	linkcolor=blue,
	filecolor=magenta,      
	urlcolor=black,
	citecolor=blue,
	linkbordercolor={0 0 1}
}

\input{preamble_symbols.tex}

\input{shortcuts.tex}

\renewcommand{\algorithmicrequire}{\textbf{Input:}}
\renewcommand{\algorithmicensure}{\textbf{Output:}}

\PassOptionsToPackage{end}{algorithmic}

\usepackage[accepted]{icml2023}

\icmltitlerunning{Efficient Online Learning with Memory via Frank-Wolfe Optimization}

\begin{document}

\twocolumn[
%!TEX root = main.tex

\icmltitle{
 Efficient Online Learning with Memory via Frank-Wolfe Optimization:
	 Algorithms with Bounded Dynamic Regret and Applications to Control
  }
% \begin{center}
%     \textbf{Hongyu Zhou $\qquad$ Zirui Xu $\qquad$ Vasileios Tzoumas} \\
%     \vspace{2mm}
%     Department of Aerospace Engineering, University of Michigan, Ann Arbor \\
%     \{zhouhy, vtzoumas\}@umich.edu
% \end{center}

% \aistatsaddress{ 
% 	Department of Aerospace Engineering \\	
% 	University of Michigan, Ann Arbor \\
% 	{zhouhy@umich.edu}
% 	\And Department of Aerospace Engineering\\
% 	University of Michigan, Ann Arbor  \\
% 	{ziruixu@umich.edu}
% 	\And Department of Aerospace Engineering\\
% 	University of Michigan, Ann Arbor  \\
% 	{vtzoumas@umich.edu}}

% It is OKAY to include author information, even for blind
% submissions: the style file will automatically remove it for you
% unless you've provided the [accepted] option to the icml2021
% package.

% List of affiliations: The first argument should be a (short)
% identifier you will use later to specify author affiliations
% Academic affiliations should list Department, University, City, Region, Country
% Industry affiliations should list Company, City, Region, Country

% You can specify symbols, otherwise they are numbered in order.
% Ideally, you should not use this facility. Affiliations will be numbered
% in order of appearance and this is the preferred way.
% \icmlsetsymbol{equal}{*}
\icmlsetsymbol{equal}{*}

\begin{icmlauthorlist}
\icmlauthor{Hongyu Zhou}{ }
\icmlauthor{Zirui Xu}{ }
\icmlauthor{Vasileios Tzoumas}{ }
\end{icmlauthorlist}

\icmlaffiliation{ }{Department of Aerospace Engineering, University of Michigan, Ann Arbor}

\icmlcorrespondingauthor{Hongyu Zhou}{zhouhy@umich.edu}

% You may provide any keywords that you
% find helpful for describing your paper; these are used to populate
% the "keywords" metadata in the PDF but will not be shown in the document
\icmlkeywords{Online Convex Optimization, Non-Stochastic Control}

\vskip 0.3in
% \vspace{-3mm}
]

% this must go after the closing bracket ] following \twocolumn[ ...

% This command actually creates the footnote in the first column
% listing the affiliations and the copyright notice.
% The command takes one argument, which is text to display at the start of the footnote.
% The \icmlEqualContribution command is standard text for equal contribution.
% Remove it (just {}) if you do not need this facility.

\printAffiliationsAndNotice{}  % leave blank if no need to mention equal contribution
% \printAffiliationsAndNotice{\icmlEqualContribution} % otherwise use the standard text.

\input{Abstract.tex}

\input{1-Introduction.tex}

\input{1-RelatedWork}
\input{2-Problem.tex}

\input{3-Meta-OFW-Alg.tex}
\input{4-Meta-OFW-Analysis.tex}

\input{5-Control.tex}
\input{6-Conclusion.tex}

% \input{Acknowledgements}

\bibliography{References}
\bibliographystyle{icml2023}

\appendix
\input{Appendix/Appendix} %Proofs

\end{document}

%% file: preamble_symbols.tex
\newtheorem{theorem}{Theorem}
\newtheorem{problem}{Problem}

\newtheorem{corollary}{Corollary}

\newtheorem{lemma}{Lemma}
\newtheorem{assumption}{Assumption}
\newtheorem{definition}{Definition}
\newtheorem{proposition}{Proposition}

\newtheorem{remark}{Remark}

\newcommand{\bdmath}{\begin{dmath}}
\newcommand{\edmath}{\end{dmath}}
\newcommand{\beq}{\begin{equation}}
\newcommand{\eeq}{\end{equation}}
\newcommand{\bdm}{\begin{displaymath}}
\newcommand{\edm}{\end{displaymath}}
\newcommand{\bea}{\begin{eqnarray}}
\newcommand{\eea}{\end{eqnarray}}
\newcommand{\beal}{\beq \begin{array}{lll}}
\newcommand{\eeal}{\end{array} \eeq}
\newcommand{\beas}{\begin{eqnarray*}}
\newcommand{\eeas}{\end{eqnarray*}}
\newcommand{\ba}{\begin{array}}
\newcommand{\ea}{\end{array}}
\newcommand{\bit}{\begin{itemize}}
\newcommand{\eit}{\end{itemize}}
\newcommand{\ben}{\begin{enumerate}}
\newcommand{\een}{\end{enumerate}}

\newcommand{\calB}{{\cal B}}

\newcommand{\calH}{{\cal H}}

\newcommand{\calM}{{\cal M}}

\newcommand{\calO}{{\cal O}}

\newcommand{\calX}{{\cal X}}

\definecolor{myblue}{RGB}{65 105 225}

\newcommand{\hide}[1]{}

\newcommand{\hiddenText}{{\color{gray} hidden text.}}
\newcommand{\hideWithText}[1]{\hiddenText}

\newcommand{\scenario}[1]{{\fontsize{9}{8.7}\selectfont\sf#1}\xspace}

%% file: shortcuts.tex
\newcommand{\TimeSeq}{\{1, \dots, T\}}

\newcommand{\myIdentity}{\mathbf{I}}

\newcommand{\myx}{\mathbf{x}}
\newcommand{\myv}{\mathbf{v}}
\newcommand{\SumOneT}{\sum_{t=1}^{T}}

\newcommand{\ie}{\emph{i.e.},\xspace}
\newcommand{\eg}{\emph{e.g.},\xspace}

\newcommand{\red}[1]{{\color{red}#1}}
\newcommand{\blue}[1]{{\color{blue}#1}}

\newcommand{\myParagraph}[1]{{\bf #1.}\xspace}

\newcommand{\OGD}{\scenario{\textbf{OGD}}}
\newcommand{\OMD}{\scenario{\textbf{OMD}}}
\newcommand{\OFW}{\scenario{\textbf{OFW}}}

\newcommand{\MetaOFW}{\scenario{\textbf{Meta-OFW}}}
\newcommand{\Hedge}{\scenario{\textbf{Hedge}}}
\newcommand{\Ader}{\scenario{\textbf{Ader}}}
\newcommand{\Scream}{\scenario{\textbf{Scream}}}
\newcommand{\DyReg}{\operatorname{Regret}_T^D}
\newcommand{\DyRegNSC}{\operatorname{Regret-NSC}_T^D}
\newcommand{\Reg}{\operatorname{S-Reg}_T}
\newcommand{\DReg}{\operatorname{D-Reg}_T}

%% file: Abstract.tex
\begin{abstract}
Projection operations are a typical computation bottleneck in online learning. 
In this paper, we enable projection-free online learning within the framework of \textit{Online Convex Optimization with Memory} (OCO-M) ---OCO-M captures how the history of decisions affects the current outcome by allowing the online learning loss functions to depend on both current and past decisions. 
Particularly, we introduce the first projection-free meta-base learning algorithm with memory that minimizes dynamic regret, \ie that minimizes the suboptimality against \textit{any} sequence of time-varying decisions. 
We are motivated by artificial intelligence applications where autonomous agents need to adapt to time-varying environments in real-time, accounting for how past decisions affect the present.  
Examples of such applications are: online control of dynamical systems; statistical arbitrage; and  time series prediction. 
The algorithm builds on the {Online Frank-Wolfe} (\OFW) and \Hedge algorithms. 
We demonstrate how our algorithm can be applied to the online control of linear time-varying systems in the presence of  unpredictable process noise.  To this end, we develop a controller with memory and bounded dynamic regret against any optimal {time-varying} linear feedback control policy.
We validate our algorithm in simulated scenarios of online control of linear time-invariant systems. 
% Our algorithm is observed to be superior in computational efficiency, achieving comparable or superior loss performance.

\end{abstract}

%% file: 1-Introduction.tex
\section{Introduction}\label{sec:Intro}

Online Convex Optimization (OCO) \cite{shalev2012online,hazan2016introduction} 
% is an online learning framework that has 
has found widespread application in statistics, information theory, and operation research \citep{cesa2006prediction}. OCO can be interpreted as a sequential game between an optimizer and an adversary over $T$ time steps: at each time step $t=1,\ldots, T$, first the optimizer chooses a decision $\myx_t$ from a convex set $\calX$; 
then, the adversary reveals a convex loss function $f_t$ and the optimizer suffers the loss $f_t(\myx_t)$. The optimizer aims to minimize its cumulative loss, despite knowing each $f_t$ only after $\myx_t$ has been already decided. 
% over a given time horizon $T$. 

\textit{Static regret} is the standard approach to measure the suboptimality of the optimizer's decisions $\myx_1,\ldots, \myx_T$.  Particularly, given a decision $\myx \in \calX$ to compare $\myx_1,\ldots, \myx_T$ with, the static regret of $\myx_1,\ldots, \myx_T$ with respect to $\myv$ is defined as follows~\citep{hazan2016introduction}:
\begin{equation}
    \Reg= \sum_{t=1}^{T} f_t(\myx_t) - \sum_{t=1}^{T}f_t(\myv).
    \label{eq:S-Reg}
\end{equation}
That is, when $\myv$ minimizes $\sum_{t=1}^{T}f_t(\myv)$, then $\Reg$ captures the suboptimality of $\myx_1, \dots, \myx_T$ against the optimal \textit{static} decision that would have been made in hindsight.

Algorithms that guarantee \textit{static no-regret} have been widely adopted in applications pertained to recommendation systems, communication-channel allocation, and action prediction~\citep{cesa2006prediction}.\footnote{An algorithm has \textit{static no-regret}  when $\Reg/T$ tends to $0$ when $T$ tends to $+\infty$, implying  $f_t(\myx_t)$ tends to $f_t(\myv)$ for $t$ large.}

But the application of such algorithms to complex artificial intelligence tasks such as \textit{online control under unpredictable disturbances}~\citep{shi2019neural} and
\textit{collaborative multi-robot motion planning} \citep{xu2023online}
% that require adaptive decision-making 
is hindered by three main technological challenges: 
\begin{itemize}[leftmargin=9pt]\setlength\itemsep{-1.mm}
    \item \textbf{Challenge I: Dynamic Environments.} Complex tasks such as the above require decisions that {adapt} to changing environments.  For example, \textit{target tracking with multiple robots} requires the robots to continuously change their position to track moving targets~\citep{xu2023online}.  Therefore, measuring performance against a static (optimal) decision per the static regret in \cref{eq:S-Reg} is insufficient.  Instead, we need to measure performance against \textit{time-varying} (optimal) decisions.

    \item \textbf{Challenge II: Past Decisions Affect the Present.} In complex tasks such as the aforementioned, past decisions often affect the present outcome.  Therefore, the OCO framework we discussed above, where each loss function $f_t$  depends on the most recent decision $\myx_t$ only, fails to capture the effect of earlier decisions to the present.  Instead, we need an OCO framework with memory, where each loss function $f_t$ depends on $\myx_t$ as well as on the past $\myx_{t-m}, \dots, \myx_{t-1}$, for some $m\geq 0$.
    % The standard OCO framework considers only a \textit{memoryless} adversary, \ie the loss is only a function of learner's current decision, and cannot deal with adversaries with memory who decide the loss function based on the learner’s current and past decisions, which is common in many real-world decision-making problems, \eg
    % statistical arbitrage \citep{anava2013online}, multistep-ahead time series prediction \citep{anava2015online}, and online control of dynamical systems \citep{agarwal2019online}. 
    
    \item \textbf{Challenge III: Fast Decision-Making.} Complex control tasks often require decisions to be made fast.  For example, such is the case for the effective \textit{online control of quadrotors against wind disturbances}~\citep{romero2022time}.
     But the current OCO algorithms typically rely on projection operations which can be computationally expensive since they require solving quadratic programs \citep{kalhan2021dynamic}.  Instead, we need fast OCO algorithms that are inevitably projection-free.
    % The learner needs to adapt to changes efficiently due to time-varying environments. For example, during adversarial target tracking, the pursuer needs to adapt to targets’ actions efficiently \citep{zhang2022adversarial,xu2023online}. 
    % Current control algorithms usually relies on model predictive control~\citep{rawlings2017model,borrelli2017predictive} which simulates the future, or online learning~\citep{agarwal2019online,simchowitz2020improper}, which updates control policy performing projection based on past observations. However, such methods can be computationally hard and prevent robots adapting to changing environments efficiently.
\end{itemize}
\vspace{-3mm}

All in all, the above challenges give rise to the need below:

\vspace{.5mm}
\noindent \myParagraph{Need}\textit{We need online learning algorithms for OCO with Memory (OCO-M) that are projection-free and guarantee near-optimal decisions in dynamic environments. The decisions' near-optimality may be captured by bounding their suboptimality with respect to optimal decisions that adapt to the changing environment knowing its future evolution, \ie by bounding \emph{dynamic regret}.} 
\vspace{.5mm}

% Particularly, , in contrast to the static regret in \cref{eq:S-Reg}dynamic regret is introduced to capture how environments change \citep{zinkevich2003online}. 
\textit{Dynamic regret for the classical OCO without memory} is defined as follows \citep{zinkevich2003online}: 
% captures the suboptimality of $\myx_1, \dots, \myx_T$ against a
given a time-varying comparator sequence $\myv_1, \dots, \myv_T$, then\footnote{A related measure to dynamic regret is \textit{adaptive regret} \citep{hazan2007adaptive}.  Adaptive regret captures the worst-case static regret on {any} contiguous time interval.
% and measures the local behavior of learner's decisions against the best fixed decision in hindsight over the time interval. 
\cite{zhang2020online} studies the relation of dynamic regret to adaptive regret.}
% as follows \citep{zinkevich2003online}:
    \begin{equation}
        \DReg = \sum_{t=1}^{T} f_t(\myx_t) - \sum_{t=1}^{T}f_t(\myv_t).
        \label{eq:D-Reg}
    \end{equation}
Dynamic regret contrasts static regret: static regret compares $(\myx_1, \dots, \myx_T)$ against a merely static $\myv$.  
Thus, when $\myv_1, \dots, \myv_T$ minimize $\sum_{t=1}^{T}f_t(\myv_t)$, then $\DReg$ captures the suboptimality of $\myx_1, \dots, \myx_T$ against the optimal \textit{time-varying} decisions that would have been made in hindsight. 
% above reduces to the static regret if the given time-varying decisions $\myv_1, \dots, \myv_t$ are the optimal static decision $\myx^{*}$.
Hence, dynamic regret bounds are typically larger than static regret bounds, depending on terms that capture the change of the environment.  Such terms are \textit{loss variation} $V_T$, \textit{gradient variation} $D_T$, and \textit{path length}~$C_T$:\footnote{Obtaining a no-regret algorithm hence requires the growth of the metrics in \cref{eq:DyReg_metric_VT,eq:DyReg_metric_DT,eq:DyReg_metric_CT} to be sublinear \citep{besbes2015non,mokhtari2016online,kalhan2021dynamic}. $V_T$ and $D_T$ are small when the loss function and decisions change slowly.}
% \begin{equation}
% 	\begin{aligned}
% 		V_{T} & \triangleq \sum_{t=1}^{T} \sup _{\mathbf{x} \in \mathcal{X}}\left|f_{t}(\mathbf{x})-f_{t-1}(\mathbf{x})\right|, \\ 
% 		D_{T} &  \triangleq \sum_{t=1}^{T}\left\|\nabla f_{t}\left(\mathbf{x}_{t}\right)-\nabla f_{t-1}\left(\mathbf{x}_{t-1}\right)\right\|_2^{2}, \\
% 		C_{T} & \triangleq \sum_{t=1}^{T}\left\|\mathbf{v}_{t}-\mathbf{v}_{t-1}\right\|_2.
% 	\end{aligned}
% 	\label{eq:DyReg_metric}
% \end{equation} 
\begin{align}
    V_{T} & \triangleq \sum_{t=1}^{T} \sup _{\mathbf{x} \in \mathcal{X}}\left|f_{t}(\mathbf{x})-f_{t-1}(\mathbf{x})\right|, \label{eq:DyReg_metric_VT}\\
    D_{T} &  \triangleq \sum_{t=1}^{T}\left\|\nabla f_{t}\left(\mathbf{x}_{t}\right)-\nabla f_{t-1}\left(\mathbf{x}_{t-1}\right)\right\|_2^{2},     \label{eq:DyReg_metric_DT} 
\end{align}
\begin{align}
    C_{T} & \triangleq \sum_{t=1}^{T}\left\|\mathbf{v}_{t}-\mathbf{v}_{t-1}\right\|_2.     \label{eq:DyReg_metric_CT}
\end{align}

\textit{Dynamic regret for OCO-M with memory $m$}, where the loss function at each time step $t$ takes the form $f_t(\myx_{t-m},\dots,\myx_t) : \calX^{m+1}\mapsto\mathbb{R}$, 
% \ie it depends on the both current and the last $m$ decisions, instead of only the current decision which is the case for memoryless OCO.  The dynamic regret becomes
is defined as follows:
\begin{equation}\label{def:rel-work-dyn-reg-memory}
    \DyReg = \sum_{t=1}^{T} f_t(\myx_{t-m}, \dots, \myx_{t}) - \sum_{t=1}^{T}f_t(\myv_{t-m}, \dots, \myv_{t}),
\end{equation}
where it is assumed that $\myx_{t-m}=\mathbf{0}$ for $t\leq m$.

\input{Table/Table-comparison}

\myParagraph{Contributions} 
We aim to address the Need by means of the following  contributions:
\begin{itemize}[leftmargin=9pt]\setlength\itemsep{-1.mm}
     \item \textit{Algorithmic Contributions}: 
     We introduce the first projection-free algorithm for OCO-M with bounded dynamic regret (\Cref{sec:Meta-OFW-Alg,sec:Meta-OFW-Analysis}) ---the regret bound is presented in \Cref{table:comparison}. The algorithm builds on the projection-free algorithms \Hedge \citep{freund1997decision} and \OFW \citep{hazan2012projection,kalhan2021dynamic}. 
    %  We provide the first projection-free algorithm for OCO-M . based on \OFW enabling online non-stationary learning in the framework of OCO with memory (OCO-M) with dynamic regret bound of $\calO\left(\sqrt{T(1+V_{T}+D_{T}+C_{T})}\right)$ (\Cref{sec:Meta-OFW-Analysis}). 
    
     We apply our algorithm to the online control of linear time-varying systems in the presence of  unpredictable noise (\Cref{sec:control}). 
    %  We thus introduce the first projection-free controller with memory, and with bounded dynamic regret against any optimal {time-varying} linear feedback controller. 
     We thus introduce a projection-free controller with memory and bounded dynamic regret against any optimal {time-varying} linear feedback control gains.   Particularly, our comparator class of optimal {time-varying} linear feedback control gains does \underline{not} require the a priori knowledge of stabilizing control gains.  Instead, the state-of-the-art OCO-M controller  by~\cite{zhao2022non} requires a comparator class of optimal time-varying policies  where an a priori knowledge of stabilizing control gains is necessary.
    %  , instead of against only static linear feedback policies.
     
    \item \textit{Technical Contributions}: To enable aforementioned algorithmic and regret bound contributions,
    % the projection-free algorithm, and its aforementioned dynamic regret bound and application to the online control of dynamical systems, 
    we make the following technical innovations:
        \begin{itemize}[leftmargin=9pt]
            \item We analyze dynamic regret of the \OFW algorithm (\Cref{sec:Meta-OFW-Alg}). The analysis enables the  state-of-the-art bound in \citep[Theorem~1]{kalhan2021dynamic} to hold true for any convex loss functions 
            % and any comparator sequences 
            in the evaluation of \OFW's regret (see \Cref{table:comparison}).  Instead, \citep[Theorem~1]{kalhan2021dynamic} holds true for smooth convex functions only.
            % , and for comparator sequences that  minimize in hindsight the cumulative loss.  

            \item We prove that 
            the \textit{Disturbance-Action Control} (DAC) policy \citep{agarwal2019online} ---widely used in online non-stochastic control to reduce the online control problem to OCO-M \citep{agarwal2019online,gradu2020adaptive,hazan2020nonstochastic}--- is able to approximate time-varying linear feedback controllers  (\Cref{prop:DAC_sufficiency} in \Cref{app-subsec:DAC_sufficiency}).
            % \Cref{app-subsec:DAC_sufficiency}). 
            % The DAC policy is used .  
            Previous results have established that a DAC policy can approximate time-\underline{in}variant linear feedback controllers only \citep{agarwal2019online}, instead of a time-varying controllers.
\end{itemize}
\end{itemize}

\myParagraph{Numerical Evaluations} {
We validate our algorithm in simulated scenarios of online control of linear time-invariant systems (\Cref{app:sec_control_exp}).  We compare our algorithm with \OGD \citep{zinkevich2003online}, \Ader \citep{zhang2018adaptive}, and \Scream \citep{zhao2022non} algorithms.
% two OCO-with-No-Memory algorithms, \OGD \citep{zinkevich2003online} and \Ader \citep{zhang2018adaptive}, and the OCO-M algorithm \Scream \citep{zhao2022non}. 
Our algorithm is observed 3 times faster than the state-of-the-art OCO-M algorithm \Scream \citep{zhao2022non} as system dimension increases, and achieves comparable or superior loss performance over all compared algorithms.
}

% \myParagraph{Organization}
% \Cref{sec:lit_review} reviews the related work. \Cref{sec:problem} introduces the problem of OCO-M. \Cref{sec:Meta-OFW-Alg} presents a projection-free meta-base algorithm (\ie~\MetaOFW algorithm) to the problem of OCO-M. \Cref{sec:Meta-OFW-Analysis} develops performance guarantee for \MetaOFW algorithm. \Cref{sec:control} presents applications of \MetaOFW algorithm with numerical experiments to the online non-stochastic control problem. The Appendix contains all proofs.

% \input{Table/Table-comparison}

%% file: Table/Table-comparison.tex
\renewcommand{\arraystretch}{1.2} 
\begin{table*}[t]
  \centering
    %  \captionsetup{font=footnotesize}
     \caption{\textbf{Comparison of related work and our work on contributed algorithms with bounded dynamic regret bounds for Online Convex Optimization}. GO denotes 
     the number of gradient oracle calls per iteration of the respective algorithm.
    }
     \label{table:comparison}
     \resizebox{2\columnwidth}{!}{
    %  \Large{
     \begin{tabular}{llcccc}
     \toprule
	 Reference & Loss function & Projection-free & Memory & GO & Regret Rate \cr
	\midrule
	 \cite{zinkevich2003online} & Convex & No & No & $\calO(1)$ & $\calO(\sqrt{T}(1+C_T))$ \cr
 	 \cite{jadbabaie2015online} & Convex smooth & No & No & $\calO(1)$ & $\calO\left(\sqrt{(1+D_T)}+\min\left\{\sqrt{(1+D_T)C_T},(1+D_T)^{\frac{1}{3}}T^{\frac{1}{3}}V_T^{\frac{1}{3}}\right\}\right)$ \cr
	 \cite{mokhtari2016online} & Strongly convex & No & No & $\calO(1)$ & $\calO(1+C_T)$ \cr
	 \cite{yang2016tracking} & Convex smooth & No & No & $\calO(1)$ & $\calO(C_T)$ \cr
	 \cite{zhang2018adaptive} & Convex & No & No & $\calO(1)$ & $\calO(\sqrt{T(1+C_T)})$ \cr
	 \cite{kalhan2021dynamic} & Convex smooth & \textbf{Yes} & No & $\calO(1)$ & $\calO\left(\sqrt{T}(1+V_{T}+\sqrt{D_T}\right)$ \cr
	 Ours (\Cref{theorem:OFW_Memoryless} and \Cref{theorem:OFW_Memoryless_Full}) & Convex & \textbf{Yes} & No & $\calO(1)$ & $\calO\left(\sqrt{T}(1+V_{T}+\sqrt{D_T}\right)$, $\calO\left(\sqrt{T\left(V_{T}+D_{T}\right)}\right)$ \cr
	 \cite{zhao2022non} & Convex & No & \textbf{Yes} & \textbf{$\calO(1)$} & $\calO(\sqrt{T(1+C_T)})$ \cr
	 Ours (\Cref{theorem:MetaOFW}) & Convex & \textbf{Yes} & \textbf{Yes} & \textbf{$\calO(1)$} & $\calO(\sqrt{T(1+V_T+\bar{D}_T+C_T)})$ \cr
     \bottomrule
     \end{tabular}
    % }
         }
 	\vspace{-5mm}
\end{table*}

%% file: 1-RelatedWork.tex
% \input{Table/Table-comparison}

\section{Related Work}\label{sec:lit_review}

We review the literature by first reviewing \textit{OCO without Memory} and \textit{OCO with Memory}; then, we review \textit{Online Learning for Control via OCO with Memory}.
% , \ie algorithms that assume no stochastic model about the environments 
% but involve projection to update the control policy 
% and select control inputs based on past information only.
% \red{not sure if we need to discuss robust control here; I'd remove. the related work should discuss the minimal amount of work most relevant to this paper.  robust control sounds irrelevant :P (I'm just direct in providing comments here :))}and \textit{robust control} algorithms that assumes known stochastic models about the environments, typically a Gaussian model, and select inputs based on simulating the future system dynamics across a {lookahead} horizon.

\myParagraph{OCO without Memory}
The \textit{OCO without Memory} literature is vast~\citep{hazan2016introduction}.  We here focus on algorithms that guarantee bounded dynamic regret; a representative subset is presented in \Cref{table:comparison}.
% in \Cref{table:comparison}, with an emphasis on algorithms that guarantee dynamic regret bounds. 
% A fundamental OCO algorithm with bounded dynamic regret is the

\cite{zhang2018adaptive} prove that the optimal dynamic regret for OCO without Memory is $\Omega\left(\sqrt{T(1+C_{T})}\right)$, and provide an algorithm matching this bound.  The algorithm is based on \textit{Online Gradient Descent} (\OGD), which is a projection-based algorithm: at each time step $t$, \OGD chooses a decision $\myx_{t}$ by first computing an intermediate decision $\mathbf{x}_{t}^\prime=\myx_{t-1}-\eta\nabla f_{t-1}(\myx_{t-1})$ ---given the previous decision $\myx_{t-1}$, the gradient of the previously revealed loss $f_{t-1}(\myx_{t-1})$, and a step size $\eta
>0$--- and then projects $\mathbf{x}_{t}^\prime$ back to the feasible convex set $\calX$ to output the final decision $\mathbf{x}_{t}$.
% ; \green{\ie $\myx_{t}=\Pi_\calX(\mathbf{x}_{t}^\prime)$, where $\Pi_\calX(\cdot)$ denotes the projection operation. Do we need this? it seems we only use $\Pi$ here} 
This projection operation is often computationally expensive since it requires solving a quadratic program \citep{rockafellar1976monotone}.  
When the projection operation is indeed computationally expensive, the \textit{Online Frank-Wolfe} (\OFW) algorithm is employed as a projection-free alternative~\citep{frank1956algorithm,hazan2012projection}: \OFW seeks a feasible descent direction by solving the linear program  $\mathbf{x}_{t-1}^\prime=\arg \min _{\mathbf{x} \in \mathcal{X}}\left\langle \nabla f_{t-1}(\mathbf{x}_{t-1}), \mathbf{x}\right\rangle$ and then updating $\mathbf{x}_{t}=(1-\eta) \mathbf{x}_{t-1}+\eta \mathbf{x}_{t-1}^\prime$. \cite{kalhan2021dynamic} generalize the \OFW method to OCO without Memory to achieve a bounded dynamic regret and \OFW has been observed 20 times faster than \OGD~\citep{kalhan2021dynamic}.\footnote{Additional examples of works utilizing {\fontsize{8.2}{8.2}\selectfont\sf{\textbf{OFW}}} for OCO without Memory are: \cite{hazan2012projection,jaggi2013revisiting,garber2015faster,wan2021projection,kalhan2021dynamic,kretzu2021revisiting,garber2022new,wan2023improved}. Examples of works utilizing {\fontsize{8.2}{8.2}\selectfont\sf{\textbf{OGD}}} for OCO without Memory are: \cite{zinkevich2003online,jadbabaie2015online,mokhtari2016online,yang2016tracking,zhang2018adaptive,chang2020unconstrained}.}

\myParagraph{OCO with Memory}
% OCO-M captures how the history of past decisions affects the current loss functions
% by allowing the online learning loss functions to depend on both current and past decisions. capture the temporal effects in online learning problems 
% \citep{weinberger2002delayed,anava2015online}. 
% Adaptive online learning algorithms with memory have been shown to have better empirical performance than those without memory \citep{gramacy2002adaptive}, and to be effective in real-world applications \citep{nguyen2012adaptive}. 
% Specifically, in OCO-M with memory $m$, the loss function at each time step $t$ takes the form $f_t(\myx_{t-m},\dots,\myx_t) : \calX^{m+1}\mapsto\mathbb{R}$, \ie it depends on the both current and the last $m$ decisions, instead of only the current decision which is the case for memoryless OCO.  The dynamic regret becomes
% \begin{equation}\label{def:rel-work-dyn-reg-memory}
%     \DyReg = \sum_{t=1}^{T} f_t(\myx_{t-m}, \dots, \myx_{t}) - \sum_{t=1}^{T}f_t(\myv_{t-m}, \dots, \myv_{t}),
% \end{equation}
% where it is assumed that $\myx_{t-m}=\mathbf{0}$ for $t\leq m$.
% 
% 
\cite{zhao2022non} prove that the optimal dynamic regret for OCO-M is $\Omega(\sqrt{ T\left(1+C_T\right)})$, and provide an algorithm matches thing bound based on \OGD. Earlier works have provided static regret bounds for OCO-M, such as the bound $\calO(T^{2/3})$ by \cite{weinberger2002delayed}, and the bound $\calO(\sqrt{T})$ by \cite{anava2015online}. 
We provide the first projection-free algorithm for OCO-M that also guarantees bounded dynamic regret.
% against any comparators.

\myParagraph{Online Learning for Control via OCO-M} 
OCO-M has been recently applied to the control of linear dynamical systems in the presence of adversarial (non-stochastic) noise
% the development of online non-stochastic control
\citep{agarwal2019online,simchowitz2020improper,shalev2012online}.
% , \ie of controlling a linear dynamical system under adversarial (non-stochastic) noise and general convex loss functions. 
% Specifically, the non-stochastic control problem can be reduced to OCO-M via control policies that assign control a function of past noise acting on the dynamical system \citep{}. 
The noise is adversarial in the sense that it may adapt to the system's evolution.  Generally, the noise can evolve arbitrarily, subject to a given upper bound on its magnitude ---the upper bound ensures problem feasibility.
% , and tunes the algorithms’ response to the nevertheless unknown noise.  
Thus, no stochastic model is assumed regarding the noise's evolution, in contrast to classical control that typically assumes Gaussian noise~\citep{aastrom2012introduction}.  
% These algorithms make no assumptions about the environments~\citep{shalev2012online,hazan2016introduction}. Instead, they assume the noise induced by environments can evolve arbitrarily, even adversarially, only subject to a given upper bound on its magnitude. \green{I understand what you mean here, but it seems a bit weird to first say we make no assumptions and then say instead we assume ...} The upper bound ensures the problem is feasible and tunes the algorithms' response to the nevertheless unknown noise.   

The current OCO-M algorithms for control prescribe control policies by optimizing linear feedback control gains.
% based on past information only, \ie no future information. 
The algorithms rely on
% utilize 
projection-based methods such as \OGD, and  guarantee bounded static regret~\citep{agarwal2019online,hazan2020nonstochastic,simchowitz2020improper,li2021online}, adaptive regret~\citep{gradu2020adaptive,zhang2022adversarial}, or dynamic regret~\citep{zhao2022non}.
Specifically, the said OCO-M regret bounds are against optimal static feedback control gains with the exception of the bound by \cite{zhao2022non} which is against a class of optimal time-varying policies; however, the definition of this class requires an a priori knowledge of linear feedback control gains that ensure stability.  We provide a projection-free controller with memory and bounded dynamic regret against any optimal {time-varying} linear feedback control policy without the need to specify to the optimal policy any stabilizing feedback control gains.

%% file: 2-Problem.tex
\section{Problem Formulation}\label{sec:problem}

We formally define the problem of \textit{Online Convex Optimization with Memory} (OCO-M) (\Cref{prob:efficient_OCO_memory_dyreg}), along with 
% To this end, we present 
standard convexity (but non-smoothness) assumptions.
% ,  and the formal definition of \textit{dynamic regret} against any comparator sequence (\Cref{Def:DyReg}).

\begin{problem}[Online Convex Optimization with Memory (OCO-M)~\citep{weinberger2002delayed}]\label{prob:efficient_OCO_memory_dyreg}
There exist 2 players, an online optimizer and an adversary, who choose decisions sequentially over a time horizon $T$. At each time step $t=1,\ldots, T$, the online optimizer chooses a decision $\mathbf{x}_t$ from a convex set $\calX$; then, the adversary chooses a loss $f_t: \calX^{m+1} \mapsto \mathbb{R}$ to penalize the optimizer's most recent $m + 1$ decisions. Particularly, the adversary reveals $f_t$ to the optimizer and the optimizer computes its loss $f_t(\myx_{t-m},\dots,\myx_t)$, where $\myx_{t-m}$ is  $\mathbf{0}$ for $t\leq m$.
% which is revealed to the optimizer
% with which to penalize the learner's most recent $m + 1$ decisions, \ie 
% and the optimizer suffers a loss of $f_t(\myx_{t-m},\dots,\myx_t)$.  
The optimizer aims to minimize 
% its cumulative loss, \ie 
$\SumOneT f_t(\myx_{t-m},\dots,\myx_t)$.
\end{problem}

The challenge in solving OCO-M optimally, \ie in minimizing $\SumOneT f_t(\myx_{t-m},\dots,\myx_t)$, is that the optimizer gets to know $f_t$ only after $\myx_t$ has been chosen, instead of before.  

% \myParagraph{Objective}
% We aim to develop efficient, \ie projection-free, online learning algorithms in the framework of OCO-M, guaranteeing near-optimal performance in non-stationary environments. The performance's near-optimality may be captured by bounding the algorithms' suboptimality with respect to optimal time-varying decisions, \ie by bounding \emph{dynamic regret}.
% \red{why don't we repeat here what we write in the technological need? also, after describing the objective intuitively, you need to say that dynamic regret is defined formally as follows:... }
% \blue{we don't need to mention \cref{eq:DyReg_metric}?}
Despite the above challenge, our objective is to develop an {efficient} (projection-free) online algorithm for OCO-M that despite its efficiency  still enjoys sublinear dynamic regret.  
% Specifically, we consider the following definition of dynamic regret.
% , such that the dynamic regret in \Cref{Def:DyReg} against any sequence of comparators grows sublinearly in $T$ up to multiplicative factors of the metrics defined in \cref{eq:DyReg_metric}.

% \begin{definition}[Dynamic Regret against Time-Varying Comparators \citep{zinkevich2003online}]\label{Def:DyReg}
% Assume that $(\myx_1, \dots, \myx_T) \in \calX^T$ is the sequence of decisions chosen by the optimizer, and that $(\myv_1, \dots, \myv_T) \in \calX^T$ is any sequence of alternative decisions (comparators) the optimizer could choose instead.  Then, \emph{dynamic regret} is the cumulative loss difference incurred by $(\myx_1, \dots, \myx_T)$ relative to $(\myv_1, \dots, \myv_T)$:\footnote{For notational simplicity, we suppress $\DyReg$'s dependence on $(\myx_1, \dots, \myx_T)$ and $(\myv_1, \dots, \myv_T)$.}
%     \begin{equation}
%         \DyReg \triangleq \SumOneT f_t(\myx_{t-m},\dots,\myx_t) - \SumOneT f_t(\myv_{t-m},\dots,\myv_t),
%         \label{eq:DyReg_def}
%     \end{equation}
% where $\myx_{t-m}$ and $\myv_{t-m}$ are assumed to be $\mathbf{0}$ for $t\leq m$.
% \end{definition}

% \green{the statement is similar to what we have said about the dynamic regret without memory (below eq. 2) }For example, $(\myv_1, \dots, \myv_T) \in \calX^T$ can be the minimizer of $\SumOneT f_t(\myx_{t-m},\dots,\myx_t)$, \ie the ideal sequence of decisions the optimizer would have chosen to solve OCO-M if the optimizer knew a priori $f_t$ for all $t=1,\ldots,T$.

To achieve our objective, we 
% define unary loss function and 
adopt standard assumptions
% \Cref{assumption:convexity} through \Cref{assumption:gradient} are standard 
in online convex optimization \citep{hazan2016introduction,anava2015online,agarwal2019online,simchowitz2020improper,zhang2018adaptive,gradu2020adaptive,zhao2022non}:

\begin{assumption}[Convex and Compact Bounded Domain, Containing the Origin]\label{assumption:bounded_set}
The domain set $\mathcal{X}$ is convex and compact,  contains the zero point, and has diameter $D$, where $D$ is a given non-negative number; \ie $\mathbf{0} \in \calX$, and $\|\mathbf{x}-\mathbf{y}\|_2\, \leq D$ for all $\mathbf{x} \in \mathcal{X}, \mathbf{y} \in \mathcal{X}$.
\end{assumption}

\begin{definition}[Unary Loss Function]\label{Def:UnaryLoss}
    Given $f_t: \calX^{m+1} \mapsto \mathbb{R}$, the \emph{unary loss function} is the $\widetilde{f}_t(\myx) \triangleq f_t(\myx, \dots, \myx)$.
\end{definition}

\begin{assumption}[Convex Loss]\label{assumption:convexity}
    The loss function $f_t: \calX^{m+1} \mapsto \mathbb{R}$ is convex, \ie the unary loss function $\widetilde{f}_t(\myx)$ is convex in $\myx$,  where $m$ is the memory length, and $\myx \in \calX$.
\end{assumption}

\begin{assumption}[Bounded Loss]\label{assumption:bounded_func}
The loss function $f_t$ takes values in $[a, a+c]$,  where $a$ and $c$ are non-negative; \ie
\begin{equation*}
    0 \leq a \leq f_t\left(\mathbf{x}_0, \ldots, \mathbf{x}_m\right) \leq a+c,
\end{equation*}
for all $\left(\mathbf{x}_0, \ldots, \mathbf{x}_m\right) \in \calX^{m+1}$  and $t \in \TimeSeq$.
\end{assumption}

\begin{assumption}[Coordinate-Wise Lipschitz]\label{assumption:Lipschitzness}
	The loss function $f_t$ is coordinate-wise $L$-Lipschitz, where $L$ is a given non-negative number; \ie 
	\begin{equation*}
	    \left|f_t\left(\mathbf{x}_0, \ldots, \mathbf{x}_m\right)-f_t\left(\mathbf{y}_0, \ldots, \mathbf{y}_m\right)\right|\leq L \sum_{i=0}^m\left\|\mathbf{x}_i-\mathbf{y}_i\right\|_2,
	\end{equation*}
    for all $\left(\mathbf{x}_0, \ldots, \mathbf{x}_m\right) \in \calX^{m+1}$, and $\left(\mathbf{y}_0, \ldots, \mathbf{y}_m\right) \in \calX^{m+1}$, and for all $t \in \TimeSeq$.
\end{assumption}

\begin{assumption}[Bounded Gradient]\label{assumption:gradient}
	The gradient norm of $\widetilde{f}_t$ is at most $G$, where $G$ is a given non-negative number; \ie $\left\|\nabla \widetilde{f}_t(\mathbf{x})\right\|_2 \leq G$  for all $\mathbf{x} \in \mathcal{X}$ and $t \in \TimeSeq$.
\end{assumption}

% All in all, the loss function $f_t$ is assumed convex with bounded value and gradient, and is not necessarily smooth. 

%% file: 3-Meta-OFW-Alg.tex
\section{{\fontsize{11}{8.7}\selectfont\sf{\textbf{Meta-OFW}}} Algorithm for OCO-M}\label{sec:Meta-OFW-Alg}
We present \MetaOFW, the first projection-free
algorithm with bounded dynamic regret for OCO-M.~\MetaOFW leverages as subroutine the Online Frank-Wolfe (\OFW) algorithm.  \OFW is introduced by \cite{kalhan2021dynamic} for the OCO problem without memory. 
% \OFW was proposed for the standard OCO problem with no memory. 

We next first present the \OFW algorithm
% with the novel dynamic regret guarantees in 
(\Cref{subsec:OFW-Memoryless}), and then present the \MetaOFW algorithm (\Cref{subsec:Meta-OFW}). 
% We also present the motivation of using a meta-base algorithm in \Cref{subsec:OFW_Known_VT} due to the challenge of unknown $V_T$ when applying \OFW to OCO-M.

\subsection{The Online Frank-Wolfe (\OFW) Algorithm for OCO without Memory
% and Improved Dynamic Regret Bounds
}\label{subsec:OFW-Memoryless}

We present the \OFW algorithm (\Cref{alg:OFW_Memoryless}) along with its dynamic regret analysis (\Cref{theorem:OFW_Memoryless}).  Particularly, our analysis results in the same regret bound as \OFW's state of the art bound in \citep[Theorem~1]{kalhan2021dynamic} but under \Cref{assumption:bounded_set} and \Cref{assumption:convexity} only.
% , and against any comparator sequence in the evaluation of \OFW's  regret. 
Instead, \OFW's bound in \cite{kalhan2021dynamic} holds true under the additional assumption of smooth loss functions.
% , instead of only convex.
% , and only against the comparators that minimize in hindsight the cumulative loss $\SumOneT f_t(\myx_t)$.

\begin{theorem}[Dynamic Regret Bound of \OFW for OCO with no memory]\label{theorem:OFW_Memoryless}
Consider the OCO problem with no memory, \ie \Cref{prob:efficient_OCO_memory_dyreg} with $m=0$. Under \Cref{assumption:bounded_set} and \Cref{assumption:convexity},  \OFW achieves against any sequence of comparators $(\mathbf{v}_1, \dots, \mathbf{v}_T)\in \calX^T$ the dynamic regret
\begin{equation}
    \DyReg \leq \calO\left(\frac{1+V_{T}}{\eta} + \sqrt{T D_{T}}\right).
    \label{eq:theorem_OFW_Memoryless_bound_init}
\end{equation}
\end{theorem}
% \Cref{eq:theorem_OFW_Memoryless_bound_init} depends on loss variation $V_T$ and gradient variation $D_T$. The first term on $V_T$ can be optimized by using step size $\eta=\calO\left(\frac{1}{\sqrt{T}}\right)$, resulting in \cref{eq:theorem_OFW_Memoryless_bound_1}.

Particularly, when $\eta$ is chosen such that $\eta=\calO\left(\frac{1}{\sqrt{T}}\right)$, then
\begin{equation}
    \DyReg \leq \calO\left(\sqrt{T}\left(1+V_{T}+\sqrt{D_{T}}\right)\right).
  	\label{eq:theorem_OFW_Memoryless_bound_1}
\end{equation}
The \OFW algorithm achieves \Cref{theorem:OFW_Memoryless} by executing the following projection-free steps  (\Cref{alg:OFW_Memoryless}): \OFW first takes as input the time horizon $T$ and a constant step size $\eta$.
% , and outputs $\myx_t$ at each iteration $t=1,\ldots, T$. 
% \red{initialize?}\blue{$\myx_1$ is the initial decision}
Then, at each iteration $t=1,\ldots, T$, \OFW chooses an $\myx_t$, after which the learner suffers a loss $f_t(\myx_t)$ and evaluates the gradient $\nabla f_t(\mathbf{x}_t)$ (lines 3-4). Afterwards, \OFW seeks a direction $\myx_t^\prime$
% , \ie \textit{conditional gradient}, 
that is parallel to the gradient within the feasible set $\calX$ by solving a linear program only once per iteration (line 5). Finally, the decision for next iteration is then updated by $\mathbf{x}_{t+1}=(1-\eta) \mathbf{x}_{t}+\eta \mathbf{x}_{t}^\prime$ (line 6).

\begin{remark}[Efficiency due to only Projection-Free Operations]
\OFW in \Cref{alg:OFW_Memoryless} is projection-free:
% , in contrast to \OGD \citep{zinkevich2003online} which requires a projection operation. 
it finds a descent direction within the feasible set via solving a linear program \emph{once} per iteration (line 5).  Instead, \eg~\OGD requires solving a quadratic program for projections \citep{zinkevich2003online}. Thus, \OFW is more efficient when projections are costly.   For example, \cite{kalhan2021dynamic} demonstrates that \OFW is 20 times faster than \OGD in matrix completion scenarios. 
In the numerical evaluations in this paper (\Cref{app:sec_control_exp}), over online non-stochastic control scenarios, we observe that the proposed \OFW-based algorithm is about 3 times faster than the \OGD-based algorithm (achieving comparable or superior loss performance).
% For example, \cite{kalhan2021dynamic} demonstrates that \OFW is 20 times faster than \OGD in applications of matrix completion. 
\end{remark}

% \subsubsection{Dynamics Regret Analysis of \OFW Algorithm} 
% We analyze for the first time the dynamic regret of \Cref{alg:OFW_Memoryless} against any sequence of comparators $\myv_1, \dots, \myv_T \in \calX$, under \Cref{assumption:convexity} to \Cref{assumption:bounded_func}. In particular, by selecting proper step size, we demonstrate in \Cref{theorem:OFW_Memoryless} that the dynamic regret upper bound of \Cref{alg:OFW_Memoryless} grows in $\sqrt{T}$ with multiplicative factors of $V_T$ (for diminishing learning rate) or $\sqrt{V_T}$ (for constant learning rate) and $\sqrt{D_T}$ defined in \cref{eq:DyReg_metric}; therefore, \Cref{theorem:OFW_Memoryless} establishes convergence of \Cref{alg:OFW_Memoryless} for non-stationary learning problems. 

\input{Alg/Alg-OFW.tex}

\subsection{\MetaOFW Algorithm for OCO-M}\label{subsec:Meta-OFW}
\input{Alg/Alg-Meta-OFW.tex}
We present \MetaOFW (\Cref{alg:Meta_OFW}).  To this end, we start with the intuition on how \Cref{alg:Meta_OFW}'s steps achieve a bounded dynamic regret (the
% , the first projection-free algorithm for \Cref{prob:efficient_OCO_memory_dyreg} with provable dynamic regret bound (\Cref{theorem:MetaOFW}) ---
 rigorous dynamic regret analysis of \MetaOFW is given in \Cref{sec:Meta-OFW-Analysis}).  

\Cref{alg:Meta_OFW} utilizes multiple copies of the \OFW algorithm as base-learners ---each one with a different step size $\eta$---
and the \Hedge algorithm \citep{freund1997decision} as a meta-learner.  The  multiple copies of \OFW aim to cope with the a priori unknown loss variation $V_T$ via a trick reminiscent of the {``doubling trick''~\citep{shalev2012online}}, \ie via covering the spectrum of step sizes such that there exist a step size that approximately minimizes \cref{eq:theorem_OFW_Memoryless_bound_init} as if $V_T$ was known;
% \green{maybe we can only say there exists an $\eta$ in the spectrum that is close enough to the minimizing step size?}; 
and \Hedge fuses the decisions provided by the base-learners to output a final decision $\myx_t$ at each step~$t$.  

We discuss in more detail the role of the base- and meta-learners in  \Cref{rem:known_V_T} and \Cref{rem:Hedge} below, respectively.  To this end, we use the following notation and definitions:
\vspace{-2mm}
\begin{itemize}[leftmargin=9pt]\setlength\itemsep{-1.mm}
    \item $\lambda\triangleq m^{2} L$ is a regularizing constant;
    \item $N$ is the total number of the base-learners;
    % \item $\sum_{t=1}^T \widetilde{f}_t\left(\mathbf{x}_t\right)-\sum_{t=1}^T \widetilde{f}_t\left(\mathbf{v}_t\right)$ is the unary cost;
    
    % \item $\sum_{t=2}^T\left\|\mathbf{x}_t-\mathbf{x}_{t-1}\right\|_2$ is the switching cost, \ie cumulative movement of decisions;
    
    \item $\calB_i$ is the $i$-th base-learner running \OFW with step size $\eta_i$ and output $\myx_{t,i}$ at each iteration $t$, where $i \in \{1, \dots, N\}$;
    
    \item $g_{t}(\mathbf{x})\triangleq\left\langle\nabla \widetilde{f}_{t}\left(\mathbf{x}_{t}\right), \mathbf{x}\right\rangle$ is the linearized loss of the unary loss function $\widetilde{f}_t(\myx_t)$ over which each base-learner optimizes via the \OFW algorithm;
    % Optimizing $g_t(\cdot)$ helps regularize the unary cost since the convexity of $\widetilde{f}_{t}$ implies $\widetilde{f}_{t}(\myx_t) - \widetilde{f}_{t}(\myv_t) \leq \left<\nabla \widetilde{f}_{t}(\mathbf{x}_{t}), \myx_t-\myv_t \right>$;
    
    \item ${\ell}_{t,i}~\triangleq~g_{t}(\mathbf{x}_{t,i}) + \lambda \left\| \mathbf{x}_{t,i} - \mathbf{x}_{t-1,i} \right\|_2$ is a surrogate loss associated with the $i$-th base-learner $\calB_i$  ---the meta-learner collects ${\ell}_{t,i}$ for all base-learners, \ie for all $i \in\{1,\ldots, N\}$, and optimizes $\myx_t$ via \Hedge; 
    % The second term in $\ell_{t,i}$ helps regularize the switching cost;
    
    \item ${p}_{t,i}$ is the assigned weight to the $i$-th base-learner $\calB_i$ by \Hedge~---each ${p}_{t,i}$, $i \in\{1,\ldots, N\}$, is used to output \MetaOFW's final decision $\myx_t$ as the weighted sum of base-learners' decisions $\mathbf{x}_{t,i}$; \ie $\myx_t = \sum_{i=1}^{N}p_{t,i}\mathbf{x}_{t,i}$;
    
    \item $\boldsymbol{\ell}_{t} \in \mathbb{R}^N$ is the vector whose $i$-th entry is $\ell_{t,i}$;
    \item $\boldsymbol{p}_{t}$ is the vector with $i$-th entry as $p_{t,i}$;
    \item $\alpha\triangleq{2}(a+c)$ is a constant introduced for notational simplicity ($a$ and $c$ are per \Cref{assumption:bounded_func}).
\end{itemize}
\vspace{-2mm}

\begin{remark}[Unknown Loss Variation $V_T$ Requires Multiple \OFW Base-Learners]\label{rem:known_V_T}
The multiple \OFW base-learners aim to overcome the challenge of the a priori unknown loss variation $V_T$. To illustrate this, we first consider that $V_T$ is \textit{known} a priori, and show that a single \OFW suffices to achieve bounded dynamic regret for OCO-M. Then, we consider that $V_T$ is \textit{unknown} a priori, and show how multiple base-learners with appropriate step sizes $\eta$ can approximate the case where $V_T$ is known a priori. To these ends, we leverage the following dynamic regret bound for OCO-M \citep[Proof of Theorem~3.1]{anava2015online}:
\begin{equation}
	\begin{aligned}
		\DyReg \leq & \underbrace{ \sum_{t=1}^T \widetilde{f}_t\left(\mathbf{x}_t\right)-\sum_{t=1}^T \widetilde{f}_t\left(\mathbf{v}_t\right)}_{\text{unary cost}} \\
		& + \lambda
		\underbrace{ \sum_{t=2}^T\left\|\mathbf{x}_t-\mathbf{x}_{t-1}\right\|_2}_{\text{switching cost}} + \lambda 
		\underbrace{
% 		\lambda C_T.
		 \sum_{t=2}^T\left\|\mathbf{v}_t-\mathbf{v}_{t-1}\right\|_2 }_{\text{path length}},
	\end{aligned}
	\label{eq:DyReg_decomposition}
\end{equation}
which we can simplify to
\begin{equation}
		\DyReg \leq  \mathcal{O}\left(\sqrt{ T\left(1+{V}_T+D_T+C_T\right)}\right),
	\label{eq:DyReg_decomposition_aux}
\end{equation}
when $V_T$ is known a priori.  Particularly, assume that  $\myx_t$ is updated by an \OFW algorithm applied to $\widetilde{f}_1, \ldots, \widetilde{f}_T$ with the $V_T$-dependent step size $\eta_* = \calO \left( \sqrt{(1+{V}_T)/T} \right)$.  Then, \cref{eq:DyReg_decomposition_aux} results from \cref{eq:DyReg_decomposition} since the three terms in \cref{eq:DyReg_decomposition} can be bounded respectively as follows: (i) the unary cost can be bounded by  \cref{eq:theorem_OFW_Memoryless_bound_init} where $\eta=\eta_*$; (ii)  the switching cost can be bounded by $\eta_*TD$ due to \OFW's line 6 and due to \Cref{assumption:bounded_set}; and (iii) the path length is by definition equal to $C_T$.  Then, an application of the Cauchy-Schwarz inequality completes the proof of \cref{eq:DyReg_decomposition_aux}.  All in all, when $V_T$ is known a priori, a single \OFW suffices to achieve bounded dynamic regret for OCO-M.

But $V_T$ is unknown a priori since it depends on the loss functions, which are unknown a priori.  Instead, an upper bound to $V_T$ is known, specifically, it holds true that $V_T\leq Tc$ under \Cref{assumption:bounded_func}. Leveraging this, we can approximate the case where $V_T$ is known a priori by employing an appropriate number of \OFW base-learners, each with a different step size, per \cref{eq:base_learners_N} and \cref{eq:step_size_pool} below.   Intuitively, we can guarantee that way that
% However, this step-size tuning is infeasible in practice since the loss variation is typically revealed to the learner a posteriori. Particularly, 
% loss variation captures the non-stationarity of the environment and it is a priori unknown and unpredictable to the learner. Therefore, due to unknown $V_T$, we need a meta-base algorithm running multiple \OFW with different step sizes, such that 
there exists a base-learner $i$ with step size $\eta_{i}$ close to the unknown step size~$\eta_*$ (the full justification of \cref{eq:base_learners_N} and \cref{eq:step_size_pool} is given in \Cref{theorem:MetaOFW}'s proof in \Cref{app-subsec:theorem_MetaOFW}).  The challenge now is to fuse the decisions of the multiple \OFW base-learners to a final decision $\myx_t$.
% such that the bound in \cref{eq:DyReg_decomposition_aux} still holds true.
\end{remark}

% The challenge now is how to fuse the decisions suggested by the multiple \OFW to a final decision $\myx_t$ that results to the same regret bound as in \cref{eq:DyReg_decomposition_aux}.

\begin{remark}[The Multiple \OFW Require a \Hedge Meta- Learner]\label{rem:Hedge}
The \Hedge meta-learner in \MetaOFW aims to fuse the decisions of the multiple \OFW base-learners to a final decision $\myx_t$.
% such that \cref{eq:DyReg_decomposition_aux} holds true even if $V_T$ is unknown a priori.  
Specifically, the \OFW base-learners provide multiple decisions at each iteration, the $\myx_{t,i}$, $i\in\{1,\ldots,N\}$ (line 4 in \Cref{alg:Meta_OFW}). Then, \MetaOFW utilizes the \Hedge steps in lines 5, 9, and 10 to fuse those decisions to a single decision, aiming to ``track'' the best base-learner $\calB_i$.
% and realize the regret bound in \cref{eq:DyReg_decomposition_aux}; we prove that this is indeed the case in \Cref{sec:Meta-OFW-Analysis} (\Cref{theorem:MetaOFW}). 
\end{remark}

We next formally describe \MetaOFW. First, the algorithm specifies the number of base learners, their corresponding step sizes, and their initial weights as follows, respectively:
\begin{align}
\hspace{-3mm}	&N=\left\lceil\frac{1}{2} \log _2(1+\frac{Tc}{\alpha})\right\rceil+1 = \mathcal{O}(\log T), \label{eq:base_learners_N} \\ 
\hspace{-3mm}	&\mathcal{H}=\left\{\eta_i \mid \eta_i=2^{i-1} \sqrt{\frac{\alpha}{\lambda TD}} \leq 1, i \in \{1, \dots, N\}\right\}, \label{eq:step_size_pool} \\ 
\hspace{-3mm}	&p_{1, i}=\frac{1}{i(i+1)} \cdot \frac{N+1}{N}, \text{ for any } i \in \{1, \dots, N\}.\label{eq:initial_weight}
\end{align}

Also, \MetaOFW sets the meta-learner's learning rate per the following  \cref{eq:theorem_MetaOFW_meat_step_size}:
        \begin{equation}
        	\epsilon=\sqrt{2 /\left((2 \lambda+G)(\lambda+G) D^2 T\right)},
        	\label{eq:theorem_MetaOFW_meat_step_size}
        \end{equation}
where the dependence on $T$ can be removed by a ``doubling trick''  \citep{cesa1997use}, similarly to how \MetaOFW copes with the unknown $V_T$, 

At each iteration $t=1, \ldots, T$, \MetaOFW  receives the intermediate decisions $\myx_{t,i}$ from all the base-learners $\calB_i$, $i\in\{1,\ldots,N\}$ (line 4) to fuse them into a final decision $\myx_t = \sum_{i=1}^{N}p_{t,i}\mathbf{x}_{t,i}$ (line 5).  Then, \MetaOFW suffers a loss of $f_t(\myx_{t-m},\dots,\myx_t)$ (lines 6-7). Afterwards, \MetaOFW constructs the linearized loss $g_t(\myx)$ and switching-cost-regularized loss $\boldsymbol{\ell}_t$ (lines 8-9). To this end, \MetaOFW  needs to evaluate only once the gradient $\nabla \widetilde{f}_{t}(\mathbf{x}_{t})$. Finally, the meta-learner and base-learners update the weights $\boldsymbol{p}_{t+1}$ and $\myx_{t+1,i}$ for the next iteration (lines 10-11).

%% file: Alg/Alg-OFW.tex
%\setlength{\textfloatsep}{-0.5mm}
% \setlength{\aboveskip}{0mm}
\begin{algorithm}[t]
	\caption{\mbox{\hspace{-.05mm}Online Frank-Wolfe Algorithm (\OFW)} \citep{kalhan2021dynamic}.}
	\begin{algorithmic}[1]
		\REQUIRE Time horizon $T$; step size $\eta$.
		\ENSURE Decision $\mathbf{x}_t$ at each time step $t=1,\ldots,T$.
		\medskip
		\STATE Initialize $\mathbf{x}_1 \in \calX$; 
		\FOR {each time step $t = 1, \dots, T$}
		\STATE Suffer a loss $f_t(\mathbf{x}_t)$;
		\STATE Obtain gradient $\nabla f_t(\mathbf{x}_t)$;
		\STATE Compute $\mathbf{x}_{t}^\prime=\arg \min _{\mathbf{x} \in \mathcal{X}}\left\langle \nabla f_t(\mathbf{x}_t), \mathbf{x}\right\rangle$;
		\STATE Update $\mathbf{x}_{t+1}=(1-\eta) \mathbf{x}_{t}+\eta \mathbf{x}_{t}^\prime$;
		\ENDFOR
	\end{algorithmic}\label{alg:OFW_Memoryless}
\end{algorithm}

%% file: Alg/Alg-Meta-OFW.tex
\begin{algorithm}[t]
	\caption{\mbox{\hspace{-.05mm}Meta OFW Algorithm (\MetaOFW).}}
	\begin{algorithmic}[1]
		\REQUIRE Time horizon $T$; number of base-learners $N$ per \cref{eq:base_learners_N}; step-size pool $\calH$ per \cref{eq:step_size_pool}; initial weight of base-learners $\boldsymbol{p}_{1}$ per \cref{eq:initial_weight}; learning rate $\epsilon$ for meta-algorithm per \cref{eq:theorem_MetaOFW_meat_step_size}.
		\ENSURE Decision $\mathbf{x}_t$ at each time step $t=1,\ldots,T$.
		\medskip
		\STATE Set $\mathbf{x}_\tau = \mathbf{0}$, $\forall \tau \leq 0$;
		\STATE Initialize $\mathbf{x}_{1,i} \in \calX$, $\forall i \in \{1, \dots, N\}$; 
		\FOR {each time step $t = 1, \dots, T$}
		\STATE Receive $\mathbf{x}_{t,i}$ from base-learner $\calB_i$ for all $i$;
		\STATE Output the Decision $\mathbf{x}_t = \sum_{i=1}^{N}p_{t,i}\mathbf{x}_{t,i}$;
		\STATE Suffer loss $f_t(\mathbf{x}_{t-m}, \dots, \mathbf{x}_t)$;
		\STATE Observe the loss function $f_t: \ \calX^{m+1} \mapsto \mathbb{R}$;
		\STATE Construct linearized loss
			\vspace{-2mm}
			 $$g_{t}(\mathbf{x})=\left\langle\nabla \widetilde{f}_{t}\left(\mathbf{x}_{t}\right), \mathbf{x}\right\rangle; $$
			 \vspace{-6mm}
		\STATE Construct the switching-cost-regularized surrogate loss  $\boldsymbol{\ell}_t~\in~\mathbb{R}^N$ with
			\vspace{-3mm}
 			$${\ell}_{t,i}~=~g_{t}(\mathbf{x}_{t,i}) + \lambda \left\| \mathbf{x}_{t,i} - \mathbf{x}_{t-1,i} \right\|_2; $$
			 \vspace{-6mm} 			
		\STATE Update the weight of base-learners $\boldsymbol{p}_{t+1} \in \Delta_{N}$ by
			 \vspace{-3mm}
			$$p_{t+1,i} = \frac{p_{t,i} e^{-\epsilon \ell_{t,i}}}{\sum_{j=1}^{N}p_{t,j} e^{-\epsilon \ell_{t,j}}}; $$
			 \vspace{-5mm}
		\STATE Base-learner $\calB_i$ updates $\mathbf{x}_{t+1,i}$ with step size $\eta_{i}$ for all $i$, per \OFW in \Cref{alg:OFW_Memoryless};
		\ENDFOR
	\end{algorithmic}\label{alg:Meta_OFW}
\end{algorithm}

%% file: 4-Meta-OFW-Analysis.tex
\section{Dynamic Regret Guarantees of {\fontsize{11}{8.7}\selectfont\sf{\textbf{Meta-OFW}}}}\label{sec:Meta-OFW-Analysis}

To present \MetaOFW's dynamic regret bound, we define:
\begin{itemize}[leftmargin=9pt]\setlength\itemsep{-1.mm}
    \item $D_{T,i} \triangleq \sum_{t=1}^{T}\left\|\nabla f_{t}\left(\mathbf{x}_{t,i}\right)-\nabla f_{t-1}\left(\mathbf{x}_{t-1,i}\right)\right\|_2^{2}$ is the gradient variation associated with the base-learner $i$;
    \item $\bar{D}_T \triangleq  \max_{i \in \{1, \dots, N\}} D_{T,i}$ is the upper bound for $D_{T,i}$.
    % \item $\bar{D}_T \triangleq \sum_{t=1}^{T} \sup_{\mathbf{y},\mathbf{z} \in \mathcal{X}} \left\|\nabla f_{t}\left(\mathbf{y}\right)-\nabla f_{t-1}\left(\mathbf{z}\right)\right\|_2^{2}$ is the upper bound for $D_{T,i}$.
\end{itemize}

We present \MetaOFW's dynamic regret bound against any comparator sequence (\Cref{theorem:MetaOFW}). Particularly, the bound below holds true, even if the loss variation $V_T$, gradient va- riation $\bar{D}_T$, and path length $C_T$ are unknown to \MetaOFW. 

\begin{theorem}[Dynamic Regret Bound of \MetaOFW]\label{theorem:MetaOFW}
For any comparator sequence $(\mathbf{v}_1, \ldots, \mathbf{v}_T) \in \mathcal{X}^T$, \MetaOFW achieves a dynamic regret $\DyReg$ that enjoys the bound:
	\begin{equation}
		\begin{aligned}
			\DyReg  & \leq \mathcal{O}\left(\sqrt{ T\left(1+V_T+\bar{D}_T+C_T\right)} \right).
		\end{aligned}
		\label{eq:theorem_MetaOFW_bound}
	\end{equation}	
\end{theorem}

The dependency on $V_T$ and $\bar{D}_T$ results from \OFW being a base-learner in \MetaOFW; similar dependencies, due to projection-free subroutines in online algorithms, have been observed in the literature: see, \eg \cite{kalhan2021dynamic} and the references in \Cref{table:comparison}.  
The dependency on $\bar{D}_T$, instead of $D_T$ in \Cref{theorem:OFW_Memoryless}, is to upper bound the gradient variation $D_{T,i}$ such that the base-learner $i$ with step size close to the unknown step size $\eta^\star$ (\Cref{rem:known_V_T}) satisfies $D_{T,i}\leq \bar{D}_T$.
% $V_T$ and $\bar{D}_T$ can be sublinear in decision-making applications where the loss functions change slowly (\eg in classic linear quadratic control problems where the loss functions are time-\underline{in}variant.).

The dependency on $C_T$ results from the sequence of comparators being time-varying. 
Specifically, \cite{zhang2018adaptive} proved that any optimal dynamic regret bound  for OCO is $\Omega\left(\sqrt{T(1+C_T)}\right)$, and thus the bound necessarily depends on $C_T$ in the worst case. 

\begin{remark}[Trade-Off of Projection-Free Efficiency with Regret Optimality] \cite{zhao2022non} prove that the optimal dynamic regret for OCO-M is $\Omega(\sqrt{ T\left(1+C_T\right)})$, and provide a projection-based algorithm  using \OGD that matches this bound.  In contrast, \MetaOFW's  regret bound in \Cref{theorem:MetaOFW} cannot match the bound $\Omega(\sqrt{ T\left(1+C_T\right)})$ due to the presence of $V_T$ and $\bar{D}_T$ in \cref{eq:theorem_MetaOFW_bound}.  But \MetaOFW is projection-free and thus is more efficient than the \OGD-based algorithm in \cite{zhao2022non} \citep{hazan2012projection}.  All in all, the dependence of \cref{eq:theorem_MetaOFW_bound} on $V_T$ and $\bar{D}_T$ is the regret suboptimality cost we pay in this paper to solve OCO-M efficiently via the projection-free \OFW.
% , \ie it is the price we pay to  use \OFW and be more efficient than using \OGD .
\end{remark}

%% file: 5-Control.tex
\section{Application to Non-Stochastic Control}\label{sec:control}

We apply \MetaOFW to the online non-stochastic control problem \citep{agarwal2019online}, and present a projection-free controller with memory (\Cref{alg:Meta_OFW_Control}), and with bounded dynamic regret against any linear time-varying feedback control policy (\Cref{theorem:MetaOFW_Control}). The results of the numerical evaluations are present in \Cref{app:sec_control_exp}. 
% The resulting algorithm enjoys a dynamic regret bound $\tilde{\mathcal{O}}\left(\sqrt{ T\left(1+V_T+D_T+C_T\right)} \right)$, where $\tilde{\calO}$ hides logarithmic terms (\Cref{theorem:MetaOFW_Control}). 

% We formally present the non-stochastic problem next. 
% In particular, we present \MetaOFW for online non-stochastic control in \Cref{alg:Meta_OFW_Control} with regret bound . We first formulate the control problem.

\subsection{The Non-Stochastic Control Problem}\label{subsec:control_formulation}
We consider Linear Time-Varying systems of the form
\begin{equation}
	x_{t+1} = A_{t} x_{t} + B_{t} u_{t} + w_{t}, \quad t=0, \ldots, T,
	\label{eq_LTV}
\end{equation}
where $x_t \in\mathbb{R}^{d_x}$ is the state of the system, $u_t \in\mathbb{R}^{d_u}$ is the control input, and $w_t\in\mathbb{R}^{d_x}$ is the process noise.  The system and input matrices, $A_t$ and $B_t$, respectively, are known.

At each time step $t$, the controller chooses a control action $u_t$ and then suffers a loss $c_t(x_t,u_t)$.  The loss function $c_t$ is revealed to the controller only after the controller has chosen the control action $u_t$, similarly to the OCO setting.

\begin{assumption}[Convex and Bounded Loss Function with Bounded Gradient]\label{assumption:cost}
The cost function $c_{t}(x_t,u_t): \mathbb{R}^{d_{x}} \times \mathbb{R}^{d_{u}} \mapsto \mathbb{R}$ is convex in $x_t$ and $u_t$. Further, when $\|x\|_2\;\leq D$, $\|u\|_2\;\leq D$ for some $D>0$, then $\left|c_{t}(x, u)\right| \leq$ $\beta D^{2}$ and $\left\|\nabla_{x} c_{t}(x, u)\right\|_2\leq G_{c} D,\left\|\nabla_{u} c_{t}(x, u)\right\|_2 \leq G_{c} D$, for given positive numbers $\beta$ and $G_c$.
\end{assumption} 

\begin{assumption}[Bounded System Matrices and Noise]\label{assumption:bounded_system_noise}
The system matrices and noise are bounded, \ie $\|A_t\|_{\mathrm{op}}\; \leq \kappa_{A}$, $\|B_t\|_{\mathrm{op}}\; \leq \kappa_{B}$, and $\left\|w_t\right\|_2 \leq W$ for given positive numbers $\kappa_{A}$, $\kappa_{B}$, and $W$, where $\|\cdot\|_{\mathrm{op}}$ is the operator norm.
\end{assumption}

Per \Cref{assumption:bounded_system_noise}, we assume no stochastic model for the process noise $w_t$: the noise may even be adversarial, subject to the bounds prescribed by $W$.

% We consider dynamic policy regret for performance metric.

\begin{problem}[Non-Stochastic Control (NSC) Problem]\label{prob:NSC}
At each time step $t=0,\ldots, T$, first a control action $u_t$ is chosen;
% from a convex set and the LTV system evolves per \cref{eq_LTV}; 
then, a loss function $c_t : \mathbb{R}^{d_{x}} \times \mathbb{R}^{d_{u}} \mapsto \mathbb{R}$ is revealed and the system suffers a loss $c_t(x_t,u_t)$. The goal is to minimize the dynamic policy regret defined below.
\end{problem}

\begin{definition}[Dynamic Policy Regret]\label{def:DyReg_control}
We define the \emph{dynamic policy regret} as
\begin{equation}
	\DyRegNSC = \sum_{t=0}^{T} c_{t}\left(x_{t}, u_{t}\right)-\sum_{t=0}^{T} c_{t}\left(x_{t}^{*}, u_{t}^{*}\right),
	\label{eq:DyReg_control}
\end{equation}
where (i) both sums in \cref{eq:DyReg_control} are evaluated with the same noise $\{w_0,\ldots, w_T\}$,  which is the noise experienced by the system during its evolution per the control input $\{u_0,\ldots, u_T\}$, (ii) $u_{t}^{*} = -K_t^{*} x_{t}^{*}$ is the optimal linear feedback control input in hindsight, {\ie the optimal input given a priori knowledge of $c_t$ and of the realized noise $w_t$}, and (iii) $x_{t}^{*}$ is the state reached by applying the sequence of optimal control inputs $\{u_{0}^{*}, \dots, u_{t-1}^{*}\}$.
% , and $K_t^{*}$ is $(\kappa,\gamma)$-strongly stable defined in \Cref{def:kappa_gamma_stable}.
\end{definition}

\input{Alg/Alg-Meta-OFW-Control}

\myParagraph{Reduction to OCO-M}
We present the reduction of the non-stochastic control problem to OCO-M, following \cite{agarwal2019online}.

Per \cref{eq_LTV}, $x_t$ depends on the control actions chosen in the past, \ie $\{u_0,\dots, u_{t-1}\}$, and similarly, the control action $u_t$ depends on $x_{t-1}$, \ie $\{u_0,\dots, u_{t-2}\}$. To reduce the non-stochastic control problem to OCO-M, there are thus 2 challenges: (i) we need a control parameterization such that the cost function $c_t(x_t, u_t)$ is convex in the parameters of the control actions$\{u_0,\dots, u_{t-1}\}$, since $c_t(x_t, u_t)$ is implicitly a function of $\{u_0,\dots, u_{t-1}\}$ via $u_t$; and, similarly, (ii) we need the memory length of $c_t(x_t, u_t)$, \ie its implicit dependence on the past control inputs $\{u_0,\dots, u_{t-1}\}$,  to stop growing as $t$ increases; that is, we need $c_t(x_t, u_t)$ to instead depend on  the most recent control inputs only, in particular, on $\{u_{t-m},\dots, u_{t}\}$ for memory length $m$.
% despite the state transition of dynamical systems, being also implicitly a function of $u_0,\dots, u_{t-1}$. 
To address these challenges, \cite{agarwal2019online} propose the \textit{Disturbance-Action Control} policy and the notion of \textit{truncated loss}.

\begin{definition}[Disturbance-Action Control Policy]\label{def:DAC}
A \emph{Disturbance-Action Control (DAC)}  policy $\pi_t(K_t,M_t)$ chooses the control action $u_t$ at state $x_t$ as $u_t = -K_t x_t + \sum_{i=1}^{H}M_t^{[i-1]}w_{t-i}$,\footnote{The DAC policy depends on the past noise, which can be obtained from \cref{eq_LTV} once the next state is observed; specifically, at time $t+1$, it holds true that $w_t=x_{t+1}-A_tx_t-B_tu_t$.} where $M_t = (M_t^{[0]}, \dots, M_t^{[H-1]})$ with $\left\|M_t^{[i]}\right\|_{\mathrm{op}} \leq \kappa_B \kappa^3 (1-\gamma)^i$ and horizon $H \geq 1$, $K_t$ is a $(\kappa,\gamma)-$strongly stable matrix {which is calculated given $A_t$ and $B_t$}, and $w_\tau = 0$ for all $\tau < 0$. 
\end{definition}

Per \Cref{prop:state_transition} in \Cref{app-subsec:State_Trans_DAC} \citep{gradu2020adaptive}, $x_t$ and $u_t$ are linear in $\{M_0, \dots, M_t\}$; therefore, the cost function $c_t(x_t,u_t)$ is convex in $\{M_0, \dots, M_t\}$.

To present the notion of \textit{truncated loss}, we use the notation:
\begin{itemize}[leftmargin=9pt]\setlength\itemsep{-1.mm}
    \item $x_t\left(M_{0: t-1}\right)$ is the state reached by applying the DAC policy $\left\{\pi_\tau\left(K_\tau, M_{\tau}\right)\right\}_{\tau=0, \ldots, t-1}$; 
    \item $u_t\left(M_{0: t}\right)$ is the control action at state $x_t\left(M_{0: t-1}\right)$ per the DAC policy $\pi_t(K_t, M_t)$; 
    \item $y_{t}\left(M_{t-1-H: t-1}\right)$ is the state reached from $x_{t-1-H}={0}$ by applying $\left\{\pi_\tau\left(K_\tau, M_{\tau}\right)\right\}_{\tau=t-H-1, \ldots, t-1}$ and experiencing the noise sequence $\{w_\tau\}_{\tau=t-H-1, \ldots, t-1}$;
    \item $v_{t}\left(M_{t-1-H: t}\right)$ is the control input that would have been executed if the state  at time $t$ was the $y_{t}\left(M_{t-1-H: t-1}\right)$.
    % per the DAC policy $\pi_t(K_t, M_t)$.
\end{itemize} 

\begin{definition}[Truncated Loss]\label{def:truncated_loss}
% For the cost function $c_{t}: \mathbb{R}^{d_{x}} \times \mathbb{R}^{d_{u}} \mapsto \mathbb{R}$ and 
Given DAC policies $\left\{\pi_\tau\left(K_\tau, M_{\tau}\right)\right\}_{\tau=0, \ldots, t}$ with memory length $H$, the induced \emph{truncated loss} $f_{t}: \mathcal{M}^{H+2} \mapsto \mathbb{R}$ is defined as
\begin{equation*}
    f_{t}\left(M_{t-1-H: t}\right)\triangleq c_{t}\left(y_{t}\left(M_{t-1-H: t-1}\right), v_{t}\left(M_{t-1-H: t}\right)\right).
\end{equation*}
\end{definition} 

Thereby, the truncated loss  $f_{t}\left(M_{t-1-H: t}\right)$ depends only on the last $H+2$ time steps of the DAC policy.  That is, $f_{t}$ has a fixed memory length $H+2$, for all $t=1,\ldots,T$.

All in all, \Cref{prob:NSC} can be reduced to OCO-M when the decision variables are the $M_t$, and the loss functions are the truncated losses $f_{t}\left(M_{t-1-H: t}\right)$, for all $t=1,\ldots,T$.

\subsection{\MetaOFW for Online Non-Stochastic Control}\label{subsec:control_MetaOFW}
We present \MetaOFW's application to the online non-stochastic control problem (\Cref{alg:Meta_OFW_Control}). Particularly, \Cref{alg:Meta_OFW_Control} initializes the number of base-learners, their corresponding step sizes, and their initial weights, per the following equations, similarly to \MetaOFW:
\begin{align}
    &N=\left\lceil\frac{1}{2} \log_2(\frac{2\beta D^2 T + \phi}{\sigma})\right\rceil+1 = \mathcal{O}(\log T), \label{eq:base_learners_N_control} \\
    &\mathcal{H}=\left\{\eta_i \mid \eta_i=2^{i-1} \sqrt{\frac{\sigma}{\zeta T D_f}} \leq 1, i \in \{1, \dots, N\}\right\},	\label{eq:step_size_pool_control} \\
    &p_{0, i}=\frac{1}{i(i+1)} \cdot \frac{N+1}{N}, \text{ for any } i \in \{1, \dots, N\},\label{eq:initial_weight_control}
\end{align}
where $\sigma \triangleq 4\beta D^2$, $\phi \triangleq \sigma + 2 \beta D^2$, $\zeta \triangleq (H+2)^{2} L_{f}$, and $L_f$, $G_f$ defined as in \Cref{lemma:loss_Mspace_property} in \Cref{app-subsec:supporting_lemma}.

The algorithm also sets the step size of meta-learner as
\begin{equation}
	\epsilon=\sqrt{2 /\left((2 \zeta+G_f)(\zeta+G_f) D_f^2 T\right)}.
	\label{eq:theorem_MetaOFW_meat_step_size_control}
\end{equation}

At each iteration $t$, \Cref{alg:Meta_OFW_Control} receives $M_{t,i}$ from all base-learners (line 4). Then, \Cref{alg:Meta_OFW_Control} calculates $M_t = \sum_{i=1}^{N}p_{t,i} M_{t,i}$ and outputs the control actions $u_t = -K_t x_t + \sum_{i=1}^{H} M_t^{[i-1]} w_{t-i}$ (lines 5-6), after which the cost function is revealed and the algorithm  suffers a loss of $c_t(x_{t},u_t)$ (line 7). Next, \Cref{alg:Meta_OFW_Control} constructs the truncated loss $f_t(M_{t-H-1}, \dots, M_t)$, linearized loss $g_t(M)$, and switching-cost-regularized loss $\boldsymbol{\ell}_t$ (lines 8-10). The meta-learner and base-learners update the weights $\boldsymbol{p}_{t+1}$ and $M_{t+1,i}$ for the next iteration (lines 11-15). Finally, the noise $w_t$ is calculated when $x_{t+1}$ is observed (line 16).

\subsection{Dynamic Regret Guarantee of \Cref{alg:Meta_OFW_Control}}\label{subsec:control_regret}
\begin{theorem}[Dynamic Policy Regret Bound of \Cref{alg:Meta_OFW_Control}]\label{theorem:MetaOFW_Control}
\Cref{alg:Meta_OFW_Control} ensures that \footnote{The path length is defined as $C_T \triangleq \sum_{t=2}^{T}\left\|M_{t-1}^{*}-M_{t}^{*}\right\|_{\mathrm{F}}$.}
	\begin{equation}
		\begin{aligned}
			\DyRegNSC & \leq \tilde{\mathcal{O}}\left(\sqrt{ T\left(1+V_T+\bar{D}_T+C_T\right)} \right). 
		\end{aligned}  
		\label{eq:theorem_MetaOFW_bound_control}
	\end{equation}	
\end{theorem}

\begin{remark}[Novelty of \Cref{theorem:MetaOFW_Control}]
\Cref{theorem:MetaOFW_Control} guarantees a dynamic regret bound against an optimal \emph{time-varying linear feedback policy} in hindsight, \ie against $\{\pi_\tau(K_\tau^*,0)\}_{\tau=0,\dots,t}$, per \Cref{def:DyReg_control}. 
This is different than competing against an optimal time-varying DAC policy $\{\pi_\tau(K_\tau,M_\tau^*)\}_{\tau=0,\dots,t}$ {with pre-specified stabilizing control gains $K_\tau$} as in \cite{zhao2022non}, or an optimal time-\underline{in}variant linear feedback policy $\{\pi(K^*,0)\}$ over the entire horizon or any time interval as in \cite{agarwal2019online,li2021online,gradu2020adaptive,simchowitz2020improper,zhang2022adversarial}. 
To achieve this, we show a DAC policy $\{\pi_\tau(K_\tau,M_\tau)\}_{\tau=0,\dots,t}$ can approximate any time-varying linear feedback policy (\Cref{prop:DAC_sufficiency} in \Cref{app-subsec:DAC_sufficiency}) without the need to specify to the optimal policy any stabilizing feedback control gains.
\end{remark}

%% file: Alg/Alg-Meta-OFW-Control.tex
\begin{algorithm}[t]
	\caption{\mbox{\hspace{-.05mm}\MetaOFW for Non-Stochastic Control.}}
	\begin{algorithmic}[1]
		\REQUIRE Time horizon $T$; number of base-learners $N$ per \cref{eq:base_learners_N_control}; step size pool $\calH$ per \cref{eq:step_size_pool_control}; initial weight of base-learners $\boldsymbol{p}_{0}$ per \cref{eq:initial_weight_control}; learning rate $\epsilon$ of  meta-algorithm per \cref{eq:theorem_MetaOFW_meat_step_size_control}.
		\ENSURE Control $u_t$ at each time step $t=1,\ldots,T$.
		\medskip
		\STATE Set $M_\tau = \mathbf{0}$ and $w_\tau = 0$, $\forall \tau < 0$;
		\STATE Initialize $M_{0,i} \in \calM$, $\forall i \in \{1, \dots, N\}$; 
		\FOR {each time step $t = 0, \dots, T$}
		\STATE Receive $M_{t,i}$ from base-learner $\calB_i$ for all $i$;
		\STATE Calculate $M_t = \sum_{i=1}^{N}p_{t,i}M_{t,i}$;
		\STATE Output $u_t = -K_t x_t + \sum_{i=1}^{H} M_t^{[i-1]} w_{t-i}$;
		\STATE Observe the loss function $c_t: \mathbb{R}^{d_x} \times \mathbb{R}^{d_u} \mapsto \mathbb{R}$ and suffer the loss $c_t(x_t, u_t)$;
		\STATE Construct the truncated loss $f_t(M_{t-H-1}, \dots, M_t): \calM^{H+2} \mapsto \mathbb{R}$ ;
		\STATE Construct the linearized loss
			\vspace{-2mm}
			 $$g_{t}(M)=\left\langle\nabla_M \widetilde{f}_{t}\left(M_{t}\right), M\right\rangle_{\mathrm{F}}; $$
			 \vspace{-6mm}
		\STATE Construct the switching-cost-regularized surrogate loss  $\boldsymbol{\ell}_t~\in~\mathbb{R}^N$ with
			\vspace{-3mm}
 			$${\ell}_{t,i}~=~g_{t}(M_{t,i}) + \zeta \left\| M_{t,i} - M_{t-1,i} \right\|_{\mathrm{F}}; $$
			 \vspace{-6mm} 			
		\STATE Update the weight of base-learners $\boldsymbol{p}_{t+1} \in \Delta_{N}$ by
			 \vspace{-3mm}
			$$p_{t+1,i} = \frac{p_{t,i} e^{-\epsilon l_{t,i}}}{\sum_{j=1}^{N}p_{t,j} e^{-\epsilon l_{t,j}}}; $$
			 \vspace{-5mm}
% 		\STATE Base-learner $\calB_i$ updates $M_{t+1,i}$ with step size $\eta_{i}$ for all $i$, per \Cref{alg:OFW_Memoryless_Control};
        \FOR {each base-learner $\calB_i$}
		\STATE Compute 
			 \vspace{-3mm}
            $$M_{t,i}^\prime=\arg \min _{M \in \mathcal{M}}\left\langle \nabla_M \widetilde{f}_t(M_t), M\right\rangle_{\mathrm{F}};$$
			 \vspace{-5mm}
		\STATE Update $M_{t+1,i}=(1-\eta_i) M_{t,i}+\eta_i M_{t,i}^\prime$;
		\ENDFOR
		\STATE Observe the state $x_{t+1}$ and calculate the noise $w_t = x_{t+1} - A_t x_t - B_t u_t$;
		\ENDFOR
	\end{algorithmic}\label{alg:Meta_OFW_Control}
\end{algorithm}

%% file: 6-Conclusion.tex
\section{Conclusion} \label{sec:con}

% Projection operations are a typical computation bottleneck in online learning. 
% In this paper, we enable projection-free online learning within the framework of \textit{Online Convex Optimization with Memory} (OCO-M) ---OCO-M captures how the history of decisions affects the current outcome by allowing the online learning loss functions to depend on both current and past decisions. 
% Particularly, we introduce the first projection-free meta-base learning algorithm with memory that minimizes dynamic regret, \ie that minimizes the suboptimality against \textit{any} sequence of time-varying decisions. 
% We are motivated by artificial intelligence applications where autonomous agents need to adapt to time-varying environments in real-time, accounting for how past decisions affect the present.  
% Examples of such applications are: online control of dynamical systems; statistical arbitrage; and  time series prediction. 
% The algorithm builds on the {Online Frank-Wolf} (\OFW) and \Hedge algorithms. 
% We demonstrate how our algorithm can be applied to the online control of linear time-varying systems in the presence of unknown and unpredictable process noise, and develop the first projection-free controller with memory and bounded dynamic regret against any linear time-varying control policy.

% \myParagraph{Summary}
We provided \MetaOFW (\Cref{alg:Meta_OFW}), the first projection-free algorithm  with bounded dynamic regret for OCO with memory in time-varying environments (\Cref{theorem:MetaOFW}).
To develop \MetaOFW, we employed the projection-free algorithm \OFW along with \Hedge. 
% and provided novel dynamic regret guarantees (\Cref{theorem:OFW_Memory_Known_VT}).  
Further, we applied \MetaOFW to the online non-stochastic control problem to control linear time-varying systems that are corrupted with unknown and unpredictable noise (\Cref{alg:Meta_OFW_Control}).
We thus developed a projection-free controller with memory and bounded dynamic regret against any linear time-varying control policy (\Cref{theorem:MetaOFW_Control}), instead of against only static linear control policies. To this end, we also proved that the DAC policy class \citep{agarwal2019online} can approximate linear time-varying feedback controllers (\Cref{prop:DAC_sufficiency} in \Cref{app-subsec:DAC_sufficiency}).

% % Can put it back if accepted
% \myParagraph{Future Work} \MetaOFW is observed to have better performance than \Scream. We will investigate tighter bounds for \MetaOFW to justify the observed performance. 
% % \MetaOFW fails to match the lower dynamic regret bound for OCO-M, \ie the $\Omega(\sqrt{ T\left(1+C_T\right)})$ proved by \citep{zhao2022non}. In our future work, we will investigate novel projection-free algorithms for OCO-M to match the lower bound without sacrificing efficiency.   

% Also, we will apply the algorithms to real-world robotic systems such as quadrotors to demonstrate resilient online control against unpredictable disturbances such as wind.  To this end, we will also extend the algorithms to partially-observed systems subject to safety guarantees.

% % Additionally, our applications to online non-stochastic control focus on fully-observed systems without constraints on state and control input. While this is infeasible in general, since the systems usually lack full-state information and need to ensure safety requirements, which usually take the form of state and control input constraints. We aim to extend our method to the setting of partially-observed systems with safety guarantee.

%% file: Appendix/Appendix.tex
\onecolumn
\label{Appendix}

\input{Appendix/Appendix-OFW}

\input{Appendix/Appendix-OFW-Memory-Known-VT}

\input{Appendix/Appendix-Meta-OFW}

\input{Appendix/Appendix-Control.tex}

\input{Appendix/Appendix-Control-Lemma.tex}

\input{Appendix/Appendix-Exp.tex}

%% file: Appendix/Appendix-OFW.tex
\section{Dynamic Regret Analysis of {\fontsize{11}{8.7}\selectfont\sf{\textbf{OFW}}} in OCO without Memory}\label{app:OFW_Memoryless}

\begin{theorem}[Dynamic Regret Bound of \OFW]\label{theorem:OFW_Memoryless_Full}
Consider the OCO problem with no memory, \ie \Cref{prob:efficient_OCO_memory_dyreg} with $m=0$. Under \Cref{assumption:bounded_set} and \Cref{assumption:convexity},  \OFW achieves against any sequence of comparators $(\mathbf{v}_1, \dots, \mathbf{v}_T)\in \calX^T$ the dynamic regret
\begin{equation}
    \DyReg \leq \calO\left(\frac{1+V_{T}}{\eta} + \sqrt{T D_{T}}\right).
    \label{eq:theorem_OFW_Memoryless_Full_bound_init}
\end{equation}
Under step size $\eta=\calO\left(\frac{1}{\sqrt{T}}\right)$, we have the following dynamic regret bound,
\begin{equation}
    \DyReg \leq \calO\left(\sqrt{T}\left(1+V_{T}+\sqrt{D_{T}}\right)\right).
   	\label{eq:theorem_OFW_Memoryless_Full_bound_1}
\end{equation}
Further, select $\eta = \sqrt{\frac{c}{b}}$ with $c<b$, we have the following dynamic regret bound,
\begin{equation}
    \DyReg \leq \calO \left(\sqrt{T\left(V_{T}+D_{T}\right)}\right).
   	\label{eq:theorem_OFW_Memoryless_Full_bound_2}
\end{equation}
\end{theorem}

\subsection{Proof of \Cref{theorem:OFW_Memoryless_Full}}
\begin{proof}
The proof follows the steps of \citep[Proof of Theorem~1]{kalhan2021dynamic} but removes the assumption that the loss function $f_t$ must be smooth per the original proof in \cite{kalhan2021dynamic}.
% and (ii) enables a dynamic regret bound that holds for any sequence of comparators $\myv_1, \dots, \myv_T$ in contrast to the original proof in \citep[Theorem~1]{kalhan2021dynamic} which holds only for optimizers. 

Additionally, the proof concludes with the novel bound in \cref{eq:theorem_OFW_Memoryless_Full_bound_2}, which is enabled  
% that the original regret bound in \cite[Theorem~1]{kalhan2021dynamic} can be improved
with a constant step size (\cref{eq:theorem_OFW_Memoryless_10}) under \Cref{assumption:bounded_func}.

In more detail, to prove \Cref{theorem:OFW_Memoryless}, we take summation on both sides from $t=1$ to $T$ of \cref{eq:lemma_OFW_Memoryless} in \Cref{lemma:OFW_Memoryless}:
\begin{equation}
	\begin{aligned}
		\sum_{t=1}^{T} \left(f_{t} \left(\mathbf{x}_{t}\right)-f_{t}\left(\mathbf{v}_{t}\right)\right) \leq & \sum_{t=1}^{T} \left(f_{t, t-1}^{\sup }+f_{t-1}\left(\mathbf{v}_{t-1}\right)-f_{t}\left(\mathbf{v}_{t}\right) \right)\\
		& +(1-\eta)\left(\sum_{t=1}^{T-1}\left(f_{t}\left(\mathbf{x}_{t}\right)-f_{t}\left(\mathbf{v}_{t}\right)\right)+f_{0}\left(\mathbf{x}_{0}\right)-f_{0}\left(\mathbf{v}_{0}\right)\right) \\
		& +\eta D \sum_{t=1}^{T}\left\|\nabla f_{t}\left(\mathbf{x}_{t}\right)-\nabla f_{t-1}\left(\mathbf{x}_{t-1}\right)\right\|_2,
	\end{aligned}
\label{eq:theorem_OFW_Memoryless_1}
\end{equation}
where $f_{t, t-1}^{\text {sup }}$ is defined in \Cref{lemma:OFW_Memoryless}. Moving $(1-\eta) \sum_{t=1}^{T-1}\left(f_{t}\left(\mathbf{x}_{t}\right)-f_{t}\left(\mathbf{v}_{t}\right) \right)$ to the left side of \cref{eq:theorem_OFW_Memoryless_1}, and noting $\sum_{t=1}^{T}\left( f_{t-1}\left(\mathbf{v}_{t-1}\right)- f_{t}\left(\mathbf{v}_{t}\right)\right)=$ $f_{0}\left(\mathbf{v}_{0}\right)-f_{T}\left(\mathbf{v}_{T}\right)$, we get
\begin{equation}
	\begin{aligned}
	\eta \sum_{t=1}^{T-1}\left(f_{t}\left(\mathbf{x}_{t}\right)-f_{t}\left(\mathbf{v}_{t}\right)\right)+\left(f_{T}\left(\mathbf{x}_{T}\right)-f_{T}\left(\mathbf{v}_{T}\right)\right) \leq & \sum_{t=1}^{T} f_{t, t-1}^{\mathrm{sup}}+(1-\eta)\left(f_{0}\left(\mathbf{x}_{0}\right)-f_{0}\left(\mathbf{v}_{0}\right)\right)+f_{0}\left(\mathbf{v}_{0}\right)-f_{T}\left(\mathbf{v}_{T}\right) \\
	&+\eta D \sum_{t=1}^{T}\left\|\nabla f_{t}\left(\mathbf{x}_{t}\right)-\nabla f_{t-1}\left(\mathbf{x}_{t-1}\right)\right\|_2.
	\end{aligned}
\label{eq:theorem_OFW_Memoryless_2}
\end{equation}
Subtracting $(1-\eta)\left(f_{T}\left(\mathbf{x}_{T}\right)-f_{T}\left(\mathbf{v}_{T}\right)\right)$ from both side of \cref{eq:theorem_OFW_Memoryless_2}, we obtain
\begin{equation}
    \begin{aligned}
    	\eta \DyReg \leq & \sum_{t=1}^{T} f_{t, t-1}^{\sup }+(1-\eta)\left(f_{0}\left(\mathbf{x}_{0}\right)-f_{T}\left(\mathbf{x}_{T}\right)-f_{0}\left(\mathbf{v}_{0}\right)+f_{T}\left(\mathbf{v}_{T}\right)\right)\\
    	&+f_{0}\left(\mathbf{v}_{0}\right)-f_{T}\left(\mathbf{v}_{T}\right)+\eta D \sum_{t=1}^{T}\left\|\nabla f_{t}\left(\mathbf{x}_{t}\right)-\nabla f_{t-1}\left(\mathbf{x}_{t-1}\right)\right\|_2 \\
    	=& \sum_{t=1}^{T} f_{t, t-1}^{\sup }+\eta\left(-f_{0}\left(\mathbf{x}_{0}\right)+f_{T}\left(\mathbf{x}_{T}\right)+f_{0}\left(\mathbf{v}_{0}\right)-f_{T}\left(\mathbf{v}_{T}\right)\right)\\
    	&+f_{0}\left(\mathbf{x}_{0}\right)-f_{T}\left(\mathbf{x}_{T}\right)+\eta D \sum_{t=1}^{T}\left\|\nabla f_{t}\left(\mathbf{x}_{t}\right)-\nabla f_{t-1}\left(\mathbf{x}_{t-1}\right)\right\|_2.
    \end{aligned}
\label{eq:theorem_OFW_Memoryless_3}
\end{equation}
By \Cref{assumption:bounded_func} and the Cauchy-Schwarz inequality, it holds true that: $f_{0}\left(\mathbf{x}_{0}\right)-f_{T}\left(\mathbf{x}_{T}\right)~\leq~{2}(a+c)$, and  $-f_{0}\left(\mathbf{x}_{0}\right)+f_{T}\left(\mathbf{x}_{T}\right)+f_{0}\left(\mathbf{v}_{0}\right)-f_{T}\left(\mathbf{v}_{T}\right)~\leq~4(a+c)$. Divide both sides of \cref{eq:theorem_OFW_Memoryless_3} by $\eta$ and substitute the following inequality, which holds true due to the Cauchy-Schwarz inequality,
\begin{equation}
	\sum_{t=1}^{T}\left\|\nabla f_{t}\left(\mathbf{x}_{t}\right)-\nabla f_{t-1}\left(\mathbf{x}_{t-1}\right)\right\|_2 \leq \sqrt{T \sum_{t=1}^{T}\left\|\nabla f_{t}\left(\mathbf{x}_{t}\right)-\nabla f_{t-1}\left(\mathbf{x}_{t-1}\right)\right\|_2^{2}}=\sqrt{T D_{T}}
\label{eq:theorem_OFW_Memoryless_4}
\end{equation}
into \cref{eq:theorem_OFW_Memoryless_3} with the definition of $V_{T}$ to obtain
\begin{equation}
\DyReg \leq \frac{1}{\eta} V_{T}+\frac{{2}}{\eta}(a+c) + D \sqrt{T D_{T}} + 4(a+c).
\label{eq:theorem_OFW_Memoryless_5}
\end{equation}
Selecting $\eta=\calO(\frac{1}{\sqrt{T}})$ we get
\begin{equation}
    \DyReg \leq \calO\left(\sqrt{T}\left(1 + V_{T}+\sqrt{D_{T}}\right)\right).
\label{eq:theorem_OFW_Memoryless_6}
\end{equation} 
Further, by selecting $\eta = \sqrt{\frac{c}{b}}$ with $c<b$, the dynamic regret in \cref{eq:theorem_OFW_Memoryless_5} becomes
\begin{equation}
    \begin{aligned}
        \DyReg &\leq \sqrt{\frac{b}{c}} V_{T} + D \sqrt{T D_{T}}  + \left(\sqrt{\frac{4b}{c}}+4\right)(a+c) \\
        & = \sqrt{Tb}\frac{V_{T}}{\sqrt{Tc}} + D \sqrt{T D_{T}}  + \left(\sqrt{\frac{4b}{c}}+4\right)(a+c) \\
        & = \sqrt{T V_{T} b}\frac{\sqrt{V_{T}}}{\sqrt{Tc}} + D \sqrt{T D_{T}}  + \left(\sqrt{\frac{4b}{c}}+4\right)(a+c). 
    \end{aligned}
    \label{eq:theorem_OFW_Memoryless_7}
\end{equation}
From \Cref{assumption:bounded_func}, we have the following bound of $V_T$,
\begin{equation}
    0 \leq V_{T} =\sum_{t=1}^{T} \sup _{\mathbf{x} \in \mathcal{X}}\left|f_{t}(\mathbf{x})-f_{t-1}(\mathbf{x})\right| \leq Tc, 
\end{equation}
which implies $\frac{\sqrt{V_{T}}}{\sqrt{Tc}} \leq 1$. Substituting it into \cref{eq:theorem_OFW_Memoryless_7} gives
\begin{equation}
    \DyReg \leq \sqrt{T V_{T} b} + D \sqrt{T D_{T}}  + \left(\sqrt{\frac{4b}{c}}+4\right)(a+c).
    \label{eq:theorem_OFW_Memoryless_8}
\end{equation}
Hence, the dynamic regret is bounded as
\begin{equation}
    \DyReg \leq \calO\left(\sqrt{T}\left(\sqrt{V_{T}}+\sqrt{D_{T}}\right)\right).
\label{eq:theorem_OFW_Memoryless_9}
\end{equation}
Equivalently, due to the Cauchy-Schwarz inequality, the bound can be written as
\begin{equation}
    \DyReg \leq \calO \left(\sqrt{T\left(V_{T}+D_{T}\right)}\right),
\label{eq:theorem_OFW_Memoryless_10}
\end{equation}

\end{proof}

\subsection{Proof of \Cref{lemma:OFW_Memoryless}}

\begin{lemma}\label{lemma:OFW_Memoryless}
Under \Cref{assumption:bounded_set} and \Cref{assumption:convexity}, \Cref{alg:OFW_Memoryless} satisfies the following descent relations against any sequence of comparators $(\mathbf{v}_1, \dots, \mathbf{v}_T)\in \calX^T$,
	\begin{equation}
    	\begin{aligned}
        	f_{t}\left(\mathbf{x}_{t}\right)-f_{t}\left(\mathbf{v}_{t}\right) \leq\;& f_{t, t-1}^{\text {sup }}+(1-\eta)\left(f_{t-1}\left(\mathbf{x}_{t-1}\right)-f_{t-1}\left(\mathbf{v}_{t-1}\right)\right) \\
        	&+f_{t-1}\left(\mathbf{v}_{t-1}\right)-f_{t}\left(\mathbf{v}_{t}\right) +\eta D\left\|\nabla f_{t}\left(\mathbf{x}_{t}\right)-\nabla f_{t-1}\left(\mathbf{x}_{t-1}\right)\right\|_2
    	\end{aligned}
    	\label{eq:lemma_OFW_Memoryless}
	\end{equation}
	where $f_{t, t-1}^{\text {sup}}\triangleq\sup _{\mathbf{x} \in \mathcal{X}}\left|f_{t}(\mathbf{x})-f_{t-1}(\mathbf{x})\right|$ is the instantaneous maximum cost variation.
\end{lemma}

\begin{proof}The proof follows similar steps of \citep[Proof of Lemma~1]{kalhan2021dynamic} but removes the assumption that the loss function $f_t$ must be smooth per the original proof in \cite{kalhan2021dynamic}.

By convexity of $f_t(\cdot)$, we have
	\begin{equation}
    	f_{t}\left(\mathbf{x}_{t}\right) \leq  f_{t}\left(\mathbf{x}_{t-1}\right)+\left\langle\nabla f_{t}\left(\mathbf{x}_{t}\right), \mathbf{x}_{t}-\mathbf{x}_{t-1}\right\rangle.
    	\label{eq:lemma_OFW_Memoryless_1}
	\end{equation}
Substituting the update step $\mathbf{x}_{t}=(1-\eta) \mathbf{x}_{t-1}+\eta \mathbf{x}_{t-1}^\prime$ in \Cref{alg:OFW_Memoryless}, \ie $\mathbf{x}_{t}-\mathbf{x}_{t-1}=\eta\left(\mathbf{x}_{t-1}^\prime-\mathbf{x}_{t-1}\right)$, into \cref{eq:lemma_OFW_Memoryless_1},
	\begin{equation}
    	f_{t}\left(\mathbf{x}_{t}\right) \leq  f_{t}\left(\mathbf{x}_{t-1}\right)+\eta\left\langle\nabla f_{t}\left(\mathbf{x}_{t}\right), \mathbf{x}_{t-1}^\prime-\mathbf{x}_{t-1}\right\rangle .
    	\label{eq:lemma_OFW_Memoryless_2}
	\end{equation}
Adding and subtracting the terms $\eta\left\langle\nabla f_{t-1}\left(\mathbf{x}_{t-1}\right), \mathbf{x}_{t-1}^\prime-\mathbf{x}_{t-1}\right\rangle$ to the right hand side of \cref{eq:lemma_OFW_Memoryless_2}, we obtain
	\begin{equation}
    	\begin{aligned}
        	f_{t}\left(\mathbf{x}_{t}\right) \leq & f_{t}\left(\mathbf{x}_{t-1}\right)+\eta\left\langle\nabla f_{t}\left(\mathbf{x}_{t}\right)-\nabla f_{t-1}\left(\mathbf{x}_{t-1}\right), \mathbf{x}_{t-1}^\prime-\mathbf{x}_{t-1}\right\rangle \\
        	&+\eta\left\langle\nabla f_{t-1}\left(\mathbf{x}_{t-1}\right), \mathbf{x}_{t-1}^\prime-\mathbf{x}_{t-1}\right\rangle .
        	\end{aligned}
    	\label{eq:lemma_OFW_Memoryless_3}
	\end{equation}
Next, by the optimality condition of $\mathbf{x}_{t-1}^\prime$, \ie 
	\begin{equation}
     \left\langle\mathbf{x}_{t-1}^\prime, \nabla f_{t-1}\left(\mathbf{x}_{t-1}\right)\right\rangle = \min _{\mathbf{x} \in \mathcal{X}}\left\langle\mathbf{x}, \nabla f_{t-1}\left(\mathbf{x}_{t-1}\right)\right\rangle \leq \left\langle\mathbf{v}_{t-1}, \nabla f_{t-1}\left(\mathbf{x}_{t-1}\right)\right\rangle ,
	    \label{eq:lemma_OFW_Memoryless_4}
	\end{equation}
and substituting into \cref{eq:lemma_OFW_Memoryless_3} leads to
	\begin{equation}
    	\begin{aligned}
        	f_{t}\left(\mathbf{x}_{t}\right) \leq \;& f_{t}\left(\mathbf{x}_{t-1}\right)+\eta\left\langle\nabla f_{t}\left(\mathbf{x}_{t}\right)-\nabla f_{t-1}\left(\mathbf{x}_{t-1}\right), \mathbf{x}_{t-1}^\prime-\mathbf{x}_{t-1}\right\rangle \\
        	&+\eta\left\langle\nabla f_{t-1}\left(\mathbf{x}_{t-1}\right), \mathbf{v}_{t-1}-\mathbf{x}_{t-1}\right\rangle.
    	\end{aligned}
    	\label{eq:lemma_OFW_Memoryless_5}
	\end{equation}
By convexity of $f_{t-1}(\cdot)$ in \cref{eq:lemma_OFW_Memoryless_5}, we have
	\begin{equation}
    	\begin{aligned}
        	f_{t}\left(\mathbf{x}_{t}\right) \leq \;& f_{t}\left(\mathbf{x}_{t-1}\right)+\eta\left\langle\nabla f_{t}\left(\mathbf{x}_{t}\right)-\nabla f_{t-1}\left(\mathbf{x}_{t-1}\right), \mathbf{x}_{t-1}^\prime-\mathbf{x}_{t-1}\right\rangle \\
        	&+\eta\left(f_{t-1}\left(\mathbf{v}_{t-1}\right)-f_{t-1}\left(\mathbf{x}_{t-1}\right)\right).
    	\end{aligned}
    	\label{eq:lemma_OFW_Memoryless_6}
	\end{equation}
Subtracting $f_{t}\left(\mathbf{v}_{t}\right)$ from both sides of \cref{eq:lemma_OFW_Memoryless_6} gives
	\begin{equation}
	\begin{aligned}
	f_{t}\left(\mathbf{x}_{t}\right)-f_{t}\left(\mathbf{v}_{t}\right) \leq\; & f_{t}\left(\mathbf{x}_{t-1}\right)-f_{t}\left(\mathbf{v}_{t}\right) + \eta\left\langle\nabla f_{t}\left(\mathbf{x}_{t}\right)-\nabla f_{t-1}\left(\mathbf{x}_{t-1}\right), \mathbf{x}_{t-1}^\prime-\mathbf{x}_{t-1}\right\rangle \\
	&+\eta\left(f_{t-1}\left(\mathbf{v}_{t-1}\right)-f_{t-1}\left(\mathbf{x}_{t-1}\right)\right) .
	\end{aligned}
	\label{eq:lemma_OFW_Memoryless_7}
	\end{equation}
Next, consider the term $f_{t}\left(\mathbf{x}_{t-1}\right)-f_{t}\left(\mathbf{v}_{t}\right)$ from the right hand side of \cref{eq:lemma_OFW_Memoryless_7}. We can bound it as follows:
	\begin{equation}
    	\begin{aligned}
    	    f_{t}\left(\mathbf{x}_{t-1}\right)-f_{t}\left(\mathbf{v}_{t}\right)&=f_{t}\left(\mathbf{x}_{t-1}\right)-f_{t-1}\left(\mathbf{x}_{t-1}\right)+f_{t-1}\left(\mathbf{x}_{t-1}\right)-f_{t-1}\left(\mathbf{v}_{t-1}\right)+f_{t-1}\left(\mathbf{v}_{t-1}\right)-f_{t}\left(\mathbf{v}_{t}\right) \\
    	    &\leq  f_{t, t-1}^{\text {sup }}+f_{t-1}\left(\mathbf{x}_{t-1}\right)-f_{t-1}\left(\mathbf{v}_{t-1}\right) +f_{t-1}\left(\mathbf{v}_{t-1}\right)-f_{t}\left(\mathbf{v}_{t}\right),
    	\end{aligned}
	\label{eq:lemma_OFW_Memoryless_8}
	\end{equation}
where $f_{t, t-1}^{\text {sup }}\triangleq\sup _{\mathbf{x} \in \mathcal{X}}\left|f_{t}(\mathbf{x})-f_{t-1}(\mathbf{x})\right|$. Substituting \cref{eq:lemma_OFW_Memoryless_8} into \cref{eq:lemma_OFW_Memoryless_7}, we obtain
	\begin{equation}
    	\begin{aligned}
    	f_{t}\left(\mathbf{x}_{t}\right)- f_{t}\left(\mathbf{v}_{t}\right) \leq & f_{t, t-1}^{\text {sup }} +\eta\left\langle\nabla f_{t}\left(\mathbf{x}_{t}\right)-\nabla f_{t-1}\left(\mathbf{x}_{t-1}\right), \mathbf{x}_{t-1}^\prime-\mathbf{x}_{t-1}\right\rangle \\
    	&+(1-\eta)\left(f_{t-1}\left(\mathbf{x}_{t-1}\right)-f_{t-1}\left(\mathbf{v}_{t-1}\right)\right)+f_{t-1}\left(\mathbf{v}_{t-1}\right)-f_{t}\left(\mathbf{v}_{t}\right). 
    	\end{aligned}
	\label{eq:lemma_OFW_Memoryless_9}
	\end{equation}
Applying the Cauchy-Schwarz inequality, we get
	\begin{equation}
    	\begin{aligned}
    	f_{t}\left(\mathbf{x}_{t}\right)-f_{t}\left(\mathbf{v}_{t}\right) 
    	\leq & f_{t, t-1}^{\text {sup }} +\eta\left\|\nabla f_{t}\left(\mathbf{x}_{t}\right)-\nabla f_{t-1}\left(\mathbf{x}_{t-1}\right)\right\|_2 \left\|\mathbf{x}_{t-1}^\prime-\mathbf{x}_{t-1}\right\|_2  \\
    	&+(1-\eta)\left(f_{t-1}\left(\mathbf{x}_{t-1}\right)-f_{t-1}\left(\mathbf{v}_{t-1}\right)\right)+f_{t-1}\left(\mathbf{v}_{t-1}\right)-f_{t}\left(\mathbf{v}_{t}\right).
    	\end{aligned}
	\label{eq:lemma_OFW_Memoryless_10}
	\end{equation}
Utilizing now \Cref{assumption:bounded_set} provides the result in \Cref{lemma:OFW_Memoryless}.
\end{proof}

%% file: Appendix/Appendix-OFW-Memory-Known-VT.tex
\section{Dynamic Regret Analysis of {\fontsize{11}{8.7}\selectfont\sf{\textbf{OFW}}} in OCO-M}\label{app:OFW_Known_VT_Memory}

% \subsection{Motivating Meta-Base Algorithm: An Ideal Case with Known Loss Variation $V_T$}\label{subsec:OFW_Known_VT}
% In the following, we first demonstrate the challenge of simply applying \OFW to the problem of OCO-M, which motivates the development of the meta-base algorithm, \ie~\MetaOFW present in \Cref{subsec:Meta-OFW}. 

% We can upper-bound the dynamic regret as follows per \cite[Proof of Theorem~3.1]{anava2015online},
% \begin{equation}
% 	\begin{aligned}
% 		\DyReg \leq & \underbrace{ \sum_{t=1}^T \widetilde{f}_t\left(\mathbf{x}_t\right)-\sum_{t=1}^T \widetilde{f}_t\left(\mathbf{v}_t\right)}_{\text{unary cost}} + \underbrace{\lambda \sum_{t=2}^T\left\|\mathbf{x}_t-\mathbf{x}_{t-1}\right\|_2}_{\text{switching cost}} +  \underbrace{\lambda \sum_{t=2}^T\left\|\mathbf{v}_t-\mathbf{v}_{t-1}\right\|_2}_{\text{path length}},
% 	\end{aligned}
% 	\label{eq:DyReg_def}
% \end{equation}
% where $\lambda\triangleq m^{2} L$.

\begin{theorem}[Dynamic Regret Bound of \OFW with Loss Variation Dependent Step Size]\label{corollary:OFW_Memory_Known_VT}
Under \Cref{assumption:bounded_set} to \Cref{assumption:gradient}, running \OFW over unary functions $\widetilde{f}_1, \ldots, \widetilde{f}_T$ with step size $\eta = \calO \left( \sqrt{(1+{V}_T)/T} \right) $  achieves against any sequence of comparators  $(\mathbf{v}_1, \dots, \mathbf{v}_T)\in \calX^T$ the dynamic regret in \cref{eq:DyReg_decomposition}
\begin{equation}
		\begin{aligned}
			\DyReg & \leq \mathcal{O}\left(\sqrt{ T\left(1+{V}_T+D_T\right)} + C_T \right) \\
			 & \leq \mathcal{O}\left(\sqrt{ T\left(1+{V}_T+D_T+C_T\right)} \right).
		\end{aligned}
	\label{eq:theorem_OFW_Memory_Known_VT_bound_app}
\end{equation}
\end{theorem}

\begin{proof}

From the dynamic regret analysis in \cref{eq:theorem_OFW_Memoryless_5}, we know that running \OFW over unary function gives
\begin{equation}
	\sum_{t=1}^{T}	 \widetilde{f}_{t}\left(\mathbf{x}_{t}\right)-\sum_{t=1}^{T} \widetilde{f}_{t}\left(\mathbf{v}_{t}\right) \leq \frac{1}{\eta} (V_{T}+\alpha) + D \sqrt{T D_{T}} + \rho,
\label{eq:theorem_OFW_Memory_4}
\end{equation}
where $\alpha \triangleq{2}(a+c)$ and $\rho \triangleq {4}(a+c)$. 

Next, we consider the switching cost of the decisions, \ie $\sum_{t=2}^{T}\left\|\mathbf{x}_{t-1}-\mathbf{x}_{t}\right\|_{2}$. By the update rule of \OFW, we can derive an upper bound for the switching cost,
\begin{equation}
	\sum_{t=1}^{T}\left\|\mathbf{x}_{t}-\mathbf{x}_{t-1}\right\|_2=\eta \sum_{t=1}^{T}\left\|\mathbf{x}_{t-1}^\prime-\mathbf{x}_{t-1}\right\|_2 \leq \eta T D.
	\label{eq:theorem_OFW_Memory_5}
\end{equation}

% Combining \cref{eq:theorem_OFW_Memory_4} and \cref{eq:theorem_OFW_Memory_5} yields
% \begin{equation}
% 	\sum_{t=1}^{T}	 \widetilde{f}_{t}\left(\mathbf{x}_{t}\right)-\sum_{t=1}^{T} \widetilde{f}_{t}\left(\mathbf{v}_{t}\right) + \sum_{t=1}^{T}\left\|\mathbf{x}_{t}-\mathbf{x}_{t-1}\right\|_2 \leq \eta T \lambda D + \frac{1}{\eta}(V_T + \alpha) + D \sqrt{T D_{T}} + \rho. \qedhere
% 	\label{eq:theorem_OFW_Memory_6}
% \end{equation}
% % \begin{equation}
% % 	\sum_{t=1}^{T} f_{t}\left(\mathbf{x}_{t-m}, \ldots, \mathbf{x}_{t}\right)-\sum_{t=1}^{T} f_{t}\left(\mathbf{v}_{t-m}, \ldots, \mathbf{v}_{t}\right) \leq \eta T \lambda D + \frac{1}{\eta}(V_T + \alpha) + D \sqrt{T D_{T}} + \lambda C_{T} + \rho. \qedhere
% % 	\label{eq:theorem_OFW_Memory_6}
% % \end{equation}

Combining \cref{eq:theorem_OFW_Memory_4,eq:theorem_OFW_Memory_5}, and given the definition of path length $C_T \triangleq \sum_{t=2}^T \left\|\mathbf{v}_t-\mathbf{v}_{t-1}\right\|_2$, we have
\begin{equation}
    \DyReg \leq \eta \lambda T D + \frac{1}{\eta}(V_T + \alpha) + D \sqrt{T D_{T}} + \lambda C_T + \rho.
\end{equation}

Substituting $\eta = \sqrt{(1+V_T)/T}$ into the above equation, we directly obtain
\begin{equation}
	\begin{aligned}
		\DyReg & \leq \mathcal{O}\left(\sqrt{ T (1+V_T) } + \sqrt{ T D_T } + C_T \right) \\
		& \leq \mathcal{O}\left(\sqrt{ T\left(1+V_T+D_T\right)} + C_T \right) \\
		& \leq \mathcal{O}\left(\sqrt{T\left(1+V_{T}+D_{T}\right)+C_{T}^{2}}\right)\\
		& \leq \mathcal{O}\left(\sqrt{T\left(1+V_{T}+D_{T}\right) + T C_{T}}\right)\\
		& = \mathcal{O}\left(\sqrt{T\left(1+V_{T}+D_{T}+C_{T}\right) }\right),
		\end{aligned}
\end{equation}
where the second and third inequalities hold due to the Cauchy-Schwarz inequality, and the fourth inequality holds due to \Cref{assumption:bounded_set}, \ie $0 \leq C_T = \sum_{t=2}^{T}\left\|\mathbf{v}_{t}-\mathbf{v}_{t-1}\right\|_{2} \leq TD$.
\end{proof}

%% file: Appendix/Appendix-Meta-OFW.tex
\section{Dynamic Regret Analysis of {\fontsize{11}{8.7}\selectfont\sf{\textbf{Meta-OFW}}}}\label{app:Meta-OFW}

\subsection{Preliminaries: Online Mirror Descent}\label{app:OMD}
We present useful results of Online Mirror Descent (\OMD), which enables the dynamic regret analysis for meta-algorithm, \ie \Hedge. Consider the standard OCO setting, and the sequence of online convex functions are $\left\{h_{t}\right\}_{t=1, \ldots, T}$ with $h_{t}: \mathcal{X} \mapsto \mathbb{R}$. \OMD starts from any $\mathbf{x}_{1} \in \mathcal{X}$, and at iteration $t$, the \OMD algorithm performs the following update
\begin{equation}
	\mathbf{x}_{t+1}=\underset{\mathbf{x} \in \mathcal{X}}{\operatorname{argmin}} \quad \eta\left\langle\nabla h_{t}\left(\mathbf{x}_{t}\right), \mathbf{x}\right\rangle+\mathcal{D}_{\psi}\left(\mathbf{x}, \mathbf{x}_{t}\right),
	\label{eq:OMD_1}
\end{equation}
where $\eta>0$ is the step size. The regularizer $\psi: \mathcal{X} \mapsto \mathbb{R}$ is a differentiable convex function defined on $\mathcal{X}$ and is assumed (without loss of generality) to be 1-strongly convex w.r.t. some norm $\|\cdot\|$ over $\mathcal{X}$. The induced Bregman divergence $\mathcal{D}_{\psi}$ is defined by $\mathcal{D}_{\psi}(\mathbf{x}, \mathbf{y})=\psi(\mathbf{x})-\psi(\mathbf{y})-\langle\nabla \psi(\mathbf{y}), \mathbf{x}-\mathbf{y}\rangle$.

The following generic result gives an upper bound of dynamic regret with switching cost of \OMD, which can be regarded as a generalization of OGD from gradient descent (for Euclidean norm) to mirror descent (for general primal-dual norm).

\begin{theorem}(Dynamic Regret Bound of \OMD with Switching Cost \citep[Theorem~9]{zhao2022non}; {\citep[Theorem~2]{zhao2020understand}})\label{theorem:OMD}
Provided that $\mathcal{D}_{\psi}(\mathbf{x}, \mathbf{z})-\mathcal{D}_{\psi}(\mathbf{y}, \mathbf{z}) \leq \gamma\|\mathbf{x}-\mathbf{y}\|$ for any $\mathbf{x}, \mathbf{y}, \mathbf{z} \in \mathcal{X}$, \OMD in \cref{eq:OMD_1} achieves against any sequence of comparators $(\mathbf{v}_1, \dots, \mathbf{v}_T)\in \calX^T$ that
	\begin{equation}
		\sum_{t=1}^{T} h_{t}\left(\mathbf{x}_{t}\right)-\sum_{t=1}^{T} h_{t}\left(\mathbf{v}_{t}\right)+\lambda \sum_{t=2}^{T}\left\|\mathbf{x}_{t}-\mathbf{x}_{t-1}\right\| \leq \frac{1}{\eta}\left(R^{2}+\gamma C_{T}\right)+\eta\left(\lambda G+G^{2}\right) T,
		\label{eq:OMD_2}
	\end{equation}
	where $R^{2}\triangleq\sup _{\mathbf{x}, \mathbf{y} \in \mathcal{X}} \mathcal{D}_{\psi}(\mathbf{x}, \mathbf{y})$, $G \triangleq \sup_{\forall t}\left\|\nabla h_{t}(\cdot)\right\|_{*}$, and $\left\| \cdot \right\|_{*}$ is the dual norm, and $\lambda$ is a positive constant term. 
\end{theorem}

\begin{remark} \Cref{theorem:OMD} provides a way to analysis the dynamic regret and switching cost of \OMD algorithm. By flexibly choosing the regularizer $\psi$ and comparator sequence $\mathbf{v}_{1}, \ldots, \mathbf{v}_{T}$, we can obtain the following two corollary \citep{zhao2022non}, which correspond to dynamic regret with switching cost of \OGD (\Cref{corollary:regret_bound_OGD}) and static regret with switching cost of \Hedge (meta-regret) (\Cref{corollary:regret_bound_entropy}), respectively. 
\end{remark}
%A technical caveat is that when deriving the static regret, the Bregman divergence is not required to satisfy the Lipschitz condition.

% As we mentioned earlier, dynamic regret with switching cost of OGD is a special case of \OMD by properly choosing the regularized function $\psi$. We give a formal statement in the following corollary \citep{zhao2022non}.

\begin{corollary}\label{corollary:regret_bound_OGD}
Setting the $\ell_{2}$ regularizer $\psi(\mathbf{x})=\frac{1}{2}\|\mathbf{x}\|_{2}^{2}$ and step size $\eta>0$ for \OMD, suppose $\left\|\nabla \widetilde{f}_{t}(\mathbf{x})\right\|_{2} \leq G$ and $\left\|\mathbf{x}-\mathbf{y}\right\|_{2} \leq D$ hold for all $\mathbf{x}, \mathbf{y} \in \mathcal{X}$ and $t \in \TimeSeq$, then we have
\begin{equation}
	\sum_{t=1}^{T} \tilde{f}_{t}\left(\mathbf{x}_{t}\right)-\sum_{t=1}^{T} \widetilde{f}_{t}\left(\mathbf{v}_{t}\right) + \lambda \sum_{t=2}^{T}\left\|\mathbf{x}_{t}-\mathbf{x}_{t-1}\right\|_{2} \leq \frac{1}{2 \eta}\left(D^{2}+2 D C_{T}\right) + \eta \left(G^{2}+\lambda G\right) T,
	\label{eq:OMD_15}
\end{equation}
which holds for any comparator sequence $\mathbf{v}_{1}, \ldots, \mathbf{v}_{T} \in \mathcal{X}$, and $C_{T}=\sum_{t=2}^{T}\left\|\mathbf{v}_{t-1}-\mathbf{v}_{t}\right\|_{2}$ is the path-length that measures the cumulative movements of the comparator sequence.
\end{corollary}

Further, we present a corollary regarding the static regret with switching cost for the meta-algorithm, which is essentially a specialization of \OMD algorithm by setting the negative-entropy regularizer.

\begin{corollary}\label{corollary:regret_bound_entropy}
Setting the negative-entropy regularizer $\psi(\boldsymbol{p})=\sum_{i=1}^{N} p_{i} \log p_{i}$ and learning rate $\varepsilon>0$ for \OMD, suppose $\left\|\ell_{t}\right\|_{\infty} \leq G$ holds for any $t \in \TimeSeq$ and the algorithm starts from the initial weight $p_{1} \in \Delta_{N}$, then we have
\begin{equation}
	\sum_{t=1}^{T}\left\langle\boldsymbol{p}_{t}, \boldsymbol{\ell}_{t}\right\rangle-\sum_{t=1}^{T} \ell_{t, i} + \lambda \sum_{t=2}^{T}\left\|\boldsymbol{p}_{t}-\boldsymbol{p}_{t-1}\right\|_{1} \leq \frac{\ln \left(1 / p_{1, i}\right)}{\varepsilon}+\varepsilon\left(\lambda G+G^{2}\right) T .
	\label{eq:OMD_16}
\end{equation}
\end{corollary}

Before presenting the proof of \Cref{theorem:OMD}, we first present three useful lemmas.

\begin{lemma}\citep[Lemma~3.2]{chen1993convergence}\label{lemma:chen1993convergence}
Let $\calX$ be a convex set in a Banach space $\mathcal{B}$. Let $f: \mathcal{X} \mapsto \mathbb{R}$ be a closed proper convex function on $\mathcal{X}$. Given a convex regularizer $\psi: \mathcal{X} \mapsto \mathbb{R}$, we denote its induced Bregman divergence by $\mathcal{D}_{\psi}(\cdot, \cdot)$. Then, any update of the form
\begin{equation}
	\mathbf{x}_{k}=\underset{\mathbf{x} \in \mathcal{X}}{\arg \min }\left\{f(\mathbf{x})+\mathcal{D}_{\psi}\left(\mathbf{x}, \mathbf{x}_{k-1}\right)\right\}
	\label{eq:OMD_3}
\end{equation}
satisfies the following inequality for any $\mathbf{u} \in \mathcal{X}$,
\begin{equation}
	f\left(\mathbf{x}_{k}\right)-f(\mathbf{u}) \leq \mathcal{D}_{\psi}\left(\mathbf{u}, \mathbf{x}_{k-1}\right)-\mathcal{D}_{\psi}\left(\mathbf{u}, \mathbf{x}_{k}\right)-\mathcal{D}_{\psi}\left(\mathbf{x}_{k}, \mathbf{x}_{k-1}\right).
	\label{eq:OMD_4}
\end{equation}
\end{lemma}

\begin{lemma}(\citep[Lemma~1]{duchi2010composite};\cite{vishnoi2021algorithms})\label{lemma7:zhao2022non}
If the regularizer $\psi: \mathcal{X} \mapsto \mathbb{R}$ is $\lambda$-strongly convex with respect to a norm $\|\cdot\|$, then we have the following lower bound for the induced Bregman divergence: $\mathcal{D}_{\psi}(\mathbf{x}, \mathbf{y}) \geq \frac{\lambda}{2}\|\mathbf{x}-\mathbf{y}\|^{2}$.
\end{lemma}
% \begin{proof}
% 	By the definition of strong convexity, we know that for any $\mathbf{x}, \mathbf{y} \in \mathcal{X}, \psi(\mathbf{x}) \geq \psi(\mathbf{y})+\nabla \psi(\mathbf{y})^{\top}(\mathbf{x}-\mathbf{y})+$ $\frac{\lambda}{2}\|\mathbf{x}-\mathbf{y}\|^{2}$. Reformulating the inequality and combining the definition of Bregman divergence, we know that $D_{\psi}(\mathbf{x}, \mathbf{y}) \triangleq \psi(\mathbf{x})-\psi(\mathbf{y})-\nabla \psi(\mathbf{y})^{\top}(\mathbf{x}-\mathbf{y}) \geq \frac{\lambda}{2}\|\mathbf{x}-\mathbf{y}\|^{2}$, which ends the proof.
% \end{proof}

\begin{lemma}(Switching Cost of \OMD \citep[Lemma~10]{zhao2022non})\label{lemma:OMD_swtiching}
For \OMD in \cref{eq:OMD_1}, the instantaneous switching cost is at most
\begin{equation}
	\left\|\mathbf{x}_{t}-\mathbf{x}_{t+1}\right\| \leq \eta\left\|\nabla h_{t}\left(\mathbf{x}_{t}\right)\right\|_{*}.
	\label{eq:OMD_5}
\end{equation}
\end{lemma}

\begin{remark}
    There is an earlier result in \citep[Lemma~3]{zhao2020understand} that establishes $\left\|\mathbf{x}_{t}-\mathbf{x}_{t+1}\right\| \leq 2\eta\left\|\nabla h_{t}\left(\mathbf{x}_{t}\right)\right\|_{*}$ for the switching cost of \OMD, instead of the inequality in \cref{eq:OMD_5} where the multiplicative factor $2$ is instead absent.
\end{remark}

Based on \Cref{lemma:OMD_swtiching}, we can now prove \Cref{theorem:OMD}.

\begin{proof}[Proof of \Cref{theorem:OMD}]
The following proof is given in \citep[Theorem~9]{zhao2022non}. The dynamic regret of \OMD with switching cost can be bounded by following \citep[Theorem~2]{zhao2020understand}. The differences of \citep[Theorem~9]{zhao2022non} and \citep[Theorem~2]{zhao2020understand} are that: i) \citep[Theorem~2]{zhao2020understand} bound the switching cost by $\left\|\mathbf{x}_{t}-\mathbf{x}_{t+1}\right\| \leq 2\eta\left\|\nabla h_{t}\left(\mathbf{x}_{t}\right)\right\|_{*}$, instead of \Cref{lemma:OMD_swtiching} where the multiplicative factor $2$ is absent; ii) \citep[Theorem~2]{zhao2020understand} derives a tighter bound on term (b) in \cref{eq:OMD_10,eq:OMD_12} using \Cref{lemma7:zhao2022non}.

Notice that the dynamic regret can be decomposed in the following way:
\begin{equation}
	\begin{aligned}
		\sum_{t=1}^{T} h_{t}\left(\mathbf{x}_{t}\right)-\sum_{t=1}^{T} h_{t}\left(\mathbf{v}_{t}\right) & \leq \sum_{t=1}^{T}\left\langle\nabla h_{t}\left(\mathbf{x}_{t}\right), \mathbf{x}_{t}-\mathbf{v}_{t}\right\rangle \\
		&=\underbrace{\sum_{t=1}^{T}\left\langle\nabla h_{t}\left(\mathbf{x}_{t}\right), \mathbf{x}_{t}-\mathbf{x}_{t+1}\right\rangle}_{\text {term (a) }}+\underbrace{\sum_{t=1}^{T}\left\langle\nabla h_{t}\left(\mathbf{x}_{t}\right), \mathbf{x}_{t+1}-\mathbf{v}_{t}\right\rangle}_{\text {term (b) }} .
	\end{aligned}
	\label{eq:OMD_10}
\end{equation}
From Hölder's inequality and \Cref{lemma:OMD_swtiching}, we can bound \text {term (a)} as
\begin{equation}
	\text{term (a)} \leq \sum_{t=1}^{T}\left\|\nabla h_{t}\left(\mathbf{x}_{t}\right)\right\|_{*}\left\|\mathbf{x}_{t}-\mathbf{x}_{t+1}\right\| \leq \eta \sum_{t=1}^{T}\left\|\nabla h_{t}\left(\mathbf{x}_{t}\right)\right\|_{*}^{2} .
	\label{eq:OMD_11}
\end{equation}
For \text {term (b)}, we have
\begin{equation}
	\begin{aligned}
		\text {term (b)} & \leq \frac{1}{\eta} \sum_{t=1}^{T}\left(\mathcal{D}_{\psi}\left(\mathbf{v}_{t}, \mathbf{x}_{t}\right)-\mathcal{D}_{\psi}\left(\mathbf{v}_{t}, \mathbf{x}_{t+1}\right)-\mathcal{D}_{\psi}\left(\mathbf{x}_{t+1}, \mathbf{x}_{t}\right)\right) \\
		& \leq \frac{1}{\eta} \sum_{t=2}^{T}\left(\mathcal{D}_{\psi}\left(\mathbf{v}_{t}, \mathbf{x}_{t}\right)-\mathcal{D}_{\psi}\left(\mathbf{v}_{t-1}, \mathbf{x}_{t}\right)\right)+\frac{1}{\eta}\mathcal{D}_{\psi}\left(\mathbf{v}_{1}, \mathbf{x}_{1}\right) \\
		& \leq \frac{\gamma}{\eta} \sum_{t=2}^{T}\left\|\mathbf{v}_{t}-\mathbf{v}_{t-1}\right\|+\frac{1}{\eta} R^{2},
	\end{aligned}
	\label{eq:OMD_12}
\end{equation}
where the first inequality holds by \Cref{lemma:chen1993convergence}, the second inequality holds by the non-negativity of the Bregman divergence, and the last inequality holds due to $\mathcal{D}_{\psi}(\mathbf{x}, \mathbf{z})-\mathcal{D}_{\psi}(\mathbf{y}, \mathbf{z}) \leq \gamma\|\mathbf{x}-\mathbf{y}\|$ for any $\mathbf{x}, \mathbf{y}, \mathbf{z} \in \mathcal{X}$.

By \Cref{lemma7:zhao2022non}, the switching cost is bounded as
\begin{equation}
	\sum_{t=2}^{T}\left\|\mathbf{x}_{t}-\mathbf{x}_{t-1}\right\| \leq \eta \sum_{t=2}^{T}\left\|\nabla h_{t-1}\left(\mathbf{x}_{t-1}\right)\right\|_{*}.
	\label{eq:OMD_13}
\end{equation}
Combining \cref{eq:OMD_11,eq:OMD_12,eq:OMD_13}, we obtain
\begin{equation}
	\begin{aligned}
		\sum_{t=1}^{T} h_{t}\left(\mathbf{x}_{t}\right)-\sum_{t=1}^{T} h_{t}\left(\mathbf{v}_{t}\right) + \lambda \sum_{t=2}^{T}\left\|\mathbf{x}_{t}-\mathbf{x}_{t-1}\right\| \leq & \frac{1}{\eta}\left(R^{2}+\gamma C_{T}\right)+\eta \sum_{t=1}^{T}\left(\lambda\left\|\nabla h_{t}\left(\mathbf{x}_{t}\right)\right\|_{*}+\left\|\nabla h_{t-1}\left(\mathbf{x}_{t-1}\right)\right\|_{*}^{2}\right) \\
		\leq & \frac{1}{\eta}\left(R^{2}+\gamma C_{T}\right)+\eta\left(\lambda G+G^{2}\right) T.
	\end{aligned} \qedhere
	\label{eq:OMD_14}
\end{equation}
\end{proof}

\subsection{Proof of \Cref{theorem:MetaOFW}}\label{app-subsec:theorem_MetaOFW}
\begin{proof}
According to \citep[Proof of Theorem~3.1]{anava2015online}, the coordinate-Lipschitz continuity of $f_{t}$ (\Cref{assumption:Lipschitzness}) implies that
\begin{equation}
    \begin{aligned}
    	\left|f_{t}\left(\mathbf{x}_{t-m}, \ldots, \mathbf{x}_{t}\right)-\widetilde{f}_{t}\left(\mathbf{x}_{t}\right)\right| &\leq L \sum_{i=1}^{m}\left\|\mathbf{x}_{t}-\mathbf{x}_{t-i}\right\|_2 \\
    	&\leq L \sum_{i=1}^{m} \sum_{l=1}^{i} \left\|\mathbf{x}_{t-l+1}-\mathbf{x}_{t-l}\right\|_2\\
    	&\leq m L \sum_{i=1}^{m}\left\|\mathbf{x}_{t-i+1}-\mathbf{x}_{t-i}\right\|_2 .
    \end{aligned}
    \label{eq:theorem_OFW_Memory_1}
\end{equation}
Taking the summation from $t=1$ to $T$ gives
\begin{equation}
	\left|\sum_{t=1}^{T} f_{t}\left(\mathbf{x}_{t-m}, \ldots, \mathbf{x}_{t}\right)-\sum_{t=1}^{T} \widetilde{f}_{t}\left(\mathbf{x}_{t}\right)\right| \leq m L \sum_{t=1}^{T}\sum_{i=1}^{m}\left\|\mathbf{x}_{t-i+1}-\mathbf{x}_{t-i}\right\|_2 \leq m^{2}
	L \sum_{t=1}^{T}\left\|\mathbf{x}_{t}-\mathbf{x}_{t-1}\right\|_2,
	\label{eq:theorem_OFW_Memory_2}
\end{equation}
and the dynamic regret can be thus upper bounded by
\begin{equation}
	\begin{aligned}
		\DyReg &= \sum_{t=1}^{T} f_{t}\left(\mathbf{x}_{t-m}, \ldots, \mathbf{x}_{t}\right)-\sum_{t=1}^{T} f_{t}\left(\mathbf{v}_{t-m}, \ldots, \mathbf{v}_{t}\right) \\
% 		&\leq \left| \sum_{t=1}^{T} f_{t}\left(\mathbf{x}_{t-m}, \ldots, \mathbf{x}_{t}\right)-\sum_{t=1}^{T} f_{t}\left(\mathbf{v}_{t-m}, \ldots, \mathbf{v}_{t}\right) \right| \\
% 		&\leq \left| \sum_{t=1}^{T} f_{t}\left(\mathbf{x}_{t-m}, \ldots, \mathbf{x}_{t}\right) \right| + \left|\sum_{t=1}^{T} f_{t}\left(\mathbf{v}_{t-m}, \ldots, \mathbf{v}_{t}\right) \right| \\
		& \leq \underbrace{\sum_{t=1}^{T} \widetilde{f}_{t}\left(\mathbf{x}_{t}\right)-\sum_{t=1}^{T} \tilde{f}_{t}\left(\mathbf{v}_{t}\right)}_{\text {dynamic regret over unary loss }}+\underbrace{\lambda \sum_{t=1}^{T}\left\|\mathbf{x}_{t}-\mathbf{x}_{t-1}\right\|_2}_{\text {switching cost of decisions}}+ \underbrace{\lambda \sum_{t=1}^{T}\left\|\mathbf{v}_{t}-\mathbf{v}_{t-1}\right\|_2}_{\text {switching cost of comparators}},
	\end{aligned}
	\label{eq:theorem_OFW_Memory_3}
\end{equation}
where $\mathbf{x}_\tau$ and $\mathbf{v}_\tau$ can be set as $\mathbf{0}$ for all $\tau \leq 0$, and $\lambda\triangleq m^{2} L$.

By \Cref{lemma:MetaOFW_DyReg_decomposition}, we aim to bound the meta-regret and base-regret terms.

\myParagraph{Meta-regret bound} 
Denote by $\mathbf{e}_{i}$ the $i$-th standard basis of $\mathbb{R}^{N}$-space. Since the meta-algorithm performs Hedge over the switching-cost-regularized loss $\boldsymbol{\ell}_{t} \in \mathbb{R}^{N}$, \Cref{corollary:regret_bound_entropy} implies that for any $i \in \{1, \dots, N\}$,
\begin{equation}
	\begin{aligned}
		\sum_{t=1}^{T}\left\langle\boldsymbol{p}_{t}, \boldsymbol{\ell}_{t}\right\rangle-\sum_{t=1}^{T} \ell_{t, i}+\lambda D \sum_{t=2}^{T}\left\|\boldsymbol{p}_{t}-\boldsymbol{p}_{t-1}\right\|_{1} & \leq \varepsilon\left(\lambda D G_{\text {meta }}+G_{\text {meta }}^{2}\right) T+\frac{\mathcal{D}_{\psi}\left(\mathbf{e}_{i}, \boldsymbol{p}_{1}\right)}{\varepsilon} \\
		&=\varepsilon(2 \lambda+G)(\lambda+G) D^{2} T+\frac{\ln \left(1 / p_{1, i}\right)}{\varepsilon} \\
		& \leq \varepsilon(2 \lambda+G)(\lambda+G) D^{2} T+\frac{2 \ln (i+1)}{\varepsilon} .
	\end{aligned}
	\label{eq:theorem_Meta_OFW_1}
\end{equation}
where the first inequality holds due to $G_{\text {meta}}=\max_{t \in \TimeSeq}\left\|\ell_{t}\right\|_{\infty} \leq(\lambda+G) D$, and the last inequality holds by plugging in the initialization of weights \ie $\boldsymbol{p}_{1} \in \Delta_{N}$ with $p_{1, i}=\frac{1}{i(i+1)} \cdot \frac{N+1}{N}$ for any $i \in \{1, \dots, N\}$. By choosing the optimal learning rate $\varepsilon=\varepsilon^{*}=\sqrt{\frac{2}{(2 \lambda+G)(\lambda+G) D^{2} T}}$, we can obtain the following upper bound for the meta-regret,
\begin{equation}
	\sum_{t=1}^{T}\left\langle\boldsymbol{p}_{t}, \boldsymbol{\ell}_{t}\right\rangle-\sum_{t=1}^{T} \ell_{t, i}+\lambda D \sum_{t=2}^{T}\left\|\boldsymbol{p}_{t}-\boldsymbol{p}_{t-1}\right\|_{1} \leq D \sqrt{2(2 \lambda+G)(\lambda+G) T}(1+\ln (i+1)) .
	\label{eq:theorem_Meta_OFW_2}
\end{equation}
%Note that the dependence of learning rate tuning on $T$ can be removed by either a time-varying tuning or doubling trick.

\myParagraph{Base-regret bound}
As specified by \MetaOFW, there are multiple base-learners, each performing \OFW over the linearized loss with a particular step size $\eta_{i} \in \mathcal{H}$ for base-learner $\mathcal{B}_{i}$. As a result, \cref{eq:theorem_OFW_Memory_4,eq:theorem_OFW_Memory_5} and the definition of $\bar{D}_T$ imply that the base-regret satisfies
\begin{equation}
	\sum_{t=1}^{T} g_{t}\left(\mathbf{x}_{t, i}\right)-\sum_{t=1}^{T} g_{t}\left(\mathbf{v}_{t}\right)+\lambda \sum_{t=2}^{T}\left\|\mathbf{x}_{t, i}-\mathbf{x}_{t-1, i}\right\|_{2} \leq \eta_i \lambda T D + \frac{1}{\eta_i}(V_T + \alpha) + D \sqrt{T \bar{D}_T} + \rho,
	\label{eq:theorem_Meta_OFW_3}
\end{equation}
which holds for any comparator sequence $\mathbf{v}_{1}, \ldots, \mathbf{v}_{T} \in \mathcal{X}$ as well as any base-learner $i \in \{1, \dots, N\}$.

\myParagraph{Dynamic regret bound}
Due to the boundedness of the loss variation $V_T$, we know that the optimal step size $\eta_{*}$ provably lies in the range of $\left[\eta_{1}, \eta_{N}\right]$. In particular, given \cref{eq:theorem_Meta_OFW_3}, the optimal step size $\eta_{*}$ is 
\begin{equation}
	\sqrt{\frac{\alpha}{ \lambda T D}} \leq \eta_{*} = \sqrt{\frac{V_T + \alpha}{\lambda T D}} \leq\sqrt{\frac{T c + \alpha}{\lambda T D}}.
	\label{eq:theorem_Meta_OFW_4}
\end{equation}
Furthermore, by the construction of the step size pool in \cref{eq:step_size_pool}, there exists an index $i^{*} \in \{1, \dots, N\}$, such that $\eta_{i^{*}}~\leq~\eta_{*}~\leq~\eta_{i^{*}+1}=2\eta_{i^{*}}$, with
\begin{equation}
	i^{*} \leq\left\lceil\frac{1}{2} \log _{2}\left(1+\frac{V_{T}}{\alpha}\right)\right\rceil+1.
	\label{eq:theorem_Meta_OFW_5}
\end{equation} 

Notice that the meta-base decomposition in \Cref{lemma:MetaOFW_DyReg_decomposition} holds for any index of base-learners $i$. Thus, in particular, we can choose the index $i^{*}$ and achieve the following result by using the meta-regret and base-regret bounds in \cref{eq:theorem_Meta_OFW_2,eq:theorem_Meta_OFW_3},
\begin{equation}
	\begin{aligned}
	& \sum_{t=1}^{T} \widetilde{f}_{t}\left(\mathbf{x}_{t}\right)-\sum_{t=1}^{T} \tilde{f}_{t}\left(\mathbf{v}_{t}\right)+\lambda \sum_{t=2}^{T}\left\|\mathbf{x}_{t}-\mathbf{x}_{t-1}\right\|_{2} \\
	\leq & \underbrace{\sum_{t=1}^{T}\left(\left\langle\boldsymbol{p}_{t}, \ell_{t}\right\rangle-\ell_{t, i^{*}}\right)+\lambda D \sum_{t=2}^{T}\left\|\boldsymbol{p}_{t}-\boldsymbol{p}_{t-1}\right\|_{1}}_{\text {meta-regret }}+\underbrace{\sum_{t=1}^{T}\left(g_{t}\left(\mathbf{x}_{t, i^{*}}\right)-g_{t}\left(\mathbf{v}_{t}\right)\right)+\lambda \sum_{t=2}^{T}\left\|\mathbf{x}_{t, i^{*}}-\mathbf{x}_{t-1, i^{*}}\right\|_{2}}_{\text {base-regret }} \\
	\leq & D \sqrt{2(2 \lambda+G)(\lambda+G) T}\left(1+\ln \left(i^{*}+1\right)\right)+\left(\eta_{i^{*}} \lambda T D + \frac{1}{\eta_{i^{*}}}(V_T + \alpha) + D \sqrt{T \bar{D}_T} + \rho \right) \\
	\leq & D \sqrt{2(2 \lambda+G)(\lambda+G) T}\left(1+\ln \left(i^{*}+1\right)\right)+ \eta_{*} \lambda TD+\frac{2}{\eta_{*}}\left(V_{T}+\alpha\right) + D \sqrt{T \bar{D}_T} + \rho \\
	\leq & \underbrace{2 D(\lambda+G) \sqrt{T}\left(1+\ln \left(\left\lceil\frac{1}{2} \log _{2}\left(1+\frac{V_{T}}{\alpha}\right)\right\rceil+2\right)\right)}_{\leq \mathcal{O}\left(\sqrt{T}\left(1+\log \log V_T \right)\right)}+\underbrace{3\sqrt{\lambda TD\left(V_{T}+\alpha \right)}+D \sqrt{T \bar{D}_T} }_{\leq \mathcal{O}\left(\sqrt{T\left(1+V_{T}+\bar{D}_T\right)}\right)} + \rho\\
	\leq & \mathcal{O}\left(\sqrt{T\left(1+V_{T}+\bar{D}_T\right)}\right) .
	\end{aligned}
	\label{eq:theorem_Meta_OFW_6}
\end{equation}

Combining \cref{eq:DyReg_decomposition} and \cref{eq:theorem_Meta_OFW_6} gives
\begin{equation}
	\begin{aligned}
		\DyReg \leq & \sum_{t=1}^{T} \widetilde{f}_{t}\left(\mathbf{x}_{t}\right)-\sum_{t=1}^{T} \widetilde{f}_{t}\left(\mathbf{v}_{t}\right) + \lambda \sum_{t=2}^{T}\left\|\mathbf{x}_{t}-\mathbf{x}_{t-1}\right\|_{2}+\lambda \sum_{t=2}^{T}\left\|\mathbf{v}_{t}-\mathbf{v}_{t-1}\right\|_{2}\\
		\leq & \mathcal{O}\left(\sqrt{T\left(1+V_{T} + \bar{D}_T\right)}\right)+\mathcal{O}\left(C_{T}\right) \\
		\leq & \mathcal{O}\left(\sqrt{T\left(1+V_{T}+\bar{D}_T\right)+C_{T}^{2}}\right)\\
		\leq & \mathcal{O}\left(\sqrt{T\left(1+V_{T}+\bar{D}_T\right) + T C_{T}}\right)\\
		= & \mathcal{O}\left(\sqrt{T\left(1+V_{T}+\bar{D}_T+C_{T}\right) }\right),\\
	\end{aligned}
	\label{eq:theorem_Meta_OFW_7}
\end{equation}
where the third inequality holds due to Cauchy-Schwartz inequality, and the fourth inequality holds due to \Cref{assumption:bounded_set}, \ie $0 \leq C_T = \sum_{t=2}^{T}\left\|\mathbf{v}_{t}-\mathbf{v}_{t-1}\right\|_{2} \leq TD$.
\end{proof}

% \subsection{Proof of \Cref{lemma:MetaOFW_DyReg_decomposition}}
\subsection{Decomposition of Unary Cost and Switching Cost \citep{zhao2022non}}
    \begin{lemma}[Decomposition of Unary Cost and Switching Cost \citep{zhao2022non}]\label{lemma:MetaOFW_DyReg_decomposition}
% The switching cost of meta-learner can be decomposed as
% 	\begin{equation}
% 		\begin{aligned} 								
% 			\sum_{t=2}^T\left\|\mathbf{x}_t-\mathbf{x}_{t-1}\right\|_2 \leq & D \sum_{t=2}^T\left\|\boldsymbol{p}_t-\boldsymbol{p}_{t-1}\right\|_1 \\
% 			& + \sum_{t=2}^T \sum_{i=1}^N p_{t, i}\left\|\mathbf{x}_{t, i}-\mathbf{x}_{t-1, i}\right\|_2 .
% 		\end{aligned}
% 	\label{eq:switching_cost_decomposition}
% 	\end{equation}
The unary and switching costs in \cref{eq:DyReg_decomposition} can be decomposed as follows:
	\begin{equation}
		\begin{aligned} 
			 \sum_{t=1}^T 	\tilde{f}_t\left(\mathbf{x}_t\right)-\sum_{t=1}^T \tilde{f}_t\left(\mathbf{v}_t\right)+\lambda \sum_{t=2}^T\left\|\mathbf{x}_t-\mathbf{x}_{t-1}\right\|_2 
			 %\leq & \sum_{t=1}^T\left\langle\nabla \widetilde{f}_t\left(\mathbf{x}_t\right), \mathbf{x}_t-\mathbf{v}_t\right\rangle+\lambda D \sum_{t=2}^T\left\|\boldsymbol{p}_t-\boldsymbol{p}_{t-1}\right\|_1 \\ 
			 %& +\lambda \sum_{t=2}^T \sum_{i=1}^N p_{t, i}\left\|\mathbf{x}_{t, i}-\mathbf{x}_{t-1, i}\right\|_2 \\
			 & \leq \underbrace{\sum_{t=1}^T\left(\left\langle\boldsymbol{p}_t, \boldsymbol{\ell}_t\right\rangle-\ell_{t, i}\right)+\lambda D \sum_{t=2}^T\left\|\boldsymbol{p}_t-\boldsymbol{p}_{t-1}\right\|_1}_{\text{meta-regret }} \\
			 &\quad+\underbrace{\sum_{t=1}^T\left(g_t\left(\mathbf{x}_{t, i}\right)-g_t\left(\mathbf{v}_t\right)\right)+\lambda \sum_{t=2}^T\left\|\mathbf{x}_{t, i}-\mathbf{x}_{t-1, i}\right\|_2}_{\text{base-regret }} ,
		\end{aligned}
		\label{eq:MetaOFW_DyReg_decomposition}
	\end{equation}
	which holds for any $i \in \{1, \dots, N\}$.
\end{lemma}

\begin{proof}
By the meta-base structure, the final decision of each round is $\mathbf{x}_{t}=\sum_{i=1}^{N} p_{t, i} \mathbf{x}_{t, i}$. Therefore, we can expand the switching cost of the final prediction sequence as
\begin{equation}
	\begin{aligned}
		\left\|\mathbf{x}_{t}-\mathbf{x}_{t-1}\right\|_{2} &=\left\|\sum_{i=1}^{N} p_{t, i} \mathbf{x}_{t, i}-\sum_{i=1}^{N} p_{t-1, i} \mathbf{x}_{t-1, i}\right\|_{2} \\
		& \leq\left\|\sum_{i=1}^{N} p_{t, i} \mathbf{x}_{t, i}-\sum_{i=1}^{N} p_{t, i} \mathbf{x}_{t-1, i}\right\|_{2}+\left\|\sum_{i=1}^{N} p_{t, i} \mathbf{x}_{t-1, i}-\sum_{i=1}^{N} p_{t-1, i} \mathbf{x}_{t-1, i}\right\|_{2} \\
		& \leq \sum_{i=1}^{N} p_{t, i}\left\|\mathbf{x}_{t, i}-\mathbf{x}_{t-1, i}\right\|_{2}+D \sum_{i=1}^{N}\left|p_{t, i}-p_{t-1, i}\right| \\
		&=\sum_{i=1}^{N} p_{t, i}\left\|\mathbf{x}_{t, i}-\mathbf{x}_{t-1, i}\right\|_{2}+D\left\|\boldsymbol{p}_{t}-\boldsymbol{p}_{t-1}\right\|_{1},
	\end{aligned}
	\label{eq:lemma_swtiching_cost}
\end{equation}
where the first inequality holds due to the triangle inequality and the second inequality is true owing to the boundedness of the feasible domain (\Cref{assumption:bounded_set}). 

By \cref{eq:lemma_swtiching_cost} and convexity of $\widetilde{f}_{t}(\cdot)$, we have
\begin{equation}
	\begin{aligned}
	& \sum_{t=1}^{T} \widetilde{f}_{t}\left(\mathbf{x}_{t}\right)-\sum_{t=1}^{T} \widetilde{f}_{t}\left(\mathbf{v}_{t}\right)+\lambda \sum_{t=2}^{T}\left\|\mathbf{x}_{t}-\mathbf{x}_{t-1}\right\|_{2} \\
	{\leq} & \sum_{t=1}^{T}\left\langle\nabla \widetilde{f}_{t}\left(\mathbf{x}_{t}\right), \mathbf{x}_{t}-\mathbf{v}_{t}\right\rangle+\lambda D \sum_{t=2}^{T}\left\|\boldsymbol{p}_{t}-\boldsymbol{p}_{t-1}\right\|_{1}+\lambda \sum_{t=2}^{T} \sum_{i=1}^{N} p_{t, i}\left\|\mathbf{x}_{t, i}-\mathbf{x}_{t-1, i}\right\|_{2}. 
	\end{aligned}
	\label{eq:lemma_MetaOFW_DyReg_decomposition_1}
\end{equation}
Next, we manipulate \cref{eq:lemma_MetaOFW_DyReg_decomposition_1} using the definition of linearized loss $g_{t}(\mathbf{x})=\left\langle\nabla \widetilde{f}_{t}\left(\mathbf{x}_{t}\right), \mathbf{x}\right\rangle$,
\begin{equation}
	\begin{aligned}
		& \sum_{t=1}^{T} \widetilde{f}_{t}\left(\mathbf{x}_{t}\right)-\sum_{t=1}^{T} \widetilde{f}_{t}\left(\mathbf{v}_{t}\right)+\lambda \sum_{t=2}^{T}\left\|\mathbf{x}_{t}-\mathbf{x}_{t-1}\right\|_{2} \\
		\leq & \sum_{t=1}^{T} \sum_{i=1}^{N} p_{t, i}\left(\left\langle\nabla \widetilde{f}_{t}\left(\mathbf{x}_{t}\right), \mathbf{x}_{t, i}\right\rangle+\lambda\left\|\mathbf{x}_{t, i}-\mathbf{x}_{t-1, i}\right\|_{2}\right)-\sum_{t=1}^{T}\left(\left\langle\nabla \widetilde{f}_{t}\left(\mathbf{x}_{t}\right), \mathbf{x}_{t, i}\right\rangle+\lambda\left\|\mathbf{x}_{t, i}-\mathbf{x}_{t-1, i}\right\|_{2}\right) \\
	&+\lambda D \sum_{t=2}^{T}\left\|\boldsymbol{p}_{t}-\boldsymbol{p}_{t-1}\right\|_{1}+\sum_{t=1}^{T}\left(\left\langle\nabla \widetilde{f}_{t}\left(\mathbf{x}_{t}\right), \mathbf{x}_{t, i}\right\rangle-\left\langle\nabla \widetilde{f}_{t}\left(\mathbf{x}_{t}\right), \mathbf{v}_{t}\right\rangle\right)+\lambda \sum_{t=2}^{T}\left\|\mathbf{x}_{t, i}-\mathbf{x}_{t-1, i}\right\|_{2} \\
	=& \underbrace{\sum_{t=1}^{T}\left(\left\langle\boldsymbol{p}_{t}, \boldsymbol{\ell}_{t}\right\rangle-\ell_{t, i}\right)+\lambda D \sum_{t=2}^{T}\left\|\boldsymbol{p}_{t}-\boldsymbol{p}_{t-1}\right\|_{1}}_{\text {meta-regret }}+\underbrace{\sum_{t=1}^{T}\left(g_{t}\left(\mathbf{x}_{t, i}\right)-g_{t}\left(\mathbf{v}_{t}\right)\right)+\lambda \sum_{t=2}^{T}\left\|\mathbf{x}_{t, i}-\mathbf{x}_{t-1, i}\right\|_{2}}_{\text {base-regret }}.\qedhere
	\end{aligned}
	\label{eq:lemma_MetaOFW_DyReg_decomposition_2}
\end{equation}

\end{proof}

\subsection{Proof of \Cref{corollary:regret_bound_entropy} \citep{zhao2022non}}

\begin{proof}
From the proof of \Cref{theorem:OMD}, we can obtain that
\begin{equation}
	\sum_{t=1}^{T}\left\langle\boldsymbol{p}_{t}, \boldsymbol{\ell}_{t}\right\rangle-\sum_{t=1}^{T} \ell_{t, i} + \lambda \sum_{t=2}^{T}\left\|\boldsymbol{p}_{t}-\boldsymbol{p}_{t-1}\right\|_{1} \leq \frac{\mathcal{D}_{\psi}\left(\mathbf{e}_{i}, \boldsymbol{p}_{1}\right)}{\varepsilon}+\varepsilon\left(\lambda G+G^{2}\right) T ,
	\label{eq:OMD_17}
\end{equation}
where $\mathbf{e}_{i}$ the $i$-th standard basis of $\mathbb{R}^{N}$-space. When choosing the negative-entropy regularizer, the induced Bregman divergence becomes Kullback-Leibler divergence, i.e., $\mathcal{D}_{\psi}(\boldsymbol{q}, \boldsymbol{p})=\mathrm{KL}(\boldsymbol{q}, \boldsymbol{p})=\sum_{i=1}^{N} q_{i} \ln \left(q_{i} / p_{i}\right)$. Therefore, $\mathcal{D}_{\psi}\left(\boldsymbol{e}_{i}, \boldsymbol{p}_{1}\right)=\ln \left(1 / p_{1, i}\right)$, which implies the desired result.\qedhere
\end{proof}

%% file: Appendix/Appendix-Control.tex
\section{Dynamic Regret Analysis of Non-Stochastic Control}\label{app:control}

\subsection{Definition of Strongly Stable Controller}\label{def:kappa_gamma_stable}
\begin{definition}
A linear policy $K$ is $(\kappa, \gamma)$-strongly stable if there exist matrices $L$ and $Q$ satisfying $A-BK = QLQ^{-1}$, such that following two conditions are satisfied:
\begin{enumerate}
    \item The spectral norm of L is strictly smaller than one, \ie $\|L\|_{\mathrm{op}} \leq 1-\gamma$;
    
    \item The controller and the transforming matrices are bounded, $\|K\|_{\mathrm{op}}$, $\|Q\|_{\mathrm{op}}$, and $\|Q\|^{-1}_{\mathrm{op}}\leq \kappa$.
\end{enumerate}
\end{definition}

\subsection{Proof of \Cref{theorem:MetaOFW_Control}}

\begin{proof}
We decompose the dynamic regret as follows,
\begin{equation}
	\begin{aligned}
		&\sum_{t=0}^{T} c_{t}\left(x_{t}, u_{t}\right)-\sum_{t=0}^{T} c_{t}\left(x_{t}^{*}, u_{t}^{*}\right) \\
		=& \sum_{t=0}^{T} c_{t}\left(x_{t}\left(M_{0: t-1}\right), u_{t}\left(M_{0: t}\right)\right)-\sum_{t=0}^{T} c_{t}\left(x_{t}^{*}\left(0 \right), u_{t}^{*}\left(0\right)\right) \\
		=& \underbrace{\sum_{t=0}^{T} c_{t}\left(x_{t}\left(M_{0: t-1}\right), u_{t}\left(M_{0: t}\right)\right)-\sum_{t=0}^{T} f_{t}\left(M_{t-1-H: t}\right)}_{\text{term (a)}}  + \underbrace{\sum_{t=0}^{T} f_{t}\left(M_{t-1-H: t}\right)-\sum_{t=0}^{T} f_{t}\left(M_{t-1-H: t}^{*}\right)}_{\text{term (b)}} \\
		&+\underbrace{\sum_{t=0}^{T} f_{t}\left(M_{t-1-H: t}^{*}\right)-\sum_{t=0}^{T} c_{t}\left(x_{t}\left(M_{0: t-1}^{*}\right), u_{t}\left(M_{0: t}^{*}\right)\right)}_{\text{term (c)}} + \underbrace{\sum_{t=0}^{T} c_{t}\left(x_{t}\left(M_{0: t-1}^{*}\right), u_{t}\left(M_{0: t}^{*}\right)\right) - \sum_{t=0}^{T} c_{t}\left(x_{t}^{*}\left(0 \right), u_{t}^{*}\left(0\right)\right)}_{\text{term (d)}},
	\end{aligned}
\end{equation}
where term (a) and term (c) are the approximation errors induced by truncation of loss function, term (b) is the dynamic regret over the truncated loss functions $\left\{f_{t}\right\}_{t=0, \ldots, T}$, and term (d) is the approximation error of DAC controller.

By \Cref{theorem:truncate_loss}, term (a) and term (c) are bounded by
\begin{equation}
	\text{term (a)} + \text{term (c)} \leq 4 T G_{c} D^{2} \kappa^{3}(1-\gamma)^{H+1}.
	\label{eq:theorem_MetaOFW_Control_1}
\end{equation}

Then by \Cref{prop:DAC_sufficiency}, we can bound term (d) as
\begin{equation}
	\text{term (d)}  \leq  \frac{4 T G_{c} D W H \kappa_B^{2} \kappa^{6}(1-\gamma)^{H-1}}{\gamma} .
	\label{eq:theorem_MetaOFW_Control_2}
\end{equation}

Next, we focus on the term (b). Similar to \cref{eq:DyReg_decomposition}, we decompose term (b) as follows,
\begin{equation}
	\begin{aligned}
		\text{term (b)}  &=\sum_{t=0}^{T} f_{t}\left(M_{t-1-H: t}\right)-\sum_{t=0}^{T} f_{t}\left(M_{t-1-H: t}^{*}\right) \\
		& \leq \sum_{t=0}^{T} \widetilde{f}_{t}\left(M_{t}\right)-\sum_{t=0}^{T} \widetilde{f}_{t}\left(M_{t}^{*}\right)+\zeta \sum_{t=1}^{T}\left\|M_{t-1}-M_{t}\right\|_{\mathrm{F}}+\zeta \sum_{t=1}^{T}\left\|M_{t-1}^{*}-M_{t}^{*}\right\|_{\mathrm{F}} \\
		& \leq \sum_{t=0}^{T}\left\langle\nabla_{M} \widetilde{f}_{t}\left(M_{t}\right), M_{t}-M_{t}^{*}\right\rangle+\zeta \sum_{t=1}^{T}\left\|M_{t-1}-M_{t}\right\|_{\mathrm{F}}+\zeta \sum_{t=1}^{T}\left\|M_{t-1}^{*}-M_{t}^{*}\right\|_{\mathrm{F}} \\
		&=\underbrace{\sum_{t=0}^{T} g_{t}\left(M_{t}\right)-\sum_{t=0}^{T} g_{t}\left(M_{t}^{*}\right)+\zeta \sum_{t=1}^{T}\left\|M_{t-1}-M_{t}\right\|_{\mathrm{F}}}_{\text{Dynamic Regret with Switching Cost over $\{g_{t}\}_{t=0,\dots,T}$}}+\underbrace{\zeta \sum_{t=1}^{T}\left\|M_{t-1}^{*}-M_{t}^{*}\right\|_{\mathrm{F}}}_{\text{Path Length of Comparators}},
	\end{aligned}
	\label{eq:theorem_MetaOFW_Control_3}
\end{equation}
where $\zeta=(H+2)^{2} L_{f}$ and $g_{t}(M)=\left\langle\nabla_{M} \widetilde{f}_{t}\left(M_{t}\right), M\right\rangle$ is the surrogate linearized loss. 

We now aim to analyze the dynamic regret with switching cost in \cref{eq:theorem_MetaOFW_Control_3}, which can be decomposed into meta-regret and base-regret similar to \Cref{lemma:MetaOFW_DyReg_decomposition}:
\begin{equation}
	\begin{aligned}
		& \sum_{t=0}^{T} g_{t}\left(M_{t}\right)-\sum_{t=0}^{T} g_{t}\left(M_{t}^{*}\right)+\zeta \sum_{t=1}^{T}\left\|M_{t-1}-M_{t}\right\|_{\mathrm{F}} \\
		=& \underbrace{ \sum_{t=0}^{T}\left\langle\boldsymbol{p}_{t}, \boldsymbol{\ell}_{t}\right\rangle-\sum_{t=0}^{T} \ell_{t, i} + \zeta D_f \sum_{t=1}^{T}\left\|\boldsymbol{p}_{t-1}-\boldsymbol{p}_{t}\right\|_{1}}_{\text{meta-regret }} + \underbrace{\left( \sum_{t=0}^{T} g_{t}\left(M_{t, i}\right)-\sum_{t=0}^{T} g_{t}\left(M_{t}^{*}\right)\right) + \zeta \sum_{t=1}^{T}\left\|M_{t-1, i}-M_{t, i}\right\|_{\mathrm{F}}}_{\text{base-regret }},
		\label{eq:theorem_MetaOFW_Control_4}
	\end{aligned}
\end{equation}
where $\ell_{t} \in \Delta_{N}$ is the surrogate loss vector of the meta-algorithm with $\ell_{t, i}=\zeta\left\|M_{t-1, i}-M_{t, i}\right\|_{\mathrm{F}}+g_{t}\left(M_{t, i}\right), \text { for } i \in \{1, \dots, N\}$. Note that the regret decomposition holds for any base-learner $\calB_i$. 

\myParagraph{Meta-regret bound}
By \Cref{corollary:regret_bound_entropy}, we obtain
\begin{equation}
	\begin{aligned}
		\text { meta-regret } & \leq \varepsilon\left(2 \zeta+G_{f}\right)\left(\zeta_{f}+G_{f}\right) D_{f}^{2} (T+1)+\frac{\ln \left(1 / p_{1, i}\right)}{\varepsilon} \\
		&=D_{f} \sqrt{2\left(2 \zeta+G_{f}\right)\left(\zeta+G_{f}\right) (T+1)}(1+\ln (1+i)),
		\label{eq:theorem_MetaOFW_Control_5}
	\end{aligned}
\end{equation}
where the equality holds by choosing the optimal learning rate $\varepsilon=\varepsilon^{*}=\sqrt{\frac{2}{D_f^{2}(2 \zeta+G_f)(\zeta+G_f) (T+1)}}$.

\myParagraph{Base-regret bound}
By \Cref{theorem:OFW_Control} and the definition of $\bar{D}_T$, we have
\begin{equation}
    \begin{aligned}
        \text { base-regret } &\leq \eta_i \zeta T D_f + \frac{1}{\eta_i} (V_{T}+\sigma) + D_f \sqrt{(T+1) D_{T}} + \theta \\
        &\leq \eta_i \zeta T D_f + \frac{1}{\eta_i} (V_{T}+\sigma) + D_f \sqrt{(T+1) \bar{D}_T} + \theta,
    \end{aligned}
	\label{eq:theorem_MetaOFW_Control_6}
\end{equation}
where $V_{T} \triangleq \sum_{t=0}^{T} \sup _{M \in \mathcal{M}}\left|f_{t}(M)-f_{t-1}(M)\right|$, $D_{T} \triangleq \sum_{t=0}^{T}\left\|\nabla_M f_{t}\left(M_{t}\right)-\nabla_M f_{t-1}\left(M_{t-1}\right)\right\|_{\mathrm{F}}^{2}$, $\sigma \triangleq 4\beta D^2$, and $\theta \triangleq 8 \beta D^2$.

\myParagraph{Overall regret bound}
Due to the boundedness of the loss variation $V_T$, the optimal step size $\eta_{*}$ provably lies in the range of $\left[\eta_{1}, \eta_{N}\right]$. In particular, the optimal step size $\eta_{*}$ is 
\begin{equation}
	\sqrt{\frac{\sigma}{\zeta T D_f}} \leq \eta_{*} = \sqrt{\frac{V_T + \sigma}{\zeta T D_f}} \leq\sqrt{\frac{2 \beta D^2 T + \phi}{\zeta T D_f}},
	\label{eq:theorem_MetaOFW_Control_7}
\end{equation}
where $\phi = \sigma + 2 \beta D^2$. 

Furthermore, by the construction of the step size pool in \cref{eq:step_size_pool}, there exists an index $i^{*} \in \{1, \dots, N\}$, such that $\eta_{i^{*}}~\leq~\eta_{*}~\leq~\eta_{i^{*}+1}=2\eta_{i^{*}}$, with
\begin{equation}
	i^{*} \leq\left\lceil\frac{1}{2} \log _{2}\left(1+\frac{V_{T}}{\sigma}\right)\right\rceil+1.
	\label{eq:theorem_MetaOFW_Control_8}
\end{equation} 

Since the meta-base decomposition in \cref{eq:theorem_MetaOFW_Control_4} holds for any index $i$, we can choose the index $i^{*}$ and achieve the following result by using the meta-regret and base-regret bounds in \cref{eq:theorem_MetaOFW_Control_5,eq:theorem_MetaOFW_Control_6},
\begin{equation}
	\begin{aligned}
		& \sum_{t=0}^{T} g_{t}\left(M_{t}\right)-\sum_{t=0}^{T} g_{t}\left(M_{t}^{*}\right)+\zeta \sum_{t=1}^{T}\left\|M_{t-1}-M_{t}\right\|_{\mathrm{F}} \\
		\leq & D_{f} \sqrt{2\left(2 \zeta+G_{f}\right)\left(\zeta+G_{f}\right) (T+1)}(1+\ln (i^{*}+1))+\left(\eta_{i^{*}} T D_f + \frac{1}{\eta_{i^{*}}}(V_T + \sigma) + D_f \sqrt{(T+1) \bar{D}_T} + \theta \right) \\
		\leq & D_{f} \sqrt{2\left(2 \zeta+G_{f}\right)\left(\zeta+G_{f}\right) (T+1)}(1+\ln (i^{*}+1))+ \eta_{*} T D_f+\frac{2}{\eta_{*}}\left(V_{T}+\sigma\right) + D_f \sqrt{(T+1) \bar{D}_T} + \theta \\
		\leq & D_{f} \sqrt{2\left(2 \zeta+G_{f}\right)\left(\zeta+G_{f}\right) (T+1)}(1+\ln (i^{*}+1))\left(1+\ln \left(\left\lceil\frac{1}{2} \log _{2}\left(1+\frac{ V_{T}}{\sigma}\right)\right\rceil+2\right)\right) \\
		& + 3\sqrt{T\left(V_{T}+k_{3}\right)}+D_f \sqrt{(T+1) \bar{D}_T} + \theta.
	\end{aligned}
	\label{eq:theorem_MetaOFW_Control_9}
\end{equation}

Combining \cref{eq:theorem_MetaOFW_Control_1,eq:theorem_MetaOFW_Control_2,eq:theorem_MetaOFW_Control_3,eq:theorem_MetaOFW_Control_9}, and $C_T \triangleq \sum_{t=2}^{T}\left\|M_{t-1}^{*}-M_{t}^{*}\right\|_{\mathrm{F}}$ gives
\begin{equation}
	\begin{aligned}
	&\sum_{t=0}^{T} c_{t}\left(x_{t}, u_{t}\right)-\sum_{t=0}^{T} c_{t}\left(x_{t}^{*}, u_{t}^{*}\right) \\
	\leq & 4 T G_{c} D^{2} \kappa^{3}(1-\gamma)^{H+1} + \frac{4 T G_{c} D W H \kappa_B^{2} \kappa^{6}(1-\gamma)^{H-1}}{\gamma} \\
	 & +  D_{f} \sqrt{2\left(2 \zeta+G_{f}\right)\left(\zeta+G_{f}\right) (T+1)}(1+\ln (i^{*}+1))\left(1+\ln \left(\left\lceil\frac{1}{2} \log _{2}\left(1+\frac{ V_{T}}{\sigma}\right)\right\rceil+2\right)\right) \\
	 & + 3\sqrt{T \left(V_{T}+k_{3}\right)}+D_f \sqrt{(T+1) \bar{D}_T} + \theta  +\zeta C_{T} .
	\end{aligned}
\end{equation}
Finally, by setting $H=\mathcal{O}(\log T)$, the final dynamic policy regret is bounded by $\widetilde{\mathcal{O}}\left(\sqrt{T\left(1 + V_{T} + \bar{D}_T + C_{T}\right)}\right)$.
\end{proof}

\subsection{Dynamic Regret Analysis of \OFW over $\mathcal{M}$-space}
In \Cref{theorem:OFW_Memoryless}, we have analyzed the dynamic regret of \MetaOFW over the Euclidean space. To utilize \MetaOFW for non-stochastic control, we need to generalized the result to $\mathcal{M}$-space, \ie, generalize the previous results from Euclidean norm to Frobenius norm over $\mathcal{M}$-space. To this end, we first present the dynamic regret analysis of \OFW over $\mathcal{M}$-space.

\input{Alg/Alg-OFW-Control}
\begin{theorem}\label{theorem:OFW_Control}
Suppose the function $\tilde{f}: \mathcal{M} \mapsto \mathbb{R}$ is convex, with bounded the gradient norm $G_{f}$, \ie $\left\|\nabla_{M} \widetilde{f}_{t}(M)\right\|_{\mathrm{F}} \leq G_{f}$ for any $M \in \mathcal{M}$ and $t \in \TimeSeq$, and bounded Euclidean diameter of $\mathcal{M}$-space $D_{f}$, i.e., $\sup _{M, M^{\prime} \in \mathcal{M}}\left\|M-M^{\prime}\right\|_{\mathrm{F}} \leq D_{f}$. Then for any comparator sequence $M_{1}, \ldots, M_{T} \in \mathcal{M}$, \Cref{alg:OFW_Memoryless_Control} satisfies that
\begin{equation}
	\sum_{t=1}^{T} \widetilde{f}_{t}\left(M_{t}\right)-\sum_{t=1}^{T} \widetilde{f}_{t}\left(M_{t}^{*}\right) + \zeta \sum_{t=2}^{T}\left\|M_{t-1}-M_{t}\right\|_{\mathrm{F}} \leq \eta T \zeta D_f + \frac{1}{\eta} (V_{T}+\sigma) + D_f \sqrt{T D_{T}} + \theta,
\end{equation}
where $V_{T} \triangleq \sum_{t=1}^{T} \sup _{M \in \mathcal{M}}\left|f_{t}(M)-f_{t-1}(M)\right|$, and $D_{T} \triangleq \sum_{t=1}^{T}\left\|\nabla_M f_{t}\left(M_{t}\right)-\nabla_M f_{t-1}\left(M_{t-1}\right)\right\|_{\mathrm{F}}^{2}$, $\sigma \triangleq 4\beta D^2$, and $\theta \triangleq 8 \beta D^2$.

\end{theorem}
\begin{proof}
Consider the term $\sum_{t=1}^{T} \widetilde{f}_{t}\left(M_{t}\right)-\sum_{t=1}^{T} \widetilde{f}_{t}\left(M_{t}^{*}\right)$. The bound can be developed similar to \Cref{theorem:OFW_Memoryless}, replacing vector inner product $\left< \cdot, \cdot \right>$ by matrix inner product $\left< \cdot, \cdot \right>_{\mathrm{F}}$ and replacing vector norm $\left\| \cdot, \cdot \right\|_2$ by Frobenius norm $\left\| \cdot, \cdot \right\|_{\mathrm{F}}$. Then from \cref{eq:theorem_OFW_Memoryless_5}, we directly obtain
\begin{equation}
	\sum_{t=1}^{T} \widetilde{f}_{t}\left(M_{t}\right)-\sum_{t=1}^{T} \widetilde{f}_{t}\left(M_{t}^{*}\right) \leq \frac{1}{\eta} (V_{T}+\sigma) + D_f \sqrt{T D_{T}} + \theta.
\end{equation}

On the other hand, the switching cost can be bounded by
\begin{equation}
	\begin{aligned}
		\sum_{t=2}^{T}\left\|M_{t}-M_{t-1}\right\|_{\mathrm{F}} =\eta \sum_{t=2}^{T} \left\|M_{t-1}^\prime-M_{t-1}\right\|_{\mathrm{F}} \leq \eta (T-1) D_{f} \leq \eta T D_{f}.
	\end{aligned}
\end{equation}
We complete the proof by combining the above equations.
\end{proof}

\subsection{State Transition under DAC Controller}\label{app-subsec:State_Trans_DAC}
\begin{proposition}[State Transition under DAC Controller]\label{prop:state_transition}
    Suppose the initial state is $x_0 = 0$, and one chooses the DAC controller $\pi\left(M_{t}, K_t\right)$ at iteration $t$, the reaching state is
\begin{equation}
	x_{t+1} = \widetilde{A}_{K_{t:t-h}} x_{t-h}+\sum_{i=0}^{H+h} \Psi_{t, i}^{K_t, h}\left(M_{t-h: t}\right) w_{t-i},
	\label{eq:prop_state}
\end{equation}
where $\Psi_{t, i}^{K, h}\left(M_{t-h: t}\right)$ is the transfer matrix defined as
\begin{equation}
	\begin{aligned}
		 \Psi_{t, i}^{K_t, h}\left(M_{t-h: t}\right) =  \widetilde{A}_{K_{t:t-i+1}} \mathbf{1}_{i \leq h}  +\sum_{j=0}^{h} \widetilde{A}_{K_{t:t-j+1}} B_{t-j} M_{t-j}^{[i-j-1]} \mathbf{1}_{1 \leq i-j \leq H},
		\label{eq:prop_Psi}
	\end{aligned}
\end{equation}	
where $\widetilde{A}_{K_{t:t-i}} \triangleq \prod_{\tau=t}^{t-i}\left(A_\tau-B_\tau K_\tau\right)$, and we define $\widetilde{A}_{K_{t:t-i}} \triangleq \myIdentity$ if $i~<~0$. The evolving equation holds for any $h~\in~\{0, \ldots, t\}$.
\end{proposition}

\begin{proof}
The proof follows the same step as in \citep[Lemma~4.3]{agarwal2019online}. We aim to show
\begin{equation}
	x_{t+1} = \widetilde{A}_{K_{t:t-h}} x_{t-h}+\sum_{i=0}^{H+h} \Psi_{t, i}^{K_t, h}\left(M_{t-h: t}\right) w_{t-i},
	\label{eq:prop_state_transition_1}
\end{equation}
where $\Psi_{t, i}^{K, h}\left(M_{t-h: t}\right)$ is the transfer matrix defined as
\begin{equation}
		\Psi_{t, i}^{K_t, h}\left(M_{t-h: t}\right) =   \widetilde{A}_{K_{t:t-i+1}} \mathbf{1}_{i \leq h}  +\sum_{j=0}^{h} \widetilde{A}_{K_{t:t-j+1}} B_{t-j} M_{t-j}^{[i-j-1]} \mathbf{1}_{1 \leq i-j \leq H}.
		\label{eq:prop_state_transition_2}
\end{equation}	

For any time $t$, we will prove the claim by induction. For $h=0$, we have
\begin{equation}
	\begin{aligned}
		x_{t+1} &= \widetilde{A}_{K_{t}} x_{t} + \sum_{i=1}^{H} B_{t} M_{t}^{[i-1]} w_{t-i} + w_{t} \\
		&= \widetilde{A}_{K_{t}} x_{t} + \sum_{i=0}^{H} \Psi_{t, i}^{K_t, 0}\left(M_{t}\right) w_{t-i}.
	\end{aligned}
		\label{eq:prop_state_transition_3}
\end{equation}

Suppose that \cref{eq:prop_state_transition_1} holds for some $h \geq 0$, then for $h+1$ we have
\begin{equation}
	\begin{aligned}
		x_{t+1} &= \widetilde{A}_{K_{t:t-h}} x_{t-h}+\sum_{i=0}^{H+h} \Psi_{t, i}^{K_t, h}\left(M_{t-h: t}\right) w_{t-i} \\
		&= \widetilde{A}_{K_{t:t-h}} \left(  \widetilde{A}_{K_{t-h-1}} x_{t-h-1} + \sum_{i=1}^{H} B_{t-h-1} M_{t-h-1}^{[i-1]} w_{t-h-1-i} + w_{t-h-1}  \right)+\sum_{i=0}^{H+h} \Psi_{t, i}^{K_t, h}\left(M_{t-h: t}\right) w_{t-i} \\
		&=  \widetilde{A}_{K_{t:t-h-1}} x_{t-h-1} + \sum_{i=0}^{H+h+1} \left( \Psi_{t, i}^{K_t, h}\left(M_{t-h: t}\right) +  \widetilde{A}_{K_{t:t-i+1}}\mathbf{1}_{i = h+1} +  \widetilde{A}_{K_{t:t-h}}B_{t-h-1} M_{t-h-1}^{[i-h-2]} \mathbf{1}_{1 \leq i-h-1 \leq H}   \right) w_{t-i} \\
		&= \widetilde{A}_{K_{t:t-h-1}} x_{t-h-1}+\sum_{i=0}^{H+h+1} \Psi_{t, i}^{K_t, h+1}\left(M_{t-h-1: t}\right) w_{t-i}.
	\end{aligned}
	\label{eq:prop_state_transition_4}
\end{equation}

\end{proof}

\subsection{Sufficiency of DAC Policy}\label{app-subsec:DAC_sufficiency}
% \subsection{Proof of \Cref{prop:DAC_sufficiency}}

\begin{proposition}[Sufficiency of DAC Policy]\label{prop:DAC_sufficiency}
Suppose the initial state is $x_0 = 0$, for a sequence of $(\kappa, \gamma)$ strongly stable time-varying controllers $K_{0}^{*}, \dots, K_{t}^{*}$, there exist a policy $\pi_t (K_t, M_t^{*})$, with $M_t^{*,[i]} \triangleq (K_t - K_t^{*}) \widetilde{A}_{K_{t-1:t-i}^{*}}$, where $\widetilde{A}_{K^{*}_{t:t-i}} \triangleq \prod_{\tau=t}^{t-i}\left(A_\tau-B_\tau K^{*}_\tau\right)$ and  $\widetilde{A}_{K^{*}_{t:t-i}} \triangleq \myIdentity$ if $i~<~0$, such that
\begin{equation}
	\begin{aligned}
		\sum_{t=0}^{T} \left(c_{t}\left(x_{t}\left(M_{0:t}^{*}\right), u_{t}\left(M_{0:t}^{*}\right)\right) - c_{t}\left(x_{t}^{*}\left(0\right), u_{t}^{*}\left(0\right)\right) \right) 		\leq  \frac{4 T G_{c} D W H \kappa_B^{2} \kappa^{6}(1-\gamma)^{H-1}}{\gamma}.
		\label{eq:prop_DAC_sufficiency_approx}
	\end{aligned}
\end{equation}
\end{proposition} 

\begin{proof}
The proof is analogous to \citep[Lemma~5.2]{agarwal2019online}, but adjusts the definition of $M_{0:t}^{*}$ to handle the time-varying controllers. 

By definition, the state propagated by the sequence of time-varying controller $K_{1}^{*}, \dots, K_{t}^{*}$ is 
\begin{equation}
	x_{t+1}^{*}\left(0\right) = \sum_{i=0}^{t} \widetilde{A}_{K_{t:t-i+1}^{*}} w_{t-i}.
	\label{eq:prop_DAC_sufficiency_1}
\end{equation}
Consider the state transition matrix $\Psi_{t, i}^{K_t, h}\left(M_{t-h: t}^{*}\right)$ for any $i \leq H$ and $H \leq h$. With \cref{eq:prop_state_transition_2}, we have
\begin{equation}
	\begin{aligned}
		\Psi_{t, i}^{K_t, h}\left(M_{t-h: t}^{*}\right) &= \widetilde{A}_{K_{t:t-i+1}} + \sum_{j=1}^{i} \widetilde{A}_{K_{t:t-i+j+1}} B_{t-i+j} \left(K_{t-i+j} - K_{t-i+j}^{*} \right) \widetilde{A}_{K_{t-i+j-1:t-i+1}}^{*} \\
		&= \widetilde{A}_{K_{t:t-i+1}} + \sum_{j=1}^{i} \widetilde{A}_{K_{t:t-i+j+1}} \left(\widetilde{A}_{K_{t-i+j}^{*}}- \widetilde{A}_{K_{t-i+j}}\right) \widetilde{A}_{K_{t-i+j-1:t-i+1}}^{*} \\
		&= \widetilde{A}_{K_{t:t-i+1}} + \sum_{j=1}^{i} \left(\widetilde{A}_{K_{t:t-i+j+1}} \widetilde{A}_{K_{t-i+j:t-i+1}^{*}} - \widetilde{A}_{K_{t:t-i+j}} \widetilde{A}_{K_{t-i+j-1:t-i+1}}^{*} \right) \\
		&= \widetilde{A}_{K_{t:t-i+1}^{*}},
	\end{aligned}
	\label{eq:prop_DAC_sufficiency_2}
\end{equation}
where the last equality holds by telescoping the summation. Therefore, by setting $h=t$ in \cref{eq:prop_state}, we have
\begin{equation}
	x_{t+1}\left(M_{0:t}^{*}\right) = \sum_{i=0}^{H} \widetilde{A}_{K_{t:t-i+1}^{*}} w_{t-i} + \sum_{i=H+1}^{t}	\Psi_{t, i}^{K_t, t}\left(M_{0: t}^{*}\right) w_{t-i}.
	\label{eq:prop_DAC_sufficiency_3}
\end{equation}
Combine \cref{eq:prop_DAC_sufficiency_1,eq:prop_DAC_sufficiency_3}, we get
\begin{equation}
	\begin{aligned}	
		\left\| x_{t+1}^{*}\left(0\right)  - x_{t+1}\left(M_{0:t}^{*}\right)\right\|_2 &\leq W \left( \sum_{i=H+1}^{t} \left\| 	\Psi_{t, i}^{K_t, t}\left(M_{0: t}^{*}\right) \right\|_{\mathrm{op}} + \sum_{i=H+1}^{t} \left\|  \widetilde{A}_{K_{t:t-i+1}^{*}} \right\|_{\mathrm{op}}  \right) \\
		& \leq W\left(\sum_{i=H+1}^{t}\left(2 \kappa^{2}(1-\gamma)^{i}+H \kappa_{B}^{2} \kappa^{5}(1-\gamma)^{i-1}\right)\right) \\
		& \leq W\left(2 \kappa^{2}(1-\gamma)^{H+1} \gamma^{-1}+H \kappa_{B}^{2} \kappa^{5}(1-\gamma)^{H} \gamma^{-1}\right) \\
		& \leq \kappa^2 W (1-\gamma)^{H+1} \gamma^{-1} \left(2(1-\gamma) + H \kappa_B^2 \kappa^3\right) \\
		&\leq H \kappa_B^2 \kappa^5 W (1-\gamma)^{H} \gamma^{-1} \left(2(1-\gamma) + 1\right) \\
		&\leq 2 H \kappa_B^2 \kappa^5 W (1-\gamma)^{H} \gamma^{-1},
	\end{aligned}
\label{eq:prop_DAC_sufficiency_4}
\end{equation}
where the second inequality holds due to \Cref{lemma:bounded_Psi}.

Similarly, we analyze the difference in control input,
\begin{equation}
	\begin{aligned}	
		\left\| u_{t+1}^{*}\left(0\right)  - u_{t+1}\left(M_{0:t+1}^{*}\right)\right\|_2 &= \left\| -K_{t+1}^{*} x_{t+1}^{*} - \left( -K_{t+1} x_{t+1}(M_{0:t}^{*}) + \sum_{i=1}^{H} M_{t+1}^{*,[i-1]} w_{t+1-i} \right) \right\|_2 \\
		&= \left\| -K_{t+1}^{*} x_{t+1}^{*} +K_{t+1} x_{t+1}(M_{0:t}^{*}) - \sum_{i=1}^{H} (K_{t+1} - K_{t+1}^{*}) \widetilde{A}_{K_{t:t-i+2}^{*}} w_{t+1-i}  \right\|_2 \\
		&= \left\| -K_{t+1}^{*} \left(x_{t+1}^{*} - \sum_{i=0}^{H-1}\widetilde{A}_{K_{t:t-i+1}^{*}} w_{t-i}  \right)+K_{t+1} \left( x_{t+1}(M_{0:t}^{*}) - \sum_{i=0}^{H-1} \widetilde{A}_{K_{t:t-i+1}^{*}} w_{t-i}  \right) \right\|_2 \\ 
		&= \left\| -K_{t+1}^{*}  \sum_{i=H}^{t}\widetilde{A}_{K_{t:t-i+1}^{*}} w_{t-i}  +K_{t+1}   \sum_{i=H}^{t} \Psi_{t, i}^{K_t, t}\left(M_{0: t}^{*}\right) w_{t-i}  \right\|_2 \\ 
		&\leq 2 H \kappa_B^2 \kappa^6 W (1-\gamma)^{H} \gamma^{-1}.
	\end{aligned}
	\label{eq:prop_DAC_sufficiency_5}
\end{equation}
Using \cref{eq:prop_DAC_sufficiency_4,eq:prop_DAC_sufficiency_5}, the Lipschitz assumption (\Cref{assumption:cost}), and the boundedness result (\Cref{lemma:bounded_variables}), we complete the proof.
\end{proof}

\subsection{Approximation of Truncated Loss}
% \subsection{Proof of \Cref{theorem:truncate_loss}}
\begin{theorem}[Approximation of Truncated Loss: Theorem $5.3$ of \cite{agarwal2019online}]\label{theorem:truncate_loss}
Define $D\triangleq\frac{W \kappa^{3}\left(1+H \kappa_{B} \tau\right)}{\gamma\left(1-\kappa^{2}(1-\gamma)^{H+1}\right)}+\frac{W \tau}{\gamma}$. For any $(\kappa, \gamma)$ strongly stable linear controller $K_t$ at iteration $t$, and any $\tau>0$ such that the sequence of $M_{0}, \ldots, M_{T}$ satisfies $\left\|M_{t}^{[i]}\right\|_{\mathrm{op}} \leq$ $\tau(1-\gamma)^{i}$, the approximation error between original loss and truncated loss is at most
\begin{equation}
	\begin{aligned}
	\left|\sum_{t=0}^{T} \left(c_{t}\left(x_{t}\left(M_{0: t-1}\right), u_{t}\left(M_{0: t}\right)\right)-f_{t}\left(M_{t-1-H: t}\right) \right) \right| \leq  2 T G_{c} D^{2} \kappa^{3}(1-\gamma)^{H+1}.
	\label{eq:theorem_truncate_loss}
\end{aligned}
\end{equation}
\end{theorem}

\begin{proof}
By the Lipschitz continuity and definition of the truncated loss, we get that
\begin{equation}
	\begin{aligned}
		&c_{t}\left(x_{t}\left(M_{0: t-1}\right), u_{t}\left(M_{0: t}\right)\right)-f_{t}\left(M_{t-H-1: t}\right) \\
		=& c_{t}\left(x_{t}\left(M_{0: t-1}\right), u_{t}\left(M_{0: t}\right)\right)-c_{t}\left(y_{t}\left(M_{t-H-1: t-1}\right), v_{t}\left(M_{t-H-1: t}\right)\right) \\
		\leq & G_{c} D\left(\left\|x_{t}\left(M_{0: t-1}\right)-y_{t}\left(M_{t-H-1: t-1}\right)\right\|_2 +\left\|u_{t}\left(M_{0: t}\right)-v_{t}\left(M_{t-H-1: t}\right)\right\|_2 \right) \\
		\leq & G_{c} D\left(\kappa^{2}(1-\gamma)^{H+1} D+\kappa^{3}(1-\gamma)^{H+1} D\right) \\
		\leq & 2 G_{c} D^{2} \kappa^{3}(1-\gamma)^{H+1},
	\end{aligned}
\end{equation}
where the first inequality uses the Lipschitz assumption \Cref{assumption:cost} and the second inequality uses boundedness in \Cref{lemma:bounded_variables}. The result in \cref{eq:theorem_truncate_loss} is obtained by summing over the iterations from $t=1$ to $T$.
\end{proof}

%% file: Alg/Alg-OFW-Control.tex
\begin{algorithm}[t]
	\caption{\mbox{\hspace{-.05mm}\OFW for Non-Stochastic Control.}}
	\begin{algorithmic}[1]
		\REQUIRE Time horizon $T$; step size $\eta$.
		\ENSURE Prediction $M_t$ at each time step $t=1,\ldots, T$.
		\medskip
		\STATE Initialize $M_0 \in \calM$; 
		\FOR {each time step $t = 1, \dots, T$}
		\STATE Obtain gradient $\nabla_M f_t(M_t)$;
		\STATE Compute $M_{t}^\prime=\arg \min _{M \in \mathcal{M}}\left\langle \nabla_M f_t(M_t), M\right\rangle_{\mathrm{F}}$;
		\STATE Update $M_{t+1}=(1-\eta) M_{t}+\eta M_{t}^\prime$;
		\ENDFOR
	\end{algorithmic}\label{alg:OFW_Memoryless_Control}
\end{algorithm}

%% file: Appendix/Appendix-Control-Lemma.tex
\subsection{Supporting Lemmas}\label{app-subsec:supporting_lemma}

In this part, we provide supporting lemmas used in the analysis of online non-stochastic control. In particular,
\begin{itemize}
	\item \Cref{lemma:M_norm} presents the relationship between the $\ell_{1}$, op norm and Frobenius norm in the $\mathcal{M}$-space.
	
	\item \Cref{lemma:bounded_Psi} shows the norm of transfer matrix in \cref{eq:prop_Psi} is upper bounded.	
	
	\item \Cref{lemma:bounded_variables} provides the boundedness of several variables of interest.
	
	\item \Cref{lemma:loss_Mspace_property} shows properties of the truncated functions $\left\{f_{t}\right\}$ and the feasible set $\mathcal{M}$.
\end{itemize}

\begin{lemma}[Norm Relations over $\calM$-space]\label{lemma:M_norm} 
For any $M=\left(M^{[0]}, \ldots, M^{[H-1]}\right) \in \mathcal{M} \subseteq\left(\mathbb{R}^{d_{u} \times d_{x}}\right)^{H}$, its $\ell_{1}$, op norm and Frobenius norm are defined by
\begin{equation}
	\|M\|_{\ell_{1}, \mathrm{op}}\triangleq\sum_{i=0}^{H-1}\left\|M^{[i]}\right\|_{\mathrm{op}}, \text { and }\|M\|_{\mathrm{F}}\triangleq\sqrt{\sum_{i=0}^{H-1}\left\|M^{[i]}\right\|_{\mathrm{F}}^{2}} .
\end{equation}
We then have the following inequalities on their relations:
\begin{equation}
	\|M\|_{\ell_{1}, \mathrm{op}} \leq \sqrt{H}\|M\|_{\mathrm{F}}, \text { and }\|M\|_{\mathrm{F}} \leq \sqrt{d}\|M\|_{\ell_{1}, \mathrm{op}},
\end{equation}
where $d=\min \left\{d_{u}, d_{x}\right\}$.
\end{lemma}

\begin{proof}
For any matrix $X \in \mathbb{R}^{m \times n}$,
\begin{equation}
	\|X\|_{\mathrm{op}} \leq\|X\|_{\mathrm{F}} \leq \sqrt{\min\{m,n\}}\|X\|_{\mathrm{op}}.	
\end{equation}

Therefore, by definition and Cauchy-Schwarz inequality, we obtain
\begin{equation}
	\|M\|_{\ell_{1}, \mathrm{op}}=\sum_{i=0}^{H-1}\left\|M^{[i]}\right\|_{\mathrm{op}} \leq \sum_{i=0}^{H-1}\left\|M^{[i]}\right\|_{\mathrm{F}} \leq \sqrt{H}\|M\|_{\mathrm{F}} .
\end{equation}

On the other hand, we have
\begin{equation}
	\|M\|_{\mathrm{F}}=\sqrt{\sum_{i=0}^{H-1}\left\|M^{[i]}\right\|_{\mathrm{F}}^{2}} \leq \sum_{i=0}^{H-1}\left\|M^{[i]}\right\|_{\mathrm{F}} \leq \sum_{i=0}^{H-1} \sqrt{d}\left\|M^{[i]}\right\|_{\mathrm{op}}=\sqrt{d}\|M\|_{\ell_{1}, \mathrm{op}} .
\end{equation}
\end{proof}

\begin{lemma}[Bounded Transfer Matrix]\label{lemma:bounded_Psi}
	Suppose $K_t$ is $(\kappa, \gamma)$-strongly stable at each iteration $t$. Suppose that for every $i \in\{0, \ldots, H-1\}$ and every $t \in \TimeSeq$, we have $\left\|M_{t}^{[i]}\right\|_{\mathrm{op}} \leq \tau(1-\gamma)^{i}$ for some $\tau>0$. Then, the transfer matrix is bounded as
	\begin{equation}
	\left\|\Psi_{t, i}^{K, h}\right\|_{\mathrm{op}} \leq \kappa^{2}(1-\gamma)^{i} \mathbf{1}_{i \leq h}+H \kappa_{B} \kappa^{2} \tau(1-\gamma)^{i-1}.
	\end{equation}
\end{lemma}

\begin{proof}
	We follow the proof of \citep[Lemma~5.4]{agarwal2019online}. By definition of the transfer matrix $\Psi_{t, i}^{K, h}$ in \cref{eq:prop_Psi}, we have
	\begin{equation}
	\begin{aligned}
	\left\|\Psi_{t, i}^{K, h}\right\|_{\mathrm{op}} &=\left\|\widetilde{A}_{K_{t:t-i+1}} \mathbf{1}_{i \leq h}  +\sum_{j=0}^{h} \widetilde{A}_{K_{t:t-j+1}} B_{t-j} M_{t-j}^{[i-j-1]} \mathbf{1}_{1 \leq i-j \leq H}\right\|_{\mathrm{op}} \\
	& \leq\left\|\widetilde{A}_{K_{t:t-i+1}} \right\|_{\mathrm{op}} \mathbf{1}_{i \leq h} +\sum_{j=0}^{h}\left\|\widetilde{A}_{K_{t:t-j+1}}\right\|_{\mathrm{op}} \left\|B_{t-j} \right\|_{\mathrm{op}} \left\| M_{t-j}^{[i-j-1]} \right\|_{\mathrm{op}} \mathbf{1}_{1 \leq i-j \leq H} \\
	& \leq \kappa^{2}(1-\gamma)^{i} \mathbf{1}_{i \leq h} + \sum_{j=0}^{H-1} \kappa^{2}(1-\gamma)^{j} \kappa_{B} \tau(1-\gamma)^{i-j-1} \\
	& \leq \kappa^{2}(1-\gamma)^{i} \mathbf{1}_{i \leq h} +\kappa^{2} \kappa_{B} \tau \sum_{j=1}^{H}(1-\gamma)^{i-1} \\
	&=\kappa^{2}(1-\gamma)^{i} \mathbf{1}_{i \leq h} + H \kappa^{2} \kappa_{B} \tau(1-\gamma)^{i-1}.
	\end{aligned} \qedhere
	\end{equation}
\end{proof}

\begin{lemma}[Bounded State and Control]\label{lemma:bounded_variables}
Suppose $K_t$ and $K_t^{*}$ are $(\kappa, \gamma)$-strongly stable linear controllers at each iteration $t \in \TimeSeq$. Suppose that for every $i \in\{0, \ldots, H-1\}$ and every $t \in \TimeSeq$, we have $\left\|M_{t}^{[i]}\right\|_{\mathrm{op}} \leq \tau(1-\gamma)^{i}$ for some $\tau>0$. Define $D\triangleq\frac{W\left(\kappa^{3}+H \kappa_{B} \kappa^{3} \tau\right)}{\gamma\left(1-\kappa^{2}(1-\gamma)^{H+1}\right)}+\frac{W \tau}{\gamma}.$ Then, we have
\begin{gather}
	\left\|x_{t}\left(M_{0: t-1}\right)\right\|_2 \leq D,\left\|y_{t}\left(M_{t-H-1: t-1}\right)\right\|_2 \leq D,  \left\|x_{t}^{K^{*}}\right\|_2 \leq D; \label{eq:lemma_bounded_variables_1} \\	
	\left\|u_{t}\left(M_{0: t}\right)\right\|_2 \leq D, \left\|v_{t}\left(M_{t-H-1: t}\right)\right\|_2 \leq D; \label{eq:lemma_bounded_variables_2}\\
	\left\|x_{t}\left(M_{0: t-1}\right)-y_{t}\left(M_{t-1-H: t-1}\right)\right\|_2 \leq \kappa^{2}(1-\gamma)^{H+1} D; \label{eq:lemma_bounded_variables_3}\\
	\left\|u_{t}\left(M_{0: t}\right)-v_{t}\left(M_{t-1-H: t}\right)\right\|_2 \leq \kappa^{3}(1-\gamma)^{H+1} D. \label{eq:lemma_bounded_variables_4}
\end{gather}
\end{lemma}

\begin{proof}
The proof is analogous to \citep[Lemma~5.5]{agarwal2019online}. We first consider \cref{eq:lemma_bounded_variables_1}:
\begin{equation}
	\begin{aligned}
		\left\|x_{t} \left(M_{0: t-1}\right)\right\|_2 &= \left\|\widetilde{A}_{K_{t-1:t-H-1}} x_{t-H-1}\left(M_{0: t-H-2}\right)+\sum_{i=0}^{2H} \Psi_{t-1, i}^{K_t, H}\left(M_{t-H-1: t-1}\right) w_{t-i-1}\right\|_2 \\
		& \leq \kappa^{2}(1-\gamma)^{H+1}\left\|x_{t-H-1}\left(M_{0: t-H-2}\right)\right\|_2 + W \sum_{i=0}^{2 H}\left\|\Psi_{t-1, i}^{K_t, H}\left(M_{t-H-1: t-1}\right)\right\|_{\mathrm{op}} \\
		& \leq \kappa^{2}(1-\gamma)^{H+1}\left\|x_{t-H-1}\left(M_{0: t-H-2}\right)\right\|_2 + W \sum_{i=0}^{2 H}\left(\kappa^{2}(1-\gamma)^{i}+H \kappa_{B} \kappa^{2} \tau(1-\gamma)^{i-1}\right) \\
		& \leq \kappa^{2}(1-\gamma)^{H+1}\left\|x_{t-H-1}\left(M_{0: t-H-2}\right)\right\|_2 + W \frac{\kappa^{2}+H \kappa_{B} \kappa^{2} \tau}{\gamma} \\
		&\leq \frac{W\left(\kappa^{2}+H \kappa_{B} \kappa^{2} \tau\right)}{\gamma\left(1-\kappa^{2}(1-\gamma)^{H+1}\right)} \leq D,
	\end{aligned}
\end{equation}
where the fourth inequality is a summation of geometric series and the ratio of this series is $\kappa^2(1-\gamma)^{H+1}$. 

Similarly, we have
\begin{equation}
\begin{aligned}
	\left\|y_{t} \left(M_{t-H-1: t-1}\right)\right\|_2 &= \left\|\sum_{i=0}^{2H} \Psi_{t-1, i}^{K_t, H}\left(M_{t-H-1: t-1}\right) w_{t-i-1}\right\|_2 \\
	& \leq W \sum_{i=0}^{2 H}\left\|\Psi_{t-1, i}^{K_t, H}\left(M_{t-H-1: t-1}\right)\right\|_{\mathrm{op}} \\
	& \leq W \sum_{i=0}^{2 H}\left(\kappa^{2}(1-\gamma)^{i}+H \kappa_{B} \kappa^{2} \tau(1-\gamma)^{i-1}\right) \\
	& \leq  W \frac{\kappa^{2}+H \kappa_{B} \kappa^{2} \tau}{\gamma} \\
	&\leq \frac{W\left(\kappa^{2}+H \kappa_{B} \kappa^{2} \tau\right)}{\gamma} \leq D,
\end{aligned}
\end{equation}
and
\begin{equation}
\left\|x_{t}^{*}\right\|_2 = \left\|\sum_{i=0}^{t-1}  \widetilde{A}_{K_{t-1:t-i}^{*}} w_{t-i-1}\right\|_2 \leq W \sum_{i=0}^{t-1} \kappa^{2}(1-\gamma)^{i} \leq \frac{W \kappa^{2}}{\gamma} \leq D.
\end{equation}

Next, we can show \cref{eq:lemma_bounded_variables_3} as follows,
\begin{equation}
	\left\|x_{t}\left(M_{0: t-1}\right)-y_{t}\left(M_{t-H-1: t-1}\right)\right\|_2 = \left\|\widetilde{A}_{K_{t-1:t-H-1}} x_{t-H-1}\left(M_{0: t-H-2}\right)\right\|_2 \leq \kappa^{2}(1-\gamma)^{H+1} D .
\end{equation}

We now consider \cref{eq:lemma_bounded_variables_2,eq:lemma_bounded_variables_4}:
\begin{equation}
	\begin{aligned}
	\left\|u_{t}\left(M_{0: t}\right)\right\|_2 &=\left\|-K_t x_{t}\left(M_{0: t-1}\right)+\sum_{i=1}^{H} M_{t}^{[i-1]} w_{t-i}\right\|_2 \\
	& \leq \kappa\left\|x_{t}\left(M_{0: t-1}\right)\right\|_2 + \sum_{i=1}^{H} W \tau(1-\gamma)^{i-1} \\
	& \leq \frac{W\left(\kappa^{3}+H \kappa_{B} \kappa^{3} \tau\right)}{\gamma\left(1-\kappa^{2}(1-\gamma)^{H+1}\right)}+\frac{W \tau}{\gamma} \leq D,
	\end{aligned}
\end{equation}
\begin{equation}
\left\|v_{t}\left(M_{t-H-1: t}\right)\right\|_2 \leq \kappa\left\|y_{t}\left(M_{t-H-1: t-1}\right)\right\|_2 + \sum_{i=1}^{H} W \tau(1-\gamma)^{i-1} \leq D,
\end{equation}
and
\begin{equation}
\left\|u_{t}\left(M_{0: t-1}\right)-v_{t}\left(M_{t-H-1: t-1}\right)\right\|_2=\left\|-K_t \left(x_{t}\left(M_{0: t-1}\right)-y_{t}\left(M_{t-H-1: t-1}\right)\right)\right\|_2 \leq \kappa^{3}(1-\gamma)^{H+1} D .
\end{equation}
\end{proof}

\begin{lemma}\label{lemma:loss_Mspace_property}
Define $D\triangleq\frac{W \kappa^{3}\left(1+H \kappa_{B} \tau\right)}{\gamma\left(1-\kappa^{2}(1-\gamma)^{H+1}\right)}+\frac{W \tau}{\gamma}$. The truncated loss $f_{t}: \mathcal{M}^{H+2} \mapsto \mathbb{R}$ is $L_f$-coordinate-wise Lipschitz and has bounded gradient norm $G_f$.  In addition, the radius of feasible set in $\mathcal{M}$-space is bounded by $D_f$. Formally,
\begin{enumerate}
	\item The truncated loss function is $L_{f}$-coordinate-wise Lipschitz with respect to the Euclidean (i.e., Frobenius) norm, \ie
	\begin{equation}
		\left|f_{t}\left(M_{t-H-1}, \ldots, M_{t-k}, \ldots, M_{t}\right) - f_{t}\left(M_{t-H-1}, \ldots, \widetilde{M}_{t-k}, \ldots, M_{t}\right)\right| \leq L_{f}\left\|M_{t-k}-\widetilde{M}_{t-k}\right\|_{\mathrm{F}},
	\end{equation}
	where $	L_{f} \leq 3 G_{c} D \sqrt{H} \kappa_{B} \kappa^{3}(1-\gamma)^{k-1} W$.
	
	\item The gradient norm of surrogate loss $\widetilde{f}_{t}: \mathcal{M} \mapsto \mathbb{R}$ is bounded by $G_{f}$, \ie $\left\|\nabla_{M} \widetilde{f}_{t}(M)\right\|_{\mathrm{F}} \leq G_{f}$ holds for any $M \in \mathcal{M}$ and any $t \in \TimeSeq$, where $G_{f} \leq 3 H d_x d_u G_{c} W \kappa_{B} \kappa^{3} \gamma^{-1}$.
	
	\item The diameter of the feasible set is at most $D_{f}$, \ie $\left\|M-M^{\prime}\right\|_{\mathrm{F}} \leq D_{f}$ holds for any $M, M^{\prime} \in \mathcal{M}$, where $D_{f} \leq 2 \sqrt{d} \kappa_{B} \kappa^{3} \gamma^{-1}$ .
\end{enumerate}
\end{lemma}

\begin{proof}

The proof follows the steps in \citep[Lemma~5.6~and~5.7]{agarwal2019online} and \citep[Lemma~20]{zhao2022non}. We first prove claim 1. To this end, we use the following notations:
\begin{equation}
	\begin{array}{l}
		M_{t-H-1: t}\triangleq\left\{M_{t-H-1} \ldots M_{t-k} \ldots M_{t}\right\}; \\
		M_{t-H-1: t-1}\triangleq\left\{M_{t-H-1} \ldots M_{t-k} \ldots M_{t-1}\right\}; \\
		\widetilde{M}_{t-H-1: t}\triangleq\left\{M_{t-H-1} \ldots \widetilde{M}_{t-k} \ldots M_{t}\right\}; \\
		\widetilde{M}_{t-H-1: t-1}\triangleq\left\{M_{t-H-1} \ldots \widetilde{M}_{t-k} \ldots M_{t-1}\right\}; \\
		y_{t}\triangleq y_{t}\left({M}_{t-H-1: t-1}\right); \\
		\widetilde{y}_{t}\triangleq y_{t}\left(\widetilde{M}_{t-H-1: t-1}\right); \\
		v_{t}\triangleq v_{t}\left(M_{t-H-1: t}\right); \\
		\widetilde{v}_{t}\triangleq v_{t}\left(\widetilde{M}_{t-H-1: t}\right).
	\end{array}
\end{equation}

By definition of $f_{t}$, we have
\begin{equation}
		f_{t}\left(M_{t-H-1: t}\right)-f_{t}\left(\widetilde{M}_{t-H-1: t}\right)  = c_{t}\left(y_{t}, v_{t}\right)-c_{t}\left(\widetilde{y}_{t}, \widetilde{v}_{t}\right) \leq  G_{c} D\left\|y_{t}-\widetilde{y}_{t}\right\|_2 + G_{c} D\left\|v_{t}-\widetilde{v}_{t}\right\|_2.
\end{equation}

Then consider $\left\|y_{t}-\widetilde{y}_{t}\right\|$ and $\left\|v_{t}-\widetilde{v}_{t}\right\|$:
\begin{equation}
\begin{aligned}
	\left\|y_{t}^{K}-\widetilde{y}_{t}^{K}\right\|_2 &=\left\|\sum_{i=0}^{2 H}\left(\Psi_{t-1, i}^{K_t, H}\left(M_{t-H-1: t-1}\right)-\Psi_{t-1, i}^{K_t, H}\left(\widetilde{M}_{t-H-1: t-1}\right)\right) w_{t-1-i}\right\|_2 \\
	&=\left\|\widetilde{A}_{{K_{t-1:t-k+1}}} B_{t-k} \sum_{i=0}^{2 H}\left(M_{t-k}^{[i-k]}-\widetilde{M}_{t-k}^{[i-k]}\right) \mathbf{1}_{0 \leq i-k \leq H-1} w_{t-1-i}\right\|_2 \\
	& \leq \kappa_{B} \kappa^{2}(1-\gamma)^{k-1} W  \sum_{i=1}^{H}\left\|M_{t-k}^{[i-1]}-\widetilde{M}_{t-k}^{[i-1]}\right\|_{\mathrm{op}} \\
%	& \leq \kappa_{B} \kappa^{2} W\left\|M_{t-k}-\widetilde{M}_{t-k}\right\|,
    & \leq \sqrt{H} \kappa_{B} \kappa^{2}(1-\gamma)^{k-1} W  \left\|M_{t-k}-\widetilde{M}_{t-k}\right\|_{\mathrm{F}},
\end{aligned}
\end{equation}
and we have
\begin{equation}
	\begin{aligned}
		\left\|v_{t}-\widetilde{v}_{t}\right\|_2 &=\left\|-K\left(y_{t}-\widetilde{y}_{t}\right)+\mathbf{1}_{\{k=0\}} \sum_{i=1}^{H}\left(M_{t-k}^{[i-1]}-\widetilde{M}_{t-k}^{[i-1]}\right) w_{t-i} \right\|_2 \\
		& \leq\left(\sqrt{H}\kappa_{B} \kappa^{3}(1-\gamma)^{k-1} W + \sqrt{H} W\right)\left\|M_{t-k}-\widetilde{M}_{t-k}\right\|_{\mathrm{F}} \\
		& \leq 2 \sqrt{H} \kappa_{B} \kappa^{3}(1-\gamma)^{k-1} W \left\|M_{t-k}-\widetilde{M}_{t-k}\right\|_{\mathrm{F}} .
	\end{aligned}
\end{equation}

Combining the above equations, we obtain
\begin{equation}
	\begin{aligned}
		f_{t}\left(M_{t-H-1: t}\right)-f_{t}\left(\widetilde{M}_{t-H-1: t}\right) & \leq G_{c} D\left\|y_{t}^{K}-\widetilde{y}_{t}^{K}\right\|_2 +G_{c} D\left\|v_{t}^{K}-\widetilde{v}_{t}^{K}\right\|_2 \\
		& \leq G_{c} D \sqrt{H} \kappa_{B} \kappa^{2}(1-\gamma)^{k-1} W\left\|M_{t-k}-\widetilde{M}_{t-k}\right\|_{\mathrm{F}} \\
		& \quad\quad\quad+2 G_{c} D \kappa_{B} \kappa^{3}(1-\gamma)^{k-1} W\left\|M_{t-k}-\widetilde{M}_{t-k}\right\|_{\mathrm{F}} \\
		& \leq 3 G_{c} D \sqrt{H} \kappa_{B} \kappa^{3}(1-\gamma)^{k-1} W\left\|M_{t-k}-\widetilde{M}_{t-k}\right\|_{\mathrm{F}} .
	\end{aligned}
\end{equation}
Therefore, we have $L_{f} \leq 3 G_{c} D \sqrt{H} \kappa_{B} \kappa^{3}(1-\gamma)^{k-1} W$.

Now consider claim 2. We need to bound $\nabla_{M_{p, q}^{[r]}} \widetilde{f}_{t}(M)$ for every $p \in\{1, \dots,d_{u}\}, q \in\{1, \dots,d_{x}\}$, and $r \in\{0, \ldots, H-1\}$,
\begin{equation}
	\left|\nabla_{M_{p, q}^{[r]}} \widetilde{f}_{t}(M)\right| \leq G_{c}\left\|\frac{\partial y_{t}(M)}{\partial M_{p, q}^{[r]}}\right\|_{\mathrm{F}}+G_{c}\left\|\frac{\partial v_{t}(M)}{\partial M_{p, q}^{[r]}}\right\|_{\mathrm{F}} .
\end{equation}

Now we aim to bound the two terms of the right-hand side respectively:
\begin{equation}
	\begin{aligned}
		\left\|\frac{\partial y_{t}(M)}{\partial M_{p, q}^{[r]}}\right\|_{\mathrm{F}} & \leq \left\|\sum_{i=0}^{2 H} \sum_{j=0}^{H}\left[\frac{\partial \widetilde{A}_{K_{t:t-j+1}} B_{t-j} M^{[i-j-1]}}{\partial M_{p, q}^{[r]}}\right] w_{t-1-i} \mathbf{1}_{0 \leq i-j \leq H-1}\right\|_{\mathrm{F}} \\
		& \leq W \kappa_{B} \kappa^{2}\left\|\frac{\partial M^{[r]}}{\partial M_{p, q}^{[r]}}\right\|_{\mathrm{F}} \sum_{i=r+1}^{r+H+1}(1-\gamma)^{i-r-1} \\
		& \leq \frac{W \kappa_{B} \kappa^{2}}{\gamma}\left\|\frac{\partial M^{[r]}}{\partial M_{p, q}^{[r]}}\right\|_{\mathrm{F}} \\
		& \leq \frac{W \kappa_{B} \kappa^{2}}{\gamma};\\
	 \left\|\frac{\partial v_{t}(M)}{\partial M_{p, q}^{[r]}}\right\|_{\mathrm{F}} & \leq \kappa\left\|\frac{\partial y_{t}(M)}{\partial M_{p, q}^{[r]}}\right\|_{\mathrm{F}}+\sum_{i=1}^{H}\left\|\frac{\partial M^{[i-1]}}{\partial M_{p, q}^{[r]}} w_{t-i}\right\|_{\mathrm{F}} \\
		& \leq \frac{W \kappa_{B} \kappa^{3}}{\gamma}+W\left\|\frac{\partial M^{[r]}}{\partial M_{p, q}^{[r]}}\right\|_{\mathrm{F}} \\
		& \leq W\left(\frac{\kappa_{B} \kappa^{3}}{\gamma}+1\right).
	\end{aligned}
\end{equation}

Therefore, we have
\begin{equation}
	\left|\nabla_{M_{p, q}^{[r]}} \widetilde{f}_{t}(M)\right| \leq G_{c} \frac{W \kappa_{B} \kappa^{2}}{\gamma}+G_{c} W\left(\frac{\kappa_{B} \kappa^{3}}{\gamma}+1\right) \leq 3 G_{c} W \kappa_{B} \kappa^{3} \gamma^{-1} .
\end{equation}
Thus, $\left\|\nabla_{M} \widetilde{f}_{t}(M)\right\|_{\mathrm{F}}$ is at most $3 H d_x d_u G_{c} W \kappa_{B} \kappa^{3} \gamma^{-1}$.

Finally, we prove claim 3. By construction of $M^{[i]}, \forall i \in \{0, \dots, H-1\}$, we require $\|M^{[i]}\|_{\mathrm{op}} \leq \kappa_{B} \kappa^{3}(1-\gamma)^{i}$. Therefore, utilizing \Cref{lemma:M_norm} we have
\begin{equation}
	\begin{aligned}
		\max _{M_{1}, M_{2} \in \mathcal{M}}\left\|M_{1}-M_{2}\right\|_{\mathrm{F}} &\leq \sqrt{d} \max _{M_{1}, M_{2} \in \mathcal{M}}\left\|M_{1}-M_{2}\right\|_{\ell_{1}, \mathrm{op}} \\
		& \leq \sqrt{d} \max _{M_{1}, M_{2} \in \mathcal{M}}\left(\left\|M_{1}\right\|_{\ell_{1}, \mathrm{op}}+\left\|M_{2}\right\|_{\ell_{1}, \mathrm{op}}\right) \\
		&=\sqrt{d} \max _{M_{1}, M_{2} \in \mathcal{M}}\left(\sum_{i=0}^{H-1}\left\|M_{1}^{[i]}\right\|_{\mathrm{op}}+\left\|M_{2}^{[i]}\right\|_{\mathrm{op}}\right) \\
		&\leq \sqrt{d} \max _{M_{1}, M_{2} \in \mathcal{M}}\left(2 \sum_{i=0}^{H-1} \kappa_{B} \kappa^{3}(1-\gamma)^{i}\right) \\
		&=2 \sqrt{d} \kappa_{B} \kappa^{3} \sum_{i=0}^{H-1}(1-\gamma)^{i} \\
		&\leq 2 \sqrt{d} \kappa_{B} \kappa^{3} \gamma^{-1}.
	\end{aligned}
\end{equation}
Hence, we finish the proof of all three claims in the statement.
\end{proof}

%% file: Appendix/Appendix-Exp.tex
\section{Numerical Evaluations}\label{app:sec_control_exp}

\input{Table/Table-results.tex}

We evaluate \MetaOFW (\Cref{alg:Meta_OFW_Control}) in simulated scenarios of online control of linear time-invariant systems.  
% All methods are tested on a desktop computer with a 2.60GHz Intel(R) Xeon(R) Gold 6240 CPU in Python script.
% \blue{Our code will be open-sourced via a link here.}

\myParagraph{Compared Algorithms} 
We compare \MetaOFW with the \OGD \citep{zinkevich2003online}, \Ader \citep{zhang2018adaptive}, and \Scream \citep{zhao2022non} algorithms. All algorithms rely on the DAC policy \citep{agarwal2019online}.

\myParagraph{Simulation Setup} 
We follow the setup as \cite{zhao2021non} and consider linear systems of the form
\begin{equation}\label{eq:aux_exp}
    \begin{aligned}
        x_{t+1} &= A x_{t} + B u_{t} + w_{t} \\
        &= A x_{t} + B u_{t} + (\Delta_{t,A}x_{t} + \Delta_{t,B}u_{t} + \tilde{w}_{t}),
    \end{aligned}
\end{equation}
    where $\tilde{w}_{t}$ and the elements of $\Delta_{t,A}$ and $\Delta_{t,B}$ are sampled from various distributions, specifically, Gaussian, Uniform, Gamma, Beta, Exponential, and Weibull distributions. We use memory length $H=10$. The loss function has the form $c_t(x_t,u_t) = 
 q_t x_t^\top x_t + r_t u_t^\top u_t$, where $q_t\in\mathbb{R}$ and $r_t\in\mathbb{R}$ are time-varying weights. Particularly, we consider two cases:
 
%  \red{after finalizing, fix extra white space, if still any}
 
\begin{enumerate}%[leftmargin=9pt]\setlength\itemsep{-1.mm}
    \item Sinusoidal weights defined as 
    \begin{equation}
        q_t = \sin (t/10\pi),\ r_t = \sin (t/20\pi).
        \label{eq_loss_1}
    \end{equation}
    \item Step weights defined as 
    \begin{equation}
        \begin{aligned}
            (q_t, r_t) = \left\{ \begin{array}{cc} \left(\frac{\log(2)}{2}, 1 \right)  , \ & t \leq T/5,  \\ 
            \left(1, 1 \right) , \ & T/5 < t \leq 2T/5, \\
            \left(\frac{\log(2)}{2}, \frac{\log(2)}{2} \right)  , \ & 2T/5 < t \leq 3T/5, \\  
            \left(1, \frac{\log(2)}{2} \right)  , \ & 3T/5 < t \leq 4T/5,  \\
            \left(\frac{\log(2)}{2}, 1 \right)   , \ & 4T/5 < t \leq T.
            \end{array} \right.
        \end{aligned}
        \label{eq_loss_2}
    \end{equation}
\end{enumerate}

\myParagraph{Results} 
We first compare \MetaOFW with the \OGD, \Ader, and \Scream algorithms in terms of cumulative loss. The results are summarized in \Cref{table_preformance_noise}, showing that \MetaOFW achieved the lowest cumulative loss across all tested cases, except under gamma distribution with sinusoidal weights; in the best-case ---{exponential distribution with step weights--- \MetaOFW is 52 times better than \Scream.}

We also vary the dimensions of the state $x_t$ and input the $u_t$, and compare \MetaOFW and \Scream in terms of cumulative loss and computation time. The results are summarized in \Cref{table_preformance}. As $d_x$ and $d_u$ increase, \MetaOFW is computationally three times faster than \Scream, achieving also lower cumulative loss than \Scream in all cases.

%% file: Table/Table-results.tex
\renewcommand{\arraystretch}{1.2} 
\begin{table*}[t]
  \centering
    %  \captionsetup{font=footnotesize}
     \caption{Comparison of the \OGD \citep{zinkevich2003online}, \Ader \citep{zhang2018adaptive}, \Scream \citep{zhao2022non}, and \MetaOFW algorithms in terms of cumulative loss for $10000$ time steps. The \blue{blue} numbers correspond to the \blue{best} performance and the \red{red} numbers correspond to the \red{worse}.}
     \label{table_preformance_noise}
     % \resizebox{\columnwidth}{!}{
     {
     \begin{tabular}{ccccccccc}
     \toprule
 	 \multirow{2}{*}{Noise Distribution} & \multicolumn{4}{c}{Sinusoidal Weights (\cref{eq_loss_1})} & \multicolumn{4}{c}{Step Weights (\cref{eq_loss_2})}		\cr
    \cmidrule(lr){2-5} \cmidrule(lr){6-9} & \MetaOFW & \Scream & \Ader & \OGD & \MetaOFW & \Scream & \Ader & \OGD  \cr
    \midrule
	 Gaussian & \blue{15625} & 19725 & 21052 & \red{33574} & \blue{9496} & 10704 & 11453 & \red{26790}	\cr
	 Uniform & \blue{18299} & 93987 & \red{107096} & 30419 & \blue{13395} & 39057 & 35313 & \red{39885} 	\cr
	 Gamma & 16239 & \blue{16138} & \red{18039} & 17484 & \blue{9184} & 61989 & \red{75505} & 45398 	\cr
	 Beta & \blue{21448} & \red{34146} & 30990 & 30253 & \blue{15982} & 29301 & \red{30799} & 28859 \cr
	 Exponential & \blue{10621} & \red{254815} & 252227 & 28859 & \blue{4366} & \red{227860} & 204844 & 53626 	\cr
	 Weibull & \blue{14068} & \red{91474} & 94040 & 38549 & \blue{5623} & 182887 & \red{993734} & 92341	\cr
    \bottomrule
    \end{tabular}}
     % }
\end{table*}

\renewcommand{\arraystretch}{1.2} 
\begin{table}[t]
  \centering
    %  \captionsetup{font=footnotesize}
     \caption{Comparison of the \Scream \citep{zhao2022non} and \MetaOFW algorithms in terms of computational time and cumulative loss over $200$ time steps for the case of Gaussian noise in \cref{eq:aux_exp}, sinusoidal weights in \cref{eq_loss_1}, and across varying system dimensions $d_x$ and $d_u$.}
     \label{table_preformance}
     % \resizebox{\columnwidth}{!}{
     {
     \begin{tabular}{ccccc}
     \toprule
 	 \multirow{2}{*}{$(d_x, \ d_u)$} & \multicolumn{2}{c}{Time (seconds)} & \multicolumn{2}{c}{Cumulative Loss}		\cr
    \cmidrule(lr){2-3} \cmidrule(lr){4-5} & \MetaOFW & \Scream & \MetaOFW & \Scream  \cr
 	\midrule
	 $(2,\ 1)$ & 52.49 & 17.64 & 915.52 & 1819.41	\cr
	 $(4,\ 2)$ & 135.04 & 177.47 & 1388.56 & 3231.89	\cr
	 $(6,\ 3)$ & 287.67 & 605.48 & 1235.83 & 2021.47	\cr
	 $(8,\ 4)$ & 504.82 & 1351.29 & 1421.05 & 1873.79	\cr
	 $(10,\ 5)$ & 786.87 & 2219.85 & 1202.43 & 1439.22	\cr
	 $(12,\ 6)$ & 998.22 & 3029.89 & 893.17 & 984.96	\cr
	 $(14,\ 7)$ & 2022.5 & 5531.36 & 797.80 & 958.09  	\cr
     \bottomrule
     \end{tabular}}
     % }
\end{table}